%% file: arxiv_main.tex
\documentclass{article}

\usepackage[margin=1in]{geometry}

\PassOptionsToPackage{authoryear,numbers,sort&compress}{natbib}
\usepackage{natbib}

\usepackage[modulo]{lineno}

\usepackage{amsmath, amsthm, bbm, amssymb, mathtools}

\usepackage[utf8]{inputenc}
\usepackage[T1]{fontenc}

\usepackage{algorithm}
\usepackage{algpseudocode}
\usepackage{float}     %
\usepackage{placeins}  %
\usepackage{url}
\usepackage{booktabs}
\usepackage{amsfonts}
\usepackage{nicefrac}
\usepackage{microtype}
\usepackage{xcolor}
\usepackage{graphicx}
\usepackage{subcaption}
\usepackage{wrapfig}

\newif\ifarxiv
\arxivtrue

\ifarxiv
\renewenvironment{wrapfigure}[2]{%
    \begin{figure}[t]
}{%
    \end{figure}
}
\fi

\usepackage{tikz}
\usepackage{xspace}
\usepackage[percent]{overpic}
\usepackage[inline]{enumitem}
\usepackage{thm-restate}

\usepackage[skip=0.75\baselineskip, indent=0pt]{parskip}

\usepackage{authblk}

\usepackage{amartya_ltx2}

\usepackage{hyperref}
\hypersetup{draft}
\usepackage[capitalise,nameinlink,noabbrev]{cleveref}

\newtheorem{theorem}{Theorem}
\newtheorem{example}{Example}
\newtheorem{corollary}{Corollary}
\newtheorem{lemma}{Lemma}
\newtheorem{proposition}{Proposition}

\newcommand{\kdppca}{\textrm{k-DP-PCA}\xspace}
\newcommand{\kdpojas}{\textrm{k-DP-Ojas}\xspace}

\newcommand{\AdaptPrivOja}{\textnormal{\textsc{AdaDPO}}\xspace}
\newcommand{\TangAlg}{\textnormal{\textsc{TAdaDPO}}\xspace}

\newcommand{\TruncatedMean}{\textnormal{\textsc{TruncatedMean}}\xspace}
\newcommand{\PrivRange}{\textnormal{\textsc{PrivRange}}\xspace}

\renewcommand{\softO}[1]{\tilde O\!\left(#1\right)}
\DeclareMathOperator{\tr}{tr}
\DeclareMathOperator{\clip}{clip}

\theoremstyle{definition}
\newtheorem{definition}{Definition}

\theoremstyle{remark}
\newtheorem*{remark}{Remark}

\let\epsilon\varepsilon

\theoremstyle{plain}
\newtheorem{assumption}{Assumption}
\renewcommand{\theassumption}{\Alph{assumption}}
\crefname{assumption}{Assumption}{Assumptions}
\Crefname{assumption}{Assumption}{Assumptions}

\newlist{asslist}{enumerate}{1}
\setlist[asslist]{%
  label=\textbf{\theassumption.\arabic*},
  ref=\theassumption.\arabic*,
  wide,
}
\crefname{asslisti}{Assumption}{Assumptions}
\Crefname{asslisti}{Assumption}{Assumptions}

\title{Optimal Rates for Adaptive Private \(k\)-PCA}

\author[1]{Johanna D\"ungler}
\author[1]{Amartya Sanyal}
\affil[1]{Department of Computer Science, University of Copenhagen}
\affil[ ]{\texttt{\{jodu, amsa\}@di.ku.dk}}

\date{} %

\begin{document}

\nocite{dwork2006calibrating}

\maketitle

\input{sections/abstract}

\input{sections/introduction}

\input{sections/problem_formulation}

\input{sections/2-2-warmup.tex}

\input{sections/adaptive_private_ojas}

\input{sections/main_tangent_results_extended}

\input{sections/related_work_and_conclusion}

\section{Acknowledgement}

JD acknowledges support from the Danish Data Science Academy, which is funded by the Novo Nordisk Foundation (NNF21SA0069429) and VILLUM FONDEN (40516). AS acknowledges the Novo Nordisk Foundation for support via the Startup grant (NNF24OC0087820) and VILLUM FONDEN via the Young Investigator program (72069). The authors would also like to thank Rasmus Pagh for very insightful discussions.

\bibliographystyle{plainnat}  
\bibliography{refs}

\appendix

\input{appendices/math_preliminaries}

\input{appendices/additional_comparisons_to_prior_work}

\input{appendices/proofs}

\input{appendices/existing_lower_bounds}

\input{appendices/novel_ojas_bound_using_frobenius}

\input{appendices/experiments}

\end{document}

%% file: sections/abstract.tex
\begin{abstract}
  Given $n$ i.i.d. random matrices $A_i \in \mathbb{R}^{d \times d}$
  with common expectation $\Sigma$, the goal of \emph{Differentially
  Private Stochastic PCA} is to identify a $k$-dimensional subspace
  capturing the leading variance directions of $\Sigma$, while
  preserving differential privacy~(DP) for each individual sample
  $A_i$. \citet{dungler2025iterative} introduced \kdppca, the first
  algorithm to simultaneously (1) achieve sample complexity
  $n=\widetilde O(d)$ for sub-Gaussian data, (2) adapt its privacy
  noise to the intrinsic randomness of the data, and (3) extend
  seamlessly to any target dimension $k\le d$. However, its sample
  complexity has a suboptimal dependence on $k$. We propose the first
  algorithm that achieves optimal sample complexity in both $d$ and
  $k$, while retaining properties (2) and (3) of \kdppca.
  Additionally, prior utility guarantees require a minimum sample size
  with exponential dependence on the eigengap. Our method removes this
  dependence. It also improves the dependence on spectral parameters,
  matching known lower bounds for the spiked covariance model. Unlike
  deflation-based approaches like \kdppca, our algorithm updates the
  full $d\times k$ subspace jointly rather than one eigenvector at a
  time. This non-deflation structure simplifies the algorithm, reduces
  the number of hyper-parameters, and improves computational
  efficiency, particularly when using block linear algebra libraries.

\end{abstract}

%% file: sections/introduction.tex
\section{Introduction}

Principal Component Analysis (PCA) is a basic technique in statistics
and machine learning. Given data vectors \(x_1,\ldots,x_n\in\reals^d\), classical PCA estimates the leading eigenspace of the empirical
second-moment matrix \(\frac{1}{n}\sum_i x_ix_i^\top\).  The task of recovering
the top \(k\) eigenvectors is typically referred to as \(k\)-PCA. In
this paper, we study a stochastic variant of this problem. The input
is a sequence of independent symmetric matrices
\(A_1,\ldots,A_n\in\reals^{d\times d}\) with common expectation
\(\Sigma\), and the goal is to estimate the top-\(k\) eigenspace of
\(\Sigma\). In addition, we want to protect the privacy of the individual samples \(A_i\) in the sense of differential privacy~\citep{dwork2006calibrating}.

Differential privacy offers a formal framework for protecting the
privacy of individuals represented in a dataset, and has been deployed
in settings involving sensitive information, including census data
analysis and large-scale industrial
analytics~\citep{abowd2020modernization, team2017learning}. 
Private PCA has commonly been studied in the fixed-dataset setting for nearly two
decades~\citep{blum2005practical,chaudhuri2013near,hardt2013beyond,dwork2014analyze}, where algorithms typically rely on deterministic norm bounds. When such methods are applied to stochastic PCA, these bounds
must be enforced through clipping, but this transfers a worst-case fixed-data
requirement into a setting where the observations are random. The resulting
privacy-utility guarantees are therefore often suboptimal: they either require
sample sizes that grow super-linearly with the ambient dimension \(d\), or add
privacy noise at a level that does not exploit the randomness already present in
the observations. Consequently, applying fixed-dataset private PCA methods
directly in the stochastic setting typically yields error bounds of order
\(O(\sqrt{dk/n} + d^{3/2}k/(\epsilon n))\), where \(\epsilon\) denotes the
privacy parameter, while the known optimal rate is of order
\(O(\sqrt{dk/n} + dk/(\epsilon n))\)~(\Cref{thm-gauss-lower-bound-k-1,spiked-cov-lower-bound}).

The first adaptive private algorithm for stochastic PCA was given
by~\citet{liu2022dp} for the case \(k=1\). Their key idea was to
privately estimate the fluctuation scale of the stochastic update and
to clip around a private center rather than around zero. This makes
the privacy noise scale with the randomness in the data.~\citet{dungler2025iterative} extended this adaptive
principle to \(k>1\). Their method, however, is deflation-based: it
estimates one direction, projects it out, and repeats. This sequential
structure introduces an additional dependence on \(k\): their
error bounds are of order \(O (\sqrt{dk^{3/2}/n} +
dk^{3/2}/(\epsilon n))\). 
Additionally, their algorithm requires separate parameter tuning for each eigenvector it estimates, making it both more cumbersome to tune and computationally ill-suited for block linear-algebra implementations that are efficient in practice.
\citet{cai2024optimal} obtained optimal rates for Gaussian
data when the unknown population covariance \(\Sigma\) satisfies the
flat-tail condition \(\lambda_{k+1}=\cdots=\lambda_d\). However,
because \(\Sigma\) is itself unknown, it is unclear if this condition can be
privately certified from the input data in general. As a result, their privacy
guarantees rely on an unverified distributional assumption on the
data-generating process.

\noindent\textbf{Our Contributions} 
Our first contribution is \AdaptPrivOja~(\Cref{algo:priv-ojas-adaptive}), a non-deflation-based adaptive
private variant of Oja's algorithm~(\Cref{algo:ojas})
that achieves the optimal dependence on \(k\) in sample complexity.
Unlike deflation-based methods such as \kdppca~\citep{dungler2025iterative},
which reduce \(k\)-PCA to \(k\) private one-dimensional problems, our algorithm estimates the target \(k\)-dimensional subspace jointly.
Moreover, its adaptive privacy mechanism estimates and uses one single
scale parameter and one clipping center for all \(k\) columns instead
of using \(k\) separate range estimators. It then adds calibrated
Gaussian noise that preserves symmetry. This joint treatment is what
avoids the extra cost of handling the \(k\) directions separately.  We
show that \AdaptPrivOja achieves the optimal dependence on \(d\) and
\(k\)~(\Cref{thm:utility-priv-ojas-adaptive}), and performs well on
simulated data.

Our second contribution is \TangAlg~(\Cref{alg:tangent-adaptive}), a refinement of \AdaptPrivOja that improves the dependence of the private error bound on the spectrum of \(\Sigma\). The main idea is that, after obtaining a sufficiently accurate initial estimate of the top-k eigenspace, the remaining estimation problem has a more favorable local geometry. \TangAlg exploits this local structure in a second phase to reduce the amount of privacy noise needed for the final refinement. We prove a general utility guarantee for this procedure in~\Cref{thm:tangent-general-utility}. In the spiked covariance model, this yields the optimal spectral dependence up to logarithmic factors, matching the lower bound of~\citet{cai2024optimal}.

\begin{corollary}[Informal version of~\Cref{cor:flat-tail}]
\label{cor:flat-tail-intro}
Assume \(A_i=x_ix_i^\top\) with \(x_i\sim\cN(0,\Sigma)\), where
\(\Sigma\) is a spiked covariance with a flat tail, meaning that for
some \(k\in[d]\), 
\(
    \lambda_{k+1}=\cdots=\lambda_d.
\)
Let \(V_k \in \mathbb{R}^{d \times k}\) be the true top-\(k\) eigenvectors of
\(\Sigma\), and let \(\lambda_k-\lambda_{k+1}>0\). Then
the output \(Q_T\) of \TangAlg satisfies
\[
\norm{Q_TQ_T^\top-V_k V_k^\top}_F
\le
\softO{
\frac{\sqrt{\lambda_1\lambda_{k+1}}}{\lambda_k-\lambda_{k+1}}
\bs{
\sqrt{\frac{dk}{n}}+
\frac{dk}{\epsilon n}
}
}.
\]
\end{corollary}

The rest of the paper is organized as follows: \Cref{sec:problem-formulation} defines the stochastic PCA problem, the
privacy model, the main assumptions, and a theoretical warm up for the following sections.~\Cref{sec:adaptive-priv-ojas} introduces \AdaptPrivOja, its guarantees,
and experiments.~\Cref{sec:tangent-adaptive} presents
\TangAlg and its guarantees. Finally,~\Cref{sec:conclusion} gives an overview of related work and presents the conclusion.

%% file: sections/problem_formulation.tex
\section{Problem Formulation}\label{sec:problem-formulation}

Let \(A_1,\ldots,A_n\in\reals^{d\times d}\) be independent symmetric
random matrices with common expectation \(\Sigma=\bE\bs{A_i}\). We assume that \(\Sigma\) is positive semidefinite with eigenvalues
\(\lambda_1\ge\cdots\ge\lambda_d\ge0\), and we denote by
\(V_k\in\reals^{d\times k}\) the matrix containing the top \(k\)
eigenvectors of \(\Sigma\). 
The goal of \emph{Stochastic $k$-PCA}  is to output \(Q\in\reals^{d\times
k}\) with orthonormal columns such that \(QQ^\top\) is close to
\(V_k V_k^\top\), with error measured by
\(\norm{QQ^\top-V_kV_k^\top}_F\). Throughout the paper, we assume a
nonzero eigengap  \(\Delta_k:=\lambda_k-\lambda_{k+1}>0\), and we write
\(\kappa:=\lambda_1/\Delta_k\) and \(\kappa_\perp:=(\lambda_1-\lambda_d)/\Delta_k\le\kappa\).

Let \(\bS_{d,k}\coloneqq\bc{Q\in\reals^{d\times k}:Q^\top
    Q=I_k} \). For \(Q\in\bS_{d,k}\), let \(\Pi_Q\coloneqq I-QQ^\top\) and
\[
    \mathcal S_Q
    \coloneqq
    \bc{Y\in\reals^{d\times k}:Q^\top Y\text{ is symmetric}}.
\]
For a square matrix \(B\), write
\(\operatorname{sym}(B)\coloneqq\frac{B+B^\top}{2}.\) The
Frobenius-orthogonal projection onto \(\mathcal S_Q\) is
\[
    \mathsf P_{\mathcal S_Q}(Y)
    \coloneqq
    Q\operatorname{sym}(Q^\top Y) + \Pi_QY.
\]
We use \(\norm{\cdot}_2\) for operator norm, \(\norm{\cdot}_F\) for
Frobenius norm, \(\ip{A}{B}=\mathrm{Tr}(A^\top B)\) for the Frobenius
inner product, and \(\norm{\cdot}_{(2,k)}\) for the Schatten-\((2,k)\) norm, and \(\norm{\cdot}_{S_p}\) for the
Schatten-\(p\) norm \(\br{\sum_i\sigma_i(\cdot)^p}^{1/p}\), so that
\(\norm{\cdot}_{S_2}=\norm{\cdot}_F\) and \(\norm{\cdot}_{(2,k)}\) is the rank-\(k\) truncation
of \(\norm{\cdot}_{S_2}\). The notation \(\tilde O(\cdot)\) hides polylogarithmic factors.

\begin{definition}[Differential privacy]
Two datasets \(S\) and \(S'\) of the same size are neighboring (\(S \sim S'\)) if they differ by replacing one entry. A randomized algorithm \(M\) is \((\epsilon,\delta)\)-differentially private if, for every neighboring pair \(S,S'\) and every measurable event \(\mathcal E\),
\[
    \Pr\br{M(S)\in\mathcal E}
    \le
    e^\epsilon\Pr\br{M(S')\in\mathcal E}+\delta .
\]
\end{definition}
A standard way to ensure differential privacy is the Gaussian mechanism.
\begin{definition}[Gaussian Mechanism]
    For a query \(f\) with sensitivity
\(
    \Delta(f):=\sup_{S\sim S'}\|f(S)-f(S')\|_2,
\)
one releases
\[
    M(S)=f(S)+g,
    \qquad
    g\sim\mathcal N(0,\sigma^2 I).
\]
For \(\sigma \gtrsim \Delta(f)\sqrt{\log(1/\delta)}/\epsilon\), this mechanism is
\((\epsilon,\delta)\)-differentially private.
\end{definition}

\AdaptPrivOja is analyzed under the following global model, which is the same type of model used in prior stochastic private PCA analyses. We emphasize that our privacy guarantees are distribution-free; the assumptions below are used only for utility. In~\Cref{ass:tangent-model}, we introduce a local tangent-space refinement of this model, which applies after a sufficiently accurate warm-up phase and leads to sharper bounds for \TangAlg.

\begin{assumption}[\((\Sigma,\bc{\lambda_i}_{i=1}^d,M,V,K,a,\gamma)\)-model]
\label{assumption-1}
The matrices \(A_1,\ldots,A_n\in\reals^{d\times d}\) are independent, symmetric, and satisfy:
\begin{asslist}
   
    \item \label{ass:A1} \(\bE\bs{A_i}=\Sigma\), where
    \(\Sigma\succeq0\) has eigenvalues
    \(\lambda_1\ge\cdots\ge\lambda_d\ge 0\), corresponding
    eigenvectors \(v_1,\dots,v_d\), and $\lambda_k - \lambda_{k+1} > 0$.

    \item \label{ass:A2}
    \(
        \norm{A_i-\Sigma}_{(2,k)}
        =
        \sup_{P\in\bS_{d,k}}\norm{P^\top(A_i-\Sigma)}_F
        \le \lambda_1 M\; \text{ a.s.}
    \)

    \item \label{ass:A3}
    \(
        \norm{\bE\bs{(A_i-\Sigma)^2}}_2
        \le
        \lambda_1^2V .
    \)

    \item \label{ass:A4} \(\max\limits_{\|u\|_2 = 1,\|v\|_2 = 1}
    \bE \bs{\exp\br{\br{\frac{\abs{u^\top (A_i-\Sigma)v}^2}{K^2\lambda_1^2\|H_u\|_2}}^{\nicefrac{1}{2a}}}}\le 2,
  \) where \(H_u = \tfrac{1}{\lambda_1^2}\mathbb E[(A_i-\Sigma)u\,u^\top(A_i-\Sigma)^\top]\) and we denote \(\gamma^2 = \max_{\|u\|=1}\|H_u\|_2\).
    
\end{asslist}
\end{assumption}

\Cref{ass:A1,ass:A2,ass:A3} are standard for matrix concentration
(e.g., under the matrix Bernstein inequality \citep{tropp2012user})
and thus also required for the utility guarantees of Oja's algorithm
even in the non-private setting.~\Cref{ass:A4} controls the size of
the bilinear form \(u^\top (A_i-\Sigma) v\) and can be seen as a
Gaussian-like tail bound, which tells us that the magnitude of the
projection of the fluctuation \(A_i - \Sigma\) is bounded with high
probability. It is a standard assumption in related private stochastic PCA analyses~\citep{liu2022dp,dungler2025iterative}.

%% file: sections/2-2-warmup.tex
\subsection{Theoretical Warmup}

One of the most widely used algorithms for Principal Component
Analysis (PCA) is Oja's
algorithm~\citep{oja1982simplified,oja1985stochastic}. It is a
streaming algorithm that receives a sequence of matrices \(A_i\) and
at each time step updates a \(k\)-dimensional subspace \(Q_t\) by
\[
    Q_t=\mathrm{QR}\br{Q_{t-1}+\eta_t A_tQ_{t-1}}.
\] where \(\eta_t\) is the learning rate at time \(t\). The standard utility theorem for Oja's method~\citep{huang2021streaming} assumes that the matrices \(A_t\) are independent, symmetric, and have common expectation \(\Sigma\). We recall the full statement and pseudocode in \Cref{thm:main-thm-streaming,algo:ojas}. These two requirements are exactly what make a private adaptive block update delicate.

If one could release \((A_t+G_t)Q_{t-1}\), where \(G_t\) is an
independent symmetric Gaussian matrix, then the update would simply be
Oja's algorithm run on perturbed symmetric matrices \(A_t+G_t\). The
challenge, however, is not merely privacy, but adaptivity: the algorithm
must privately estimate the range of the matrix being privatized, choose
a clipping threshold, and add Gaussian noise calibrated to the resulting
sensitivity.

Adding noise to the full matrix \(A_t\) would preserve the standard analysis of Oja’s algorithm, but would require clipping at the range of \(A_t\), which can
be much larger than the range of the projected update
\(A_tQ_{t-1}\). Since privacy only needs to protect
\(A_tQ_{t-1}\in\reals^{d\times k}\), this would be unnecessarily
conservative and would destroy the adaptive gain. Instead, we estimate
the range and add Gaussian noise directly in \(\reals^{d\times k}\), at
the scale of \(A_tQ_{t-1}\). This is the right scale for privacy, but the
noise is generally not of the form \(G_tQ_{t-1}\) for any symmetric
\(G_t\). Hence the effective matrices are no longer symmetric, and the
standard theorem for Oja’s algorithm no longer applies. This is one reason prior adaptive \(k\)-PCA algorithms use deflation.~\citet{dungler2025iterative} reduce the problem to a
sequence of private one-dimensional PCA calls. 
For \(k=1\), the Oja update becomes
\[
    q_t=\mathrm{QR}\br{q_{t-1}+\eta_t A_tq_{t-1}},
\]
with \(q_{t-1}\in\mathbb R^d\). In this \(k=1\) setting, the available
utility result does not require the input matrices in the update to be
symmetric. Deflation therefore avoids the symmetry obstruction that
arises for the full \(d\times k\) block update.
However, it requires fresh batches of data to maintain
independence across directions, which loses a factor in the dependence on
\(k\).

We retain the block update through two new observations. First, for fixed \(Q\), the action of any symmetric matrix on \(Q\) lies
in the linear space
\[
    \mathcal S_Q=\bc{Y\in\reals^{d\times k}:Q^\top Y\text{ is symmetric}}.
\]
Thus the privacy noise only needs to be added inside \(\mathcal S_Q\),
rather than in all of \(\reals^{d\times k}\). This already restores a
symmetric representation: every element of \(\mathcal S_Q\) can be
written as \(G_QQ\) for some symmetric matrix \(G_Q\). However, with the
natural isotropic noise on \(\mathcal S_Q\), the representing matrix
\(G_Q\) depends on \(Q\). Along the algorithm, \(Q=Q_{t-1}\) is random
and adaptive, so the resulting effective matrices are not an i.i.d.
sequence of symmetric perturbations, and the standard theorem for Oja’s algorithm still
does not apply.

The second observation is that adding slightly more noise gives a
\(Q\)-independent symmetric representation, while preserving the same
privacy guarantee.  To state this precisely, we use the following notation.

\begin{definition}[Gaussian Orthogonal Ensemble]
A symmetric matrix \(G\in\reals^{d\times d}\) is sampled from
\(\mathrm{GOE}_d(\sigma^2)\) if \(G_{ii}\sim\cN(0,2\sigma^2)\) and
\(G_{ij}\sim\cN(0,\sigma^2)\) independently for \(i<j\).
\end{definition}

\begin{restatable}{lemma}{updatestepequiv}\label{lem:update-step-equivalence}
    Fix \(Q\in\reals^{d\times k}\) with \(Q^\top Q=I_k\). Let \(Z\in\reals^{d\times k}\) have i.i.d.\ \(\cN(0,\sigma^2)\) entries, let \(N_{\mathrm{sym}} \sim \mathrm{GOE}_k(\sigma^2)\), and let \(G\sim \mathrm{GOE}_d(\sigma^2)\). Then
    \[
    QN_{\mathrm{sym}}+(I-QQ^\top)Z
    \ \stackrel{d}{=}\ 
    GQ.
    \]
\end{restatable}

To privatize the projected update \(AQ\), we need a uniform sensitivity
bound. Since \(AQ\) is not deterministically bounded under the stochastic
model, we enforce such a bound by clipping in Frobenius norm. For
\(R>0\), define
\[
    \operatorname{clip}_R(Y)=
    \begin{cases}
        Y, & \norm{Y}_F\le R,\\
        RY/\norm{Y}_F, & \norm{Y}_F>R.
    \end{cases}
\]
Thus, at an iteration with current iterate \(Q_{t-1}\), we use the
private clipped update
\[
    Q_t
    =
    \mathrm{QR}\br{
        Q_{t-1}
        +
        \eta_t\br{
            \operatorname{clip}_R(A_tQ_{t-1})
            +
            Q_{t-1}N_t
            +
            (I-Q_{t-1}Q_{t-1}^\top)Z_t
        }
    },
\]
where \(N_t\sim\mathrm{GOE}_k(\sigma^2)\) and
\(Z_t\in\reals^{d\times k}\) has i.i.d.\ \(\cN(0,\sigma^2)\) entries.

For any \(R\) for which clipping is rare, the update agrees on the
no-clipping event with non-private Oja’s algorithm applied to
\(A_t+G_{t-1}\), where \(G_{t-1}\sim\mathrm{GOE}_d(\sigma^2)\). Thus we
can privatize the full \(d\times k\) block update while preserving the
structure required by Oja's analysis. For a rigorous privacy proof of~\Cref{algo:priv-ojas-batches} see~\Cref{thm:privacy-of-ojas-batches} in~\Cref{app:proofs}.
\begin{algorithm}[t]
\caption{Private Oja's}
\label{algo:priv-ojas-batches}
\textbf{Input:} \(S=\bc{A_i}_{i=1}^n\), \(k\), batch count \(T\), privacy parameters \((\epsilon,\delta)\), learning rates \(\bc{\eta_i}_{i=1}^T\), clipping radius \(R\).
\begin{algorithmic}[1]
    \State Set \(B\gets\lfloor n/T\rfloor,~\sigma\gets2R\sqrt{2\log(1.25/\delta)}/(B\epsilon)\).
    \State Choose \(Q_0'\in\reals^{d\times k}\) uniformly at random
    and set \(Q_0\gets\mathrm{QR}(Q_0')\).  \For{\(t=1,\ldots,T\)}
    \State Set \(\Pi_{Q_{t-1}}\gets
    I- Q_{t-1}Q_{t-1}^\top\).  
    \State Draw
    \(Z_t\in\reals^{d\times k}\) with i.i.d. entries
    \(\cN(0,\sigma^2)\), and draw \(N_{\mathrm{sym},t}\sim \mathrm{GOE}_k(\sigma^2)\). 
    \State \(Q_t\gets\mathrm{QR}\left(Q_{t-1}+\eta_t\left[\frac1B\sum_{i=B(t-1)+1}^{tB}\operatorname{clip}_R(A_iQ_{t-1})+Q_{t-1}N_{\mathrm{sym},t}+\Pi_{Q_{t-1}}Z_t\right]\right)\).
    \EndFor
    \State \Return \(Q_T\).
\end{algorithmic}
\end{algorithm}
The next theorem gives a
fixed-radius guarantee, choosing \(R\) under \Cref{assumption-1} so that
clipping is unlikely and the dependence on \(k\) is correct.

\begin{restatable}[Utility of \Cref{algo:priv-ojas-batches}]{theorem}{dpojasbatchesutilitythm}
\label{thm:utility-priv-ojas-batches}
Assume~\(A_1,\ldots,A_n\) satisfy \Cref{assumption-1} and write \(\kappa = \frac{\lambda_1}{\lambda_k - \lambda_{k+1}}\). Set
\[
    R=C\lambda_1\sqrt{dk}\bigl(K\gamma\log^a(ndk/\zeta)+1\bigr),
    \qquad
    T=\lceil C_T \kappa \log(n) \rceil ,
\]
where \(C_T\) is a sufficiently large absolute constant. If
\[
n\gtrsim
\max\left\{
    \frac{k^2\lambda_1^2V}{\zeta^2(\lambda_k - \lambda_{k+1})^2},
    \kappa^{3/2} \frac{kM}{\zeta},
    \kappa^{3/2} \frac{(K\gamma+1)k^{3/2}d}{\epsilon\zeta}
\right\},
\]
then there exist learning rates \(\bc{\eta_t}\) so that with probability at least \(1-\zeta\), \Cref{algo:priv-ojas-batches} outputs \(Q_T\) satisfying
\[
    \norm{Q_TQ_T^\top-V_k V_k^\top}_F
    \le
    \tilde O\left(\kappa 
        \frac{\sqrt{kV}}{\sqrt n}
        +
        \kappa^{3/2} \frac{\sqrt{k}M}{n}
        +
        \kappa^{3/2} \frac{(K\gamma+1)kd}{\epsilon n}
    \right).
\]
\end{restatable}

The first two terms are the stochastic error of Oja's algorithm, and the third is the privacy cost. In particular, the privacy term scales as
\(dk/(\epsilon n)\), reflecting the block update over all \(k\)
directions.

\begin{example}[Gaussian covariance and the \(k\)-dependence]
\label{example:gaussian-data}
Let \(A_i=x_ix_i^\top\) with \(x_i\sim\cN(0,\Sigma)\). Then, with high
probability, \Cref{assumption-1} holds with \(M=\tilde
O(d)\), \(V=O(d)\), \(K=O(1)\), \(a=1\), and \(\gamma=O(1)\).
Thus, applying \Cref{thm:utility-priv-ojas-batches} on this high-probability
event gives
\[
    \norm{Q_TQ_T^\top-V_k V_k^\top}_F
    \le
    \softO{\kappa \br{\frac{\sqrt{kd}}{\sqrt n}+\sqrt{\kappa}\frac{dk}{\epsilon n}}}.
\]
\end{example}

However,~\Cref{algo:priv-ojas-batches} is not adaptive. Since
\(R\) is fixed before the algorithm runs and clipping is around zero, the
radius must cover both \((A_i-\Sigma)Q\) and \(\Sigma Q\). Consequently,
the privacy noise need not vanish even when the stochastic fluctuation
does. Thus the block update has the right \(k\)-dependence, but its
privacy cost is not adaptive to the randomness in the data, as illustrated
in the example below.

\begin{example}[Spiked covariance and adaptivity]
\label{spiked-cov-example}
Consider the rank-one spiked model \(x_i=s_i+n_i\), where
\(s_i\sim\mathrm{Unif}\{v,-v\}\), \(n_i\sim\cN(0,\sigma^2I_d)\), and
\(v\in\reals^d\) is unit norm. Let \(A_i=x_ix_i^\top\), so
\(
    \Sigma=\bE\bs{A_i}=vv^\top+\sigma^2I_d .
\)
As \(\sigma\to0\), the samples become nearly deterministic:
\(A_i\to vv^\top=\Sigma\), and the stochastic fluctuation scale
\(\gamma\) tends to zero. However, the radius used in
\Cref{thm:utility-priv-ojas-batches} is
\(
    R=C\lambda_1\sqrt{dk}\bigl(K\gamma\log^a(ndk/\zeta)+1\bigr).
\)
Thus \(R\) remains of order \(\lambda_1\sqrt{dk}\) even when
\(\gamma\to0\), and the privacy noise does not vanish in the nearly
deterministic regime. The adaptive algorithm in~\Cref{sec:adaptive-priv-ojas} removes
this obstruction by privately estimating a center for \(A_iQ\) and
clipping around that center.
\end{example}

%% file: sections/adaptive_private_ojas.tex
\section{Adaptive Private Oja's}\label{sec:adaptive-priv-ojas}

The warmup in the previous section gives a private block version of
Oja's algorithm with the correct dependence on \(k\), but its clipping
radius is fixed in advance and centered at zero. We now make the block
update adaptive. At each iteration \(t\), \Cref{algo:priv-ojas-adaptive}
estimates a center for the projected samples
\(
    A_iQ_{t-1}\in\reals^{d\times k}
\) and a range parameter
controlling their fluctuations around that center
(\PrivRange,~\Cref{algo:priv-range-max}). It then clips around the estimated center
and computes a truncated batch mean
(\TruncatedMean,~\Cref{algo:trunc-mean}) and finally privatizes the mean by adding Gaussian noise. The sensitivity, and hence the Gaussian
noise scale, is calibrated to the empirical fluctuation radius rather
than to a worst-case radius fixed before the algorithm begins. Conditional on \(Q_{t-1}\), the population center of the projected
samples is \(\Sigma Q_{t-1}\), so the relevant fluctuation is
\[
    A_iQ_{t-1}-\Sigma Q_{t-1}
    =
    (A_i-\Sigma)Q_{t-1}.
\]
This is the quantity controlled by the fluctuation parameter
\(\gamma\). Consequently, when the stochastic fluctuations are small,
the clipping radius and privacy noise become small as well. This is the
adaptivity captured in~\Cref{thm:utility-priv-ojas-adaptive}.

\begin{algorithm}[ht]\caption{Adaptive Private Oja's (\AdaptPrivOja)}\label{algo:priv-ojas-adaptive}
\textbf{Input:} $S = \{A_1, \dots, A_{n} \}$, $k \in [d]$ , batch count $T$, warm-up length $T_0$, base clipping radius $R$, privacy parameters $(\epsilon, \delta)$, learning rates $\{ \eta_t \}_{t=1}^{T}$, and failure probability $\zeta$
    \begin{algorithmic}[1]
        \State Set \(B\gets\lfloor n/T\rfloor,~\sigma\gets2R\sqrt{2\log(1.25/\delta)}/(B\epsilon)\)
        \State Choose $Q_0' \in \mathbb{R}^{d \times k}$ uniformly at random,~$Q_0 \gets \text{QR}[Q_0']$
        \For{$t = 1,2,\dots,T_0$} \Comment{Phase I: non-adaptive warm-up}
            \State $\bar G_t \gets \frac1B\sum_{i=B(t-1)+1}^{tB}\clip_R(A_iQ_{t-1})$ \textbf{and} $\sigma_t \gets \sigma$
            \State $Z_t \in \mathbb{R}^{d \times k}$ with coordinates $\overset{\text{i.i.d}}{\sim} \cN(0,\sigma_t^2), \quad N_{\mathrm{sym}, t} \sim GOE_k(\sigma_t^2)$
            \State $Q_t \gets \text{QR}\left[Q_{t-1} + \eta_t \left(\mathsf P_{\mathcal S_{Q_{t-1}}}(\bar{G}_t) + Q_{t-1} N_{\mathrm{sym}, t} + (I - Q_{t-1} Q_{t-1}^\top ) Z_t\right)\right]$
        \EndFor
        \For{$t = T_0+1,\dots,T$} \Comment{Phase II: adaptive}
            \State Estimate Range $\widehat\Lambda_t \gets$  \PrivRange$\left(\{ A_i Q_{t-1} \}_{i=B \cdot (t-1) +1}^{B\cdot (t-1)+B/2}, (\epsilon, \delta), \zeta/2T\right)$
            \State Set Truncation Threshold $R_t \gets 3K\sqrt{\widehat\Lambda_t}\,\log^a(Bdk/(2\zeta))$
            \State $\bar{G}_t \gets
            \TruncatedMean \left(\{A_i Q_{t-1} \}_{i=B\cdot (t-1)+B/2 + 1}^{tB} , R_t, \widehat\Lambda_t, (\epsilon/2, \delta/2), \zeta/2T\right)$
            \State Set privacy noise multiplier
            \(
                \sigma_t \gets
                \frac{8R_t\sqrt{dk}}{B\epsilon}
                \sqrt{2\log(2.5/\delta)}
            \)
            \State $Z_t \in \mathbb{R}^{d \times k}$ with coordinates $\overset{\text{i.i.d}}{\sim} \cN(0,\sigma_t^2), \quad N_{\mathrm{sym}, t} \sim GOE_k(\sigma_t^2)$
            \State $Q_t \gets \text{QR}\left[Q_{t-1} + \eta_t \left(\mathsf P_{\mathcal S_{Q_{t-1}}}(\bar{G}_t) + Q_{t-1} N_{\mathrm{sym}, t} + (I - Q_{t-1} Q_{t-1}^\top ) Z_t\right)\right]$\label{step:priv-ojas-adaptive-update-step}
        \EndFor
        \State \Return $Q_T$
    \end{algorithmic}
\end{algorithm}

\begin{restatable}[Privacy of~\Cref{algo:priv-ojas-adaptive}]{lemma}{dpojasadaptiveprivlem}\label{lem:privacy-of-ojas-adaptive} If $0 < \varepsilon \leq 1$ then~\Cref{algo:priv-ojas-adaptive} is $(\varepsilon, \delta)$-DP. 
\end{restatable}
\begin{remark}
The restriction on \(\varepsilon\) can be removed by replacing the
standard Gaussian calibration with the analytic Gaussian mechanism. This calibration is less convenient to state in closed form, but is readily implemented in our experiments.
\end{remark}

\begin{restatable}[Utility of~\Cref{algo:priv-ojas-adaptive}]{theorem}{dpojasadaptiveutilitythm}
\label{thm:utility-priv-ojas-adaptive}
Given matrices $A_1, \dots, A_n$ fulfilling~\Cref{assumption-1} write
\(\kappa = \frac{\lambda_1}{\lambda_k - \lambda_{k+1}}\) and set the base clipping radius,
the batch count and the warm-up length to
\[
    R=C\lambda_1\sqrt{dk}\bigl(K\gamma\log^a(ndk/\zeta)+1\bigr),
    \qquad
    T=\lceil C_T \kappa \log(n) \rceil,
    \qquad
    T_0=\lceil T/2\rceil,
\]
where \(C_T\) is a sufficiently large absolute constant. If
\[
n\gtrsim
\max\left\{
    \frac{k^2\lambda_1^2V}{\zeta^2(\lambda_k - \lambda_{k+1})^2},
    \kappa^{3/2}\cdot\frac{kM}{\zeta},
    \kappa^{3/2}\cdot\frac{(K\gamma+1)k^{3/2}d}{\epsilon\zeta},
    \frac{dk}{\epsilon},
    \frac{K^2d}{\epsilon}
\right\},
\]
then there exist learning rates \(\bc{\eta_t}\) such that, with probability at least $1 - \zeta$,
\Cref{algo:priv-ojas-adaptive} outputs $Q_T$ with
\[
    \| Q_T Q_T^\top - V_k V_k^\top \|_F
    \leq
    \tilde{O}\left( \kappa
        \frac{\sqrt{kV}}{\sqrt{n}}
        +
        \kappa^{3/2} \frac{\sqrt{k}M}{n}
        +
        \kappa^{3/2}\frac{ K d k \gamma}
        {\varepsilon n}
    \right).
\]
\end{restatable}
\begin{remark}
The choice $T_0=\lceil T/2\rceil$ is sufficient but far from necessary: We fix it at $T/2$ because that choice requires no knowledge of
$V$, $M$ or $\gamma$.
\end{remark}

\begin{restatable}[Rank-\(k\) spiked covariance]{corollary}{corrankkadaptiveojas}\label{cor:rank-k-adapitve-ojas}
Let \(V_k \in\mathbb R^{d\times k}\) have orthonormal columns, let
\(\Theta=\mathrm{diag}(\theta_1,\dots,\theta_k)\) with
\(\theta_1\ge\cdots\ge\theta_k>0\), and let
\(
    X_i=V_k\Theta^{1/2}+\sigma Z_i, \;
    (Z_i)_{ab}\overset{\mathrm{i.i.d.}}{\sim}\mathcal N(0,1).
\)
Set $A_i = X_i X_i^\top$, then \( \Sigma=\mathbb E[A_i]=V_k\Theta V_k^\top+k\sigma^2I_d\). Running~\Cref{algo:priv-ojas-adaptive}  with input
\(A_1,\ldots,A_n\) outputs \(Q_T\) with high probability satisfying
\[
 \| Q_TQ_T^\top - V_k V_k^\top \|_F \leq \tilde{O}\left( \frac{\sigma\sqrt{(\theta_1 + k\sigma^2)kd }}{\theta_k \sqrt{n}}+ \sigma \cdot \sqrt{\frac{\theta_1 + k \sigma^2}{\theta_k}}\frac{Kdk\,\sqrt{\theta_1+\sigma^2k}}
    {\varepsilon\theta_k n}\right)
\]
\end{restatable}
\begin{remark}
    The privacy term in~\Cref{cor:rank-k-adapitve-ojas} scales with the fluctuation level \(\sigma\) and vanishes as \(\sigma\to0\).
\end{remark}

To estimate fluctuations around the mean
(\Cref{algo:priv-range-max}), we build on the private range-estimation
idea of~\citet{liu2022dp} (\Cref{algo:priv-top-eigenval-estimation-liu}).
The new difficulty is that our updates are matrix-valued:
\(A_iQ_{t-1}\in\mathbb R^{d\times k}\), rather than vector-valued as in
the top-eigenvector setting. A naive extension would estimate a range
separately for each column and compose privacy over \(k\) runs. Instead,
\Cref{algo:priv-range-max} estimates one range parameter for the whole
batch by taking blockwise maxima over the \(k\) column fluctuations and
running the private histogram learner only once per update step. This lets us update the
entire \(k\)-dimensional subspace at once, rather than using a
deflation-based sequence of private one-dimensional problems as
in~\citet{dungler2025iterative}.

\subsection{Experiments}\label{subsec:experiments}

We compare \AdaptPrivOja against the stochastic DP-PCA methods: \kdppca and
\kdpojas~\citep{dungler2025iterative}, and against stochastic adaptations of the
DP-Gauss algorithms of~\citet{dwork2014analyze} and the noisy power method
of~\citet{hardt2014noisy}. The latter methods were designed for deterministic
bounded-data settings, so we adapt them to our stochastic model by clipping.

The algorithms of~\citet{dwork2014analyze} assume a data matrix
\(X\in\mathbb R^{n\times d}\) with rows bounded in Euclidean norm by one and
estimate the top eigenspace of \(X^\top X\). The noisy power method was originally
analyzed under single-entry matrix changes of size at most one. Later analyses
\citep{florina2016improved,nicolas2024differentially} showed privacy under
perturbations \(A'=A+C\) satisfying
\(\sqrt{\sum_i\|C_{i,:}\|_1^2}\le1\). In contrast, our setting observes
independent matrices \(A_i\),
with no deterministic norm bound, and the goal is to estimate the top eigenspace
of \(\mathbb E[A_i]=\Sigma\).

To use deterministic baselines, we clip the samples. A global
normalization \(\tilde x_i=x_i/\max_j\|x_j\|_2\) is not private, since
one sample can change the normalization of all
samples~\citep{hu2024provable}. Normalizing each sample individually,
\(\tilde x_i=x_i/\|x_i\|_2\), changes the target because generally
\(\mathbb E[xx^\top/\|x\|_2^2]\neq\Sigma\). Instead, as
in~\citep{dungler2025iterative}, we clip at a deterministic threshold
\(R\) chosen so that clipping is inactive with high probability. For
the spiked covariance model, we take
\(R=C\sqrt{\lambda_1}+\sigma\sqrt{d\log(n/\vartheta)}\), which ensures
\(\|x_i\|_2\le R\) for all \(i\) with probability at least
\(1-\vartheta\), up to constants. The DP-Gauss privacy noise is then
scaled to this clipping radius. 

With this clipping, we adapt Algorithms 1 and 2
of~\citet{dwork2014analyze}, denoted \texttt{DP-Gauss-1} and
\texttt{DP-Gauss-2}. \texttt{DP-Gauss-1} clips each sample, adds
Gaussian noise to the empirical second-moment matrix, and runs
non-private PCA on the noisy matrix. \texttt{DP-Gauss-2} privately
estimates the eigengap of the clipped covariance matrix, computes the
non-private top-\(k\) eigenspace of the clipped data, and perturbs the
eigenvectors using noise calibrated to the private eigengap estimate. 

We also adapt the noisy power method. Replacing one sample by \(a\)
adds the rank-one perturbation \(C=aa^\top\), for which
\(\sqrt{\sum_i\|C_{i,:}\|_1^2}=\|a\|_2\|a\|_1\). Thus, to satisfy the
adjacency condition
of~\citet{florina2016improved,nicolas2024differentially}, we clip so
that \(\|a\|_2\le R\) (same $R$ as for \texttt{DP-Gauss}) and
\(\|a\|_1\le R'\), and scale the privacy noise accordingly. In the
spiked covariance model, we use \(R'=\sigma
d+\sqrt{\lambda_1d}+\sigma\sqrt{d\log(n/\vartheta)}\), so that
\(\|x_i\|_1\le R'\) with probability at least \(1-\vartheta\), up to
constants. We call this baseline \texttt{DP-Power-Method}.

\begin{figure}[t]
  \centering

  \includegraphics[width=0.9\textwidth]{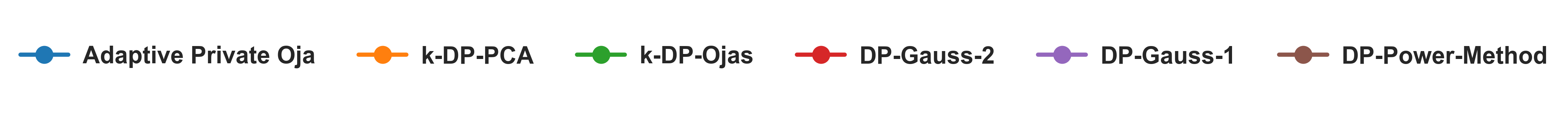}

  \vspace{0.75em}

  \begin{subfigure}[b]{\textwidth}
    \centering
    \includegraphics[width=\linewidth]{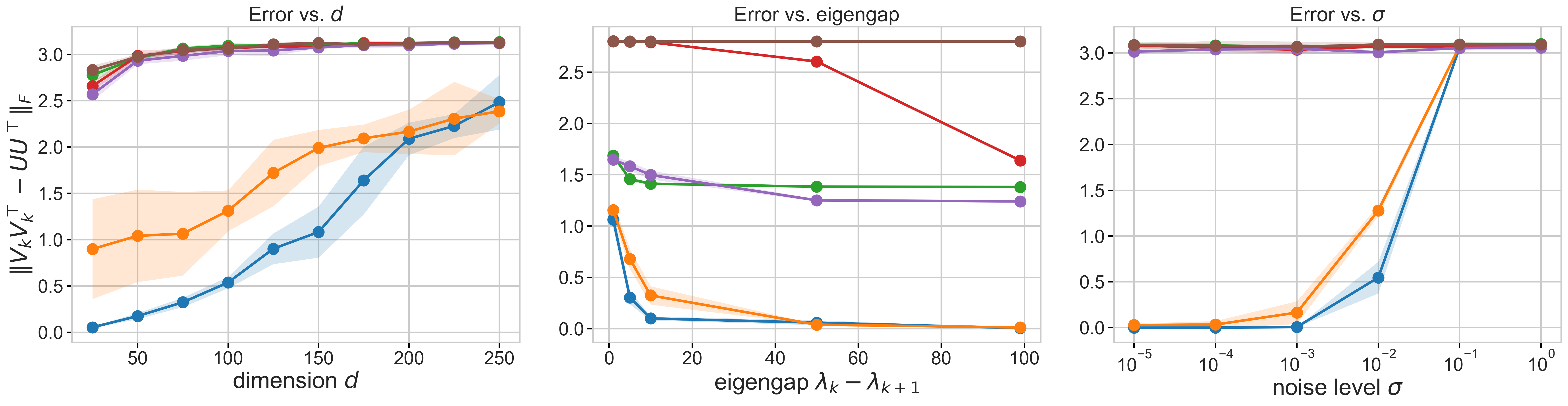}
    \phantomsubcaption
    \label{fig:adadpojas-sweep}
  \end{subfigure}

  \begin{subfigure}[b]{\textwidth}
    \centering
    \includegraphics[width=\linewidth]{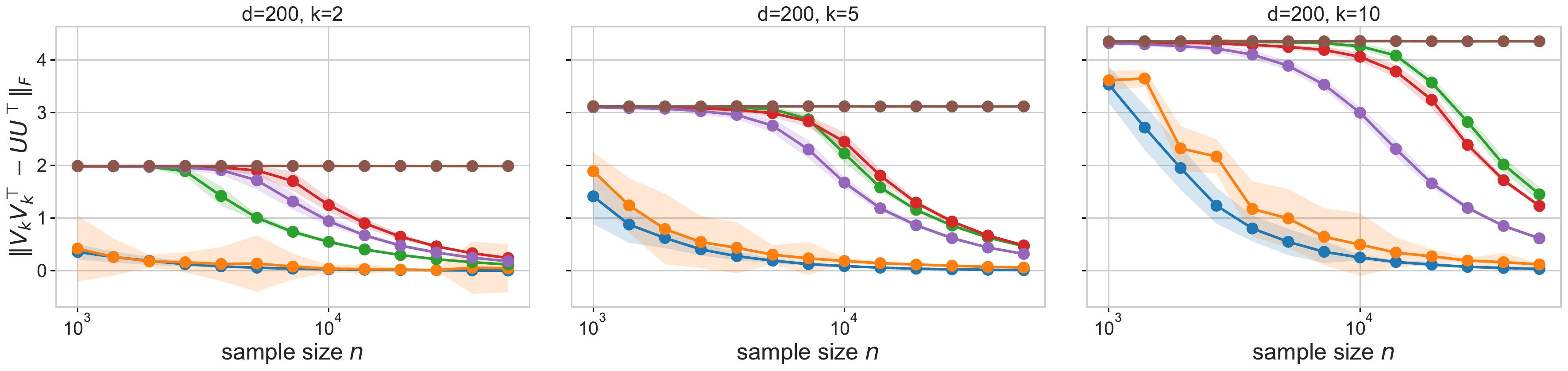}
    \phantomsubcaption
    \label{fig:adadpojas-k-and-d}
  \end{subfigure}
  
  \ifarxiv
    \vspace{0mm}
  \else
    \vspace{-2mm}
  \fi

  \caption{\small Comparison of \AdaptPrivOja with baselines on the spiked covariance model. We plot the mean over 10 trials, with shaded regions representing 95\% confidence intervals. We set $\epsilon=1$, $\delta = 0.01$. For the top row figures we set $k=5$.
  }
  \label{fig:adadpojas-spiked-comparison}
  \ifarxiv
    \vspace{0mm}
  \else
    \vspace{-5 mm}
  \fi
\end{figure}

\noindent\textbf{Experimental Results on Spiked Covariance Data} In~\Cref{fig:adadpojas-sweep} we vary the ambient dimension, eigengap, and
noise level while keeping the other parameters fixed. The plots show that
\AdaptPrivOja consistently attains lower subspace error. In~\Cref{fig:adadpojas-k-and-d} we vary the sample size across several choices of \(k\) and \(d\), showing that the advantage persists across different ranks and dimensions and that \AdaptPrivOja benefits from the predicted block \(k\)-dependence. Across
all plots, the deterministic bounded-data baselines are less competitive,
reflecting the cost of clipping and calibrating noise to worst-case
bounds rather than to the stochastic fluctuations of the data. Additional experiments and implementation details are provided in
\Cref{app:experiments}.

%% file: sections/main_tangent_results_extended.tex
\section{Fine-tuning in tangent space for optimal spectral dependence}
\label{sec:tangent-adaptive}

\AdaptPrivOja has optimal dependence on \(d,k,n\), and \(\varepsilon\) for
general Gaussian inputs, because its privacy noise adapts to the stochastic
fluctuation of the projected update. In the Gaussian spiked covariance model
of~\citet{cai2024optimal}, however, the data have additional structure. Namely,
\(A_i=x_ix_i^\top\) with
\(
    x_i\sim\cN(0,\Sigma),
    \text{ and }
    \Sigma=U\Theta U^\top+\sigma^2 I_d,
\)
where \(U\in\mathbb R^{d\times k}\) has orthonormal columns. Equivalently, all
directions outside the signal subspace have the same eigenvalue,
\(
    \lambda_{k+1}=\cdots=\lambda_d=\sigma^2 .
\)
This flat-tail structure permits a sharper dependence on the spectrum of
\(\Sigma\). The minimax lower bound~(\Cref{spiked-cov-lower-bound}) scales as
\(
    \frac{\sqrt{\lambda_1\lambda_{k+1}}}
         {\lambda_k-\lambda_{k+1}}
    \left(
        \sqrt{\frac{dk}{n}}
        +
        \frac{dk}{\varepsilon n}
    \right),
\)
up to logarithmic factors. By contrast, applying~\Cref{thm:utility-priv-ojas-adaptive} directly to
\(A_i=x_ix_i^\top\) yields
\[
    \norm{Q_TQ_T^\top-V_kV_k^\top}_F
    \le
    \widetilde O\left(
        \kappa\sqrt{\frac{dk}{n}}
        +
        \kappa^{3/2}\,\frac{Kdk}{\varepsilon n}
    \right).
\]
as shown in~\Cref{cor:gaussian-spiked-adaptive-ojas}. Thus \AdaptPrivOja has the correct dependence on the statistical and privacy
parameters, but its spectral factors, \(\kappa\) on the statistical term and \(\kappa^{3/2}\) on
the privacy term, are larger than the single factor
\(\sqrt{\lambda_1\lambda_{k+1}}/(\lambda_k-\lambda_{k+1})\) available in the flat-tail Gaussian
family.

The reason is geometric. Once an iterate \(Q\) is already close to the target
subspace, not every component of the update \(A_iQ\) contributes equally to
improving the subspace estimate. The component lying inside \(\operatorname{span}(Q)\) changes the
coordinates of the basis \(Q\), but not the subspace it spans; the
subspace is changed by the orthogonal, or tangent, component. Writing
\(
    \Pi_Q := I-QQ^\top,
\)
this tangent component is
\(
    \Pi_Q A_iQ .
\)
Accordingly, our second algorithm uses \AdaptPrivOja only as a warm-up phase.
After obtaining a private estimate \(Q_0\) within constant subspace error of the
true top-\(k\) eigenspace, it runs a second-stage refinement on fresh samples
using only the projected tangent updates
\[
    Y_i=\Pi_{t-1}A_iQ_{t-1},
    \qquad
    \Pi_{t-1}=I-Q_{t-1}Q_{t-1}^\top .
\]
This removes the component of the update that merely rotates the current basis
and focuses the private mean estimation on the residual correction that moves
the subspace.
\begin{algorithm}[t]
\caption{Tangent Adaptive Private Ojas (\TangAlg)}
\label{alg:tangent-adaptive}
\textbf{Input:} samples \(S=\bc{A_i}_{i=1}^n\), target rank \(k\), batch count $T$, privacy parameters \((\epsilon,\delta)\), failure probability \(\zeta\), learning rates $\{ \eta_t \}_{t=1}^{T}$, and a private starting point $Q_0 \in \mathbb{R}^{d \times k}$.
\begin{algorithmic}[1]
    \State Set \(B\gets\lfloor n/T\rfloor \)
    \For{\(t=1,\ldots,T \)}
        \State Set \(\Pi_{t-1}=I-Q_{t-1}Q_{t-1}^\top\)
        \State Project updates to tangent space \(
            Y_i \gets \Pi_{t-1}A_iQ_{t-1}
        \) for \( i \in [ B (t-1) +1, B t]\)
        \State Estimate Range \(\widehat\Lambda_t\gets\PrivRange(\bc{Y_i}_{i=B \cdot (t-1) +1}^{B\cdot (t-1)+B/2},\varepsilon,\delta,\zeta/(8T))\).
        \State Set Truncation Threshold 
        \( r_t = C_RK\sqrt{\widehat\Lambda_t }\log^a\left(\frac{BdkT}{\zeta}\right)\)
        \State $\bar{Y}_t \gets \TruncatedMean\!\left(
    \bc{Y_i}_{i=B(t-1)+B/2+1}^{Bt},
    r_t,
    \widehat\Lambda_t,
    (\epsilon/2, \delta/2),\zeta/(8T)
\right)$
        \State Set privacy noise multiplier
        \(
            \sigma_t
            \gets
            \frac{8r_t\sqrt{dk}}{B\epsilon}
            \sqrt{2\log(2.5/\delta)}
        \) 
        \State Draw \(Z_t\in\reals^{d\times k}\) with i.i.d. entries \(\cN(0,\sigma_t^2)\) 
        \State \(Q_t\gets\mathrm{QR}\left(Q_{t-1}+\eta_t\Pi_{t-1}(\bar Y_t+Z_t)\right)\)
    \EndFor
    \State \Return \(Q_T\)
\end{algorithmic}
\end{algorithm}

The benefit of the tangent update can be seen through the scale of the residual
fluctuations. For \(Q\in\bS_{d,k}\), define
\[
    e(Q)=\norm{QQ^\top-V_k V_k^\top}_F,
\]
where \(V_kV_k^\top\) is the projector onto the target eigenspace. The
centered tangent fluctuation is
\[
    Z_Q(A):=\Pi_Q(A-\Sigma)Q .
\]
We measure its variance by
\begin{equation*}
\label{eq:def-vq}
    v_Q^2
    :=
    \max\left\{
    \norm{\bE\bs{Z_Q(A)Z_Q(A)^\top}}_2,
    \norm{\bE\bs{Z_Q(A)^\top Z_Q(A)}}_2
    \right\}.
\end{equation*}
This parameter controls concentration of the tangent minibatch mean. We also
define the one-column tangent scale
\begin{equation*}
\label{eq:def-gq}
    g_Q^2
    :=
    \max_{r\in[k]}
    \norm{
    \Pi_Q\bE\bs{(A-\Sigma)q_rq_r^\top(A-\Sigma)^\top}\Pi_Q
    }_2 ,
\end{equation*}
where \(q_1,\ldots,q_k\) are the columns of \(Q\). This is the scale used for
private range estimation and truncation. These quantities explain the improved spectral dependence. For Gaussian
covariance data, the fourth-moment identity implies that, in a local basin
around \(V_k V_k^\top\),
\[
    v_Q
    \lesssim
    \sqrt{\lambda_1(\operatorname{tr}_{>k}(\Sigma)+k\lambda_{k+1})}
    +
    \lambda_1\sqrt{k}\,e(Q),
\]
and
\[
    g_Q
    \lesssim
    \sqrt{\lambda_1\lambda_{k+1}}
    +
    \lambda_1 e(Q).
\]
Thus, once \(Q\) is close to the target subspace, the tangent fluctuation scale
is governed by the interaction between signal directions and noise directions.
In the flat-tail spiked covariance model,
\(\operatorname{tr}_{>k}(\Sigma)=(d-k)\lambda_{k+1}\), so the leading scales
become
\(
    v_Q \approx \sqrt{d\lambda_1\lambda_{k+1}},
    \text{ and }
    g_Q \approx \sqrt{\lambda_1\lambda_{k+1}} .
\)
Substituting these scales into the tangent refinement analysis gives the
spectral factor
\[
    \frac{\sqrt{\lambda_1\lambda_{k+1}}}
         {\lambda_k-\lambda_{k+1}},
\]
rather than \(\lambda_1/(\lambda_k-\lambda_{k+1})\). Intuitively, the full
projected update \(A_iQ\) still contains large fluctuations internal to the
signal subspace, while the tangent update removes those directions and retains
only the signal-noise interaction that actually moves the subspace.

We now state the local model used to analyze this refinement phase. Unlike the
global model for \AdaptPrivOja, this model only needs to hold after the warm-up
phase has entered a constant-accuracy neighborhood of the target subspace.

\begin{assumption}[Tangent refinement model]
\label{ass:tangent-model}
There exist constants \(c_0\in(0,1)\), \(M_\perp\ge0\), \(v_0,v_1\ge0\), \(g_0,g_1\ge0\), \(K\ge1\), and \(a\ge1/2\) such that the following hold for every \(Q\in\bS_{d,k}\) satisfying
\(
    \norm{QQ^\top -V_k V_k^\top }_2\le c_0.
\)
\begin{asslist}

    \item \label{ass:T2}
    \(
        \norm{\Pi_Q(A_i-\Sigma)Q}_2\le M_\perp
    \), 
    
    \item  \label{ass:T3}
    \(
        v_Q\le v_0+v_1e(Q)
    \), 
    \item \label{ass:T4} 
    \(
        g_Q\le g_0+g_1e(Q)
    \)

    \item \label{ass:T5}
    For every column \(q_r\) of \(Q\), every unit vector \(u\perp\operatorname{span}(Q)\), and every \(\vartheta\in(0,1)\), \[
        \Pr\br{
        \abs{u^\top(A_i-\Sigma)q_r}
        >
        K g_Q \log^a(1/\vartheta)
        }
        \le \vartheta .
    \]
\end{asslist}
\end{assumption}

\begin{remark}
\Cref{ass:tangent-model} is a localized version of the global model used for
\AdaptPrivOja. The global assumptions imply the boundedness and variance parts
of this model with \(M_\perp\le\lambda_1 M\), \(v_0=\lambda_1\sqrt V\), and \(v_1=0\),
but only give the crude tail scale \(\lambda_1\gamma\). In structured settings,
such as the flat-tail spiked covariance model, the sharper \(g_Q\)-scale tail
condition can be verified directly.
\end{remark}

\begin{restatable}[Utility of \TangAlg]
{theorem}{tangentgeneralutility}
\label{thm:tangent-general-utility}
Suppose \(A_1,\ldots,A_n\) satisfy~\Cref{assumption-1}. Suppose further that,
together with the starting point \(Q_0\), they satisfy
\Cref{ass:tangent-model}, and that
\(
    \norm{Q_0Q_0^\top-V_kV_k^\top}_F\le c_0/2
\)
with probability at least \(1-\zeta/8\). Set
\[
    T=\left\lceil
        C_T \kappa_\bot \log\!\left(
        \frac{ndk}{\zeta\delta\epsilon}
        \right)
    \right\rceil,
    \qquad \kappa_\bot := \frac{\lambda_1 - \lambda_d}{\lambda_k - \lambda_{k+1}},
\]
where \(C_T\) is a sufficiently large absolute constant. If
\[
    n
    \ge
    \softOm{
    \max\left\{
        \frac{K^2d+Kdk}{\epsilon},
        \kappa_\bot\frac{k(v_0^2+v_1^2)}{(\lambda_k - \lambda_{k+1})^2},
        \kappa_\bot \frac{\sqrt{k}M_\perp}{(\lambda_k - \lambda_{k+1})},
        \kappa_\bot \frac{K(g_0+g_1)dk}{\epsilon(\lambda_k - \lambda_{k+1})}
    \right\}
    },
\]
then there exists a choice of learning rates \(\{\eta_t\}_{t=1}^T\) such that,
with probability at least \(1-\zeta\), \Cref{alg:tangent-adaptive} outputs
\(Q_T\) satisfying
\[
\norm{Q_TQ_T^\top-V_kV_k^\top}_F
\le
\operatorname{Opt}_T
+
\softO{
    \sqrt{\kappa_\bot} \frac{\sqrt{k}v_0}{(\lambda_k - \lambda_{k+1})\sqrt n}
    +
    \kappa_\bot \frac{\sqrt{k}M_\perp}{(\lambda_k - \lambda_{k+1}) n}
    +
    \kappa_\bot \frac{K g_0dk}{(\lambda_k - \lambda_{k+1})\epsilon n}
},
\]
where
\(
    \operatorname{Opt}_T
    :=
    \exp\!\left(
        -c(\lambda_k - \lambda_{k+1})\sum_{t=1}^T\eta_t
    \right)
    \norm{Q_0Q_0^\top-V_kV_k^\top}_F .
\)
\end{restatable}

The leading statistical and privacy terms are governed by the tangent
scales \(v_0\) and \(g_0\). Thus, when these are smaller than their
global counterparts, the refinement phase improves the spectral
dependence without changing the dependence on \(d,k,n\), or
\(\varepsilon\).

\begin{corollary}[Gaussian spiked covariance,~\Cref{alg:tangent-adaptive}]
\label{cor:flat-tail}
Let \(V_k\in\mathbb R^{d\times k}\) have orthonormal columns, let
\(\Theta=\mathrm{diag}(\theta_1,\dots,\theta_k)\) with
\(\theta_1\ge\cdots\ge\theta_k>0\) and \(\theta_1\le C_0\theta_k\), and let
\(
    x_i\sim\cN(0,\Sigma), \;
    \Sigma=V_k\Theta V_k^\top+\sigma^2I_d .
\)
Set \(A_i=x_ix_i^\top\), so that \(\lambda_i=\theta_i+\sigma^2\) for \(i\le k\)
and \(\lambda_{k+1}=\cdots=\lambda_d=\sigma^2\).
Running~\Cref{alg:tangent-adaptive} with input \(A_1,\ldots,A_n\) and a starting
point \(Q_0\) satisfying \(\norm{Q_0Q_0^\top-V_kV_k^\top}_F\le c_0/2\) outputs
\(Q_T\) with high probability satisfying
\[
    \norm{Q_TQ_T^\top-V_kV_k^\top}_F
    \le
    \operatorname{Opt}_T
    +
    \tilde O\!\left(
        \frac{\sigma\sqrt{(\theta_1+\sigma^2)kd}}{\theta_k\sqrt n}
        +
        \frac{Kdk\,\sigma\sqrt{\theta_1+\sigma^2}}{\varepsilon\theta_k n}
        +
        \frac{\sqrt{k}M_\perp}{\theta_k n}
    \right),
\]
where
\(
    \operatorname{Opt}_T
    :=
    \exp\!\left(-c\,\theta_k\sum_{t=1}^T\eta_t\right)
    \norm{Q_0Q_0^\top-V_kV_k^\top}_F
\)
and \(\eta_t\) are the learning rates. 
\end{corollary}
\begin{remark}
    When the linear Bernstein term is lower order, \(M_\perp\lesssim\sigma\sqrt{(\theta_1+\sigma^2)dn}\), and \(K=O(1)\), this matches the known flat-tail spiked-covariance lower-bound scaling~(\Cref{spiked-cov-lower-bound}) up to logarithmic factors.
\end{remark}

\begin{wrapfigure}{r}{0.42\textwidth}
    \ifarxiv
    \vspace{0mm}
    \else
    \vspace{-5mm}
    \fi
    \centering
    \includegraphics[width=0.40\textwidth]{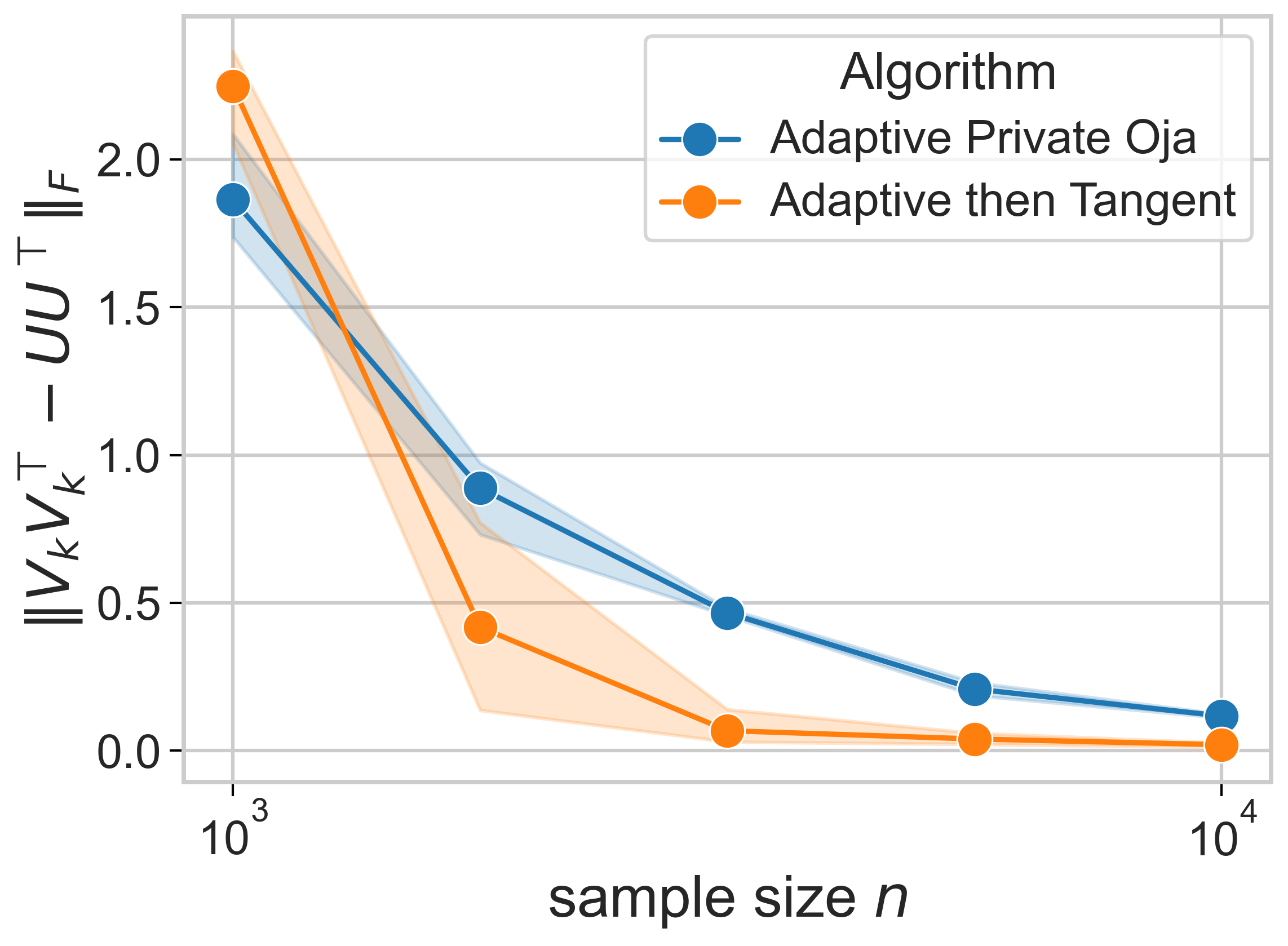}
    \caption{\small Comparison of \(\AdaptPrivOja\) with a two-stage variant
    that warm-starts with \(\AdaptPrivOja\) and then runs \(\TangAlg\).}
    \label{fig:tangent-refinement-spiked}
    \ifarxiv
    \vspace{0mm}
    \else
    \vspace{-15mm}
    \fi
\end{wrapfigure}

The experiment in \Cref{fig:tangent-refinement-spiked} illustrates the effect
of the tangent refinement phase. We first run \(\AdaptPrivOja\) to obtain a
private constant-accuracy warm start, and then run \(\TangAlg\) on fresh
samples. Once the warm start is accurate enough, the tangent phase exploits the
smaller residual fluctuations and achieves lower error than continuing with
\(\AdaptPrivOja\) alone. For the general privacy guarantee and the full proofs
for \TangAlg, see~\Cref{app:tangent}.

%% file: sections/related_work_and_conclusion.tex
\section{Related Work and Conclusion}\label{sec:conclusion}

\noindent\textbf{Related Work.} Private PCA has a long history in the fixed-dataset setting
\citep{blum2005practical,chaudhuri2013near,hardt2013beyond,dwork2014analyze,
florina2016improved,nicolas2024differentially}. These methods provide
input-wise privacy guarantees, but when applied directly to stochastic PCA
they typically rely on worst-case sensitivity bounds or clipping, leading to
suboptimal stochastic rates.

Adaptive private algorithms for stochastic PCA were introduced by
\citet{liu2022dp} for \(k=1\), using privacy noise calibrated to stochastic
fluctuations. The extension of this idea to \(k>1\) by
\citet{dungler2025iterative} is deflation-based and therefore incurs extra
dependence on \(k\). In contrast, our method treats the top-\(k\) subspace
directly through block updates. \citet{cai2024optimal} obtain optimal
guarantees for Gaussian data under a flat-tail condition on the unknown
covariance \(\Sigma\). In contrast our privacy guarantee instead holds input-wise for
every dataset.

Dimension-independent
sample complexity bounds have also been obtained under strong multiplicative eigengap
assumptions~\citep{singhal2021privately,tsfadia2024differentially}. Recent work on
private spectral perturbation and low-rank approximation~\citep{tran2025spectral}
analyzes a fixed noisy matrix, whereas our algorithm releases only adaptive projected
block actions. Other results study coherent  matrix models
\citep{d2026tight}, or decentralized privacy models \citep{campbell2025decentralized,nicolas2024differentially}. However, they use
different adjacency notions which when transformed to our more general adjacency notion yield suboptimal error-guarantees. A more detailed comparison with these results is provided in \Cref{app:related-work-details}.

\noindent\textbf{Open Problems and Conclusion.} Several directions remain open. Since \AdaptPrivOja is not deflation-based and updates the full \(d\times k\) iterate directly, it is naturally suited to streaming implementations. An interesting next step is to develop a fully online version with adaptive privacy calibration and comparable utility guarantees. A second direction is to weaken the tail assumptions behind the private range estimator. Our current analysis relies on concentration of the projected fluctuations, and extending the range-estimation and truncation steps to heavy-tailed data would substantially broaden the applicability of the method.

To the best of our knowledge, our results give
the first differentially private algorithm for stochastic \(k\)-PCA that is
sample-complexity optimal in both \(d\) and \(k\) as well as spectral parameters in some settings, while calibrating its privacy noise to the inherent stochasticity of the data.

%% file: appendices/math_preliminaries.tex
\section{Math Preliminaries}

\begin{lemma}[Variational characterization of the Schatten-\((2,k)\) norm]\label{lem:schatten-2-k-variational}
Let \(B \in \mathbb{R}^{d\times d}\), and let
\[
\mathbb S_{d,k} := \{P \in \mathbb{R}^{d\times k} : P^\top P = I_k\}.
\]
Then
\[
\sup_{P \in \mathbb S_{d,k}} \|P^\top B\|_F
=
\|B\|_{(2,k)}
:=
\left(\sum_{j=1}^k \sigma_j(B)^2\right)^{1/2},
\]
where \(\sigma_1(B)\ge \cdots \ge \sigma_d(B)\ge 0\) are the singular values of \(B\).
\end{lemma}

\begin{proof}
For any \(P \in \mathbb S_{d,k}\),
\[
\|P^\top B\|_F^2
=
\text{Tr}(B^\top P P^\top B)
=
\text{Tr}(P^\top B B^\top P).
\]
Since \(P^\top P = I_k\), the matrix \(P P^\top\) is the orthogonal projector onto a \(k\)-dimensional subspace of \(\mathbb{R}^d\). Hence
\[
\|P^\top B\|_F^2
=
\text{Tr}(P^\top B B^\top P)
\]
is the sum of the Rayleigh quotients of \(B B^\top\) over an orthonormal \(k\)-frame. By Ky Fan's maximum principle,
\[
\text{sup}_{P \in \mathbb S_{d,k}} \text{Tr}(P^\top B B^\top P)
=
\sum_{j=1}^k \lambda_j(B B^\top),
\]
where \(\lambda_1(B B^\top)\ge \cdots \ge \lambda_d(B B^\top)\ge 0\) are the eigenvalues of \(B B^\top\). Since
\[
\lambda_j(B B^\top) = \sigma_j(B)^2,
\]
it follows that
\[
\sup_{P \in \mathbb S_{d,k}} \|P^\top B\|_F^2
=
\sum_{j=1}^k \sigma_j(B)^2.
\]
Taking square roots yields the claim.
\end{proof}

\begin{lemma}[Minibatch concentration in top-\(k\) Frobenius norm]
\label{lem:minibatch-topk-concentration}
Let \(X_i := A_i-\Sigma\), and suppose \Cref{assumption-1} holds.
Let \(I_t\subset[n]\) be disjoint batches of size \(m\), and define
\[
    \bar X_t := \frac1m\sum_{i\in I_t} X_i .
\]
Then there exists a universal constant \(C>0\) such that, with probability at least
\(1-\zeta\), simultaneously for all \(t\in[T]\),
\[
    \|\bar X_t\|_{(2,k)}
    \le
    C\sqrt{k}
    \left(
        \lambda_1
        \sqrt{\frac{V\log(2dT/\zeta)}{m}}
        +
        \frac{M\log(2dT/\zeta)}{m}
    \right),
\]
\end{lemma}

\begin{proof}
Let \(X_i=A_i-\Sigma\). Then \(X_i\) are independent, symmetric, and mean-zero.

First, by \Cref{ass:A2},
\[
    \|X_i\|_{\mathrm{op}}
    \le
    \|X_i\|_{(2,k)}
    =
    \sup_{P\in\mathbb S_{d,k}}\|P^\top X_i\|_F
    \le \lambda_1 M
    \qquad\text{a.s.}
\]
Hence each summand is almost surely bounded in operator norm by \(\lambda_1 M\).

By the Liu-style variance assumption \Cref{ass:A3},
\[
    \left\|\mathbb E[X_i^2]\right\|_{\mathrm{op}}
    \le
    \lambda_1^2 V .
\]

Now fix a batch \(I_t\) of size \(m\). Matrix Bernstein applied to
\[
    \sum_{i\in I_t} X_i
\]
with almost-sure bound \(\lambda_1 M\) and variance proxy
\[
    \left\|
        \sum_{i\in I_t}\mathbb E[X_i^2]
    \right\|_{\mathrm{op}}
    \le
    m\lambda_1^2 V
\]
implies that, with probability at least \(1-\alpha\),
\[
    \left\|
        \frac1m\sum_{i\in I_t}X_i
    \right\|_{\mathrm{op}}
    \le
    C\lambda_1
    \left(
        \sqrt{\frac{V\log(2d/\alpha)}{m}}
        +
        \frac{M\log(2d/\alpha)}{m}
    \right).
\]
Set \(\alpha=\zeta/T\) and union bound over \(t\in[T]\). Then, with probability at least
\(1-\zeta\), the above operator-norm bound holds simultaneously for every batch.

Finally, for every matrix \(B\),
\[
    \|B\|_{(2,k)}
    =
    \left(\sum_{j=1}^k \sigma_j(B)^2\right)^{1/2}
    \le
    \sqrt{k}\,\|B\|_{\mathrm{op}} .
\]
Applying this inequality to \(B=\bar X_t\) gives
\[
    \|\bar X_t\|_{(2,k)}
    \le
    C\sqrt{k} \lambda_1
    \left(
        \sqrt{\frac{V\log(2dT/\zeta)}{m}}
        +
        \frac{M\log(2dT/\zeta)}{m}
    \right)
\]
simultaneously for all \(t\in[T]\), as claimed.
\end{proof}

\begin{definition}[$\zeta$-approximate Utility]\label{def-zeta-useful}
We say $U \in \mathbb{R}^{d \times k}$ is $\zeta$-approximate if $U$ has orthonormal columns and
\begin{align*}
    \langle U U^{\top}, \Sigma \rangle \geq ( 1 - \zeta^2)\langle V_k V_k^{\top}, \Sigma \rangle.
\end{align*}
\end{definition}

\begin{restatable}{lemma}{lemenergytofrob}\label{lem:energytofrob}
    If \(\langle U U^{\top}, \Sigma \rangle \geq ( 1 - \zeta^2)\langle V_k V_k^{\top}, \Sigma \rangle\) then
    \[
        \| U U^\top  - V_k V_k^\top\|_F \leq \zeta \sqrt{\frac{2 \sum_{i=1}^k \lambda_i}{\lambda_k - \lambda_{k+1}}}
    \]
\end{restatable}

\begin{proof}
Let
\[
    P := UU^\top,
    \qquad
    P_\star := V_k V_k^\top.
\]
Since \(P_\star\) projects onto the top-\(k\) eigenspace of \(\Sigma\), we have
\[
    \langle P_\star, \Sigma \rangle
    =
    \sum_{i=1}^k \lambda_i.
\]
The assumption implies
\[
    \langle P_\star, \Sigma \rangle - \langle P, \Sigma \rangle
    \leq
    \zeta^2 \langle P_\star, \Sigma \rangle
    =
    \zeta^2 \sum_{i=1}^k \lambda_i.
\]

We next lower bound the same quantity in terms of the distance between projectors.
Let \(v_1,\ldots,v_d\) be an orthonormal eigenbasis of \(\Sigma\), ordered so that
\[
    \Sigma = \sum_{i=1}^d \lambda_i v_i v_i^\top.
\]
Define
\[
    a_i := v_i^\top P v_i.
\]
Because \(P\) is a rank-\(k\) orthogonal projector, \(0 \leq a_i \leq 1\) and
\[
    \sum_{i=1}^d a_i = \operatorname{tr}(P) = k.
\]
Therefore,
\begin{align*}
    \langle P_\star, \Sigma \rangle - \langle P, \Sigma \rangle
    &=
    \sum_{i=1}^k \lambda_i (1-a_i)
    -
    \sum_{i=k+1}^d \lambda_i a_i \\
    &\geq
    \lambda_k \sum_{i=1}^k (1-a_i)
    -
    \lambda_{k+1} \sum_{i=k+1}^d a_i.
\end{align*}
Since
\[
    \sum_{i=1}^k (1-a_i)
    =
    k - \sum_{i=1}^k a_i
    =
    \sum_{i=k+1}^d a_i,
\]
we get
\[
    \langle P_\star, \Sigma \rangle - \langle P, \Sigma \rangle
    \geq
    (\lambda_k-\lambda_{k+1})
    \left(k-\sum_{i=1}^k a_i\right).
\]
Moreover,
\[
    \sum_{i=1}^k a_i
    =
    \operatorname{tr}(P_\star P),
\]
so
\[
    k-\sum_{i=1}^k a_i
    =
    k-\operatorname{tr}(P_\star P).
\]
Using the fact that \(P\) and \(P_\star\) are rank-\(k\) orthogonal projectors,
\[
    \|P-P_\star\|_F^2
    =
    \operatorname{tr}(P) + \operatorname{tr}(P_\star)
    -
    2\operatorname{tr}(PP_\star)
    =
    2\left(k-\operatorname{tr}(PP_\star)\right).
\]
Hence
\[
    \langle P_\star, \Sigma \rangle - \langle P, \Sigma \rangle
    \geq
    \frac{\lambda_k-\lambda_{k+1}}{2}
    \|P-P_\star\|_F^2.
\]
Combining this with the upper bound from the assumption gives
\[
    \frac{\lambda_k-\lambda_{k+1}}{2}
    \|P-P_\star\|_F^2
    \leq
    \zeta^2 \sum_{i=1}^k \lambda_i.
\]
Therefore,
\[
    \|P-P_\star\|_F^2
    \leq
    \zeta^2
    \frac{2\sum_{i=1}^k \lambda_i}{\lambda_k-\lambda_{k+1}}.
\]
Taking square roots and substituting back \(P=UU^\top\) and
\(P_\star=V_kV_k^\top\) yields
\[
    \| U U^\top  - V_k V_k^\top\|_F
    \leq
    \zeta \sqrt{\frac{2 \sum_{i=1}^k \lambda_i}{\lambda_k - \lambda_{k+1}}}.
\]
\end{proof}

\begin{lemma}[High-probability bound for GOE]
\label{lem:goe-topk-sch2}
Let \(G \sim \mathrm{GOE}_d(\sigma^2)\). Then for every \(1 \le k \le d\) and every \(t \ge 0\),
\[
\Pr\!\left(
\|G\|_{(2,k)}
\;\le\;
\sqrt{k}\,\bigl(2\sigma\sqrt{d}+t\bigr)
\right)
\ge
1-2\exp\!\left(-\frac{t^2}{4\sigma^2}\right).
\]
Equivalently, for every \(\delta \in (0,1)\),
\[
\Pr\!\left(
\|G\|_{(2,k)}
\;\le\;
2\sigma\sqrt{kd}
+
2\sigma\sqrt{k\log\!\left(\frac{2}{\delta}\right)}
\right)
\ge
1-\delta.
\]
\end{lemma}

\begin{proof}
By~\Cref{lem:schatten-2-k-variational},
\[
\|G\|_{(2,k)}
=
\sup_{P \in \mathbb S_{d,k}} \|P^\top G\|_F
=
\left(\sum_{i=1}^k \sigma_i(G)^2\right)^{1/2}
\le
\sqrt{k}\,\sigma_1(G)
=
\sqrt{k}\,\|G\|_{\mathrm{op}} .
\]
For \(G \sim \mathrm{GOE}_d(\sigma^2)\), the spectral norm satisfies
\[
\Pr\!\left(
\|G\|_{\mathrm{op}} \ge 2\sigma\sqrt{d}+t
\right)
\le
2\exp\!\left(-\frac{t^2}{4\sigma^2}\right),
\qquad t \ge 0,
\]
which proves the first claim. Setting \(t = 2\sigma\sqrt{\log(2/\delta)}\) yields the second.
\end{proof}

\begin{lemma}[Neumann series]\label{lem:neumann}
Let $A$ be a random square matrix, let $G$ be an event, and suppose that
$\|A\mathbf 1_G\|_2\le\varepsilon$ almost surely for a constant $\varepsilon<1$. Then on $G$ the
matrices $I-A$, $I+A$ and $I-A^2$ are invertible, and
\begin{align}
  \big\|(I-A^2)^{-1}\mathbf 1_G\big\|_2&\le\frac1{1-\varepsilon^2},\label{eq:neu1}\\
  \big\|\big[(I-A^2)^{-1}-I\big]\mathbf 1_G\big\|_2
   &\le\frac{\varepsilon^2}{1-\varepsilon^2}.\label{eq:neu2}
\end{align}
Consequently, for every random matrix $X$ with $X=X\mathbf 1_G$,
\begin{equation}\label{eq:neu3}
  \|X\|_F\le\frac1{1-\varepsilon^2}\,\big\|X(I-A^2)\big\|_F\qquad\text{pointwise.}
\end{equation}
Outside $G$ nothing is asserted: the hypothesis constrains $A$ only where $\mathbf 1_G=1$, so
$I-A^2$ need not be invertible there, and $(I-A^2)^{-1}$ is to be read as an arbitrary measurable
choice, say $I$. The choice is immaterial, since $(I-A^2)^{-1}$ occurs only against
$\mathbf 1_G$.
\end{lemma}

\begin{proof}
Fix $\omega\in G$ and put $B:=A(\omega)^2$. Submultiplicativity of $\|\cdot\|_2$ and the
hypothesis give $\|B\|_2\le\|A(\omega)\|_2^2\le\varepsilon^2<1$. Hence $\|B^j\|_2\le
\|B\|_2^j\le\varepsilon^{2j}$ and the series $\sum_{j\ge0}B^j$ converges absolutely in operator
norm. Its partial sums telescope against $I-B$,
\[
  (I-B)\sum_{j=0}^{m}B^j=I-B^{m+1}\xrightarrow[m\to\infty]{}I ,
\]
because $\|B^{m+1}\|_2\le\varepsilon^{2(m+1)}\to0$; the same computation on the other side gives
$\sum_{j=0}^mB^j(I-B)\to I$. Therefore $I-B=I-A(\omega)^2$ is invertible with
$(I-A(\omega)^2)^{-1}=\sum_{j\ge0}B^j$ and
\[
  \big\|(I-A(\omega)^2)^{-1}\big\|_2\le\sum_{j\ge0}\|B\|_2^{\,j}
  \le\sum_{j\ge0}\varepsilon^{2j}=\frac1{1-\varepsilon^2},
\]
which is \eqref{eq:neu1}. Since $A$ commutes with itself, $I-A^2=(I-A)(I+A)$, and a product of
square matrices is invertible only if both factors are; this gives the invertibility of
$I\pm A$. Subtracting the term $j=0$ from the series,
\[
  (I-A(\omega)^2)^{-1}-I=\sum_{j\ge1}B^j=B\,(I-A(\omega)^2)^{-1},
\]
so its norm is at most $\varepsilon^2(1-\varepsilon^2)^{-1}$, which is \eqref{eq:neu2}.

For \eqref{eq:neu3}: on $G$ we may write $X=X(I-A^2)(I-A^2)^{-1}$, while off $G$ both sides of
the asserted inequality are unaffected because $X=X\mathbf 1_G$ vanishes there. Hence, pointwise,
$\|X\|_F\le\|X(I-A^2)\|_F\,\|(I-A^2)^{-1}\mathbf 1_G\|_2\le(1-\varepsilon^2)^{-1}\|X(I-A^2)\|_F$ by \eqref{eq:neu1}, which is the claim.
\end{proof}

%% file: appendices/additional_comparisons_to_prior_work.tex
\section{Comparisons to Prior Work}
\label{app:related-work-details}

We provide additional details on the comparisons summarized in the related-work
section~(\Cref{sec:conclusion}).

\paragraph{Comparison to deterministic bounded-input methods.}
Several works address $k$-PCA in the standard setting, while assuming an additive eigengap~\citep{blum2005practical,chaudhuri2013near,hardt2013beyond,dwork2014analyze, nicolas2024differentially}. These works operate in a deterministic setting where each sample is assumed to be bounded ($\|x_i \| \leq \beta$). When applied to the stochastic setting, these works generally yield suboptimal error rates. This is partially due to the fact that all of these works assume a data independent bound ($\beta = 1$), which we cannot easily enforce in the stochastic setting. Considering Gaussian data with $x_i \sim \mathcal{N}(0, \Sigma)$, we know $\|x_i \| \leq \beta = O(\sqrt{\lambda_1 d \log(n/\zeta)}$ for all $i$ with probability $1 - \zeta$.~\citep{blum2005practical, dwork2014analyze, nicolas2024differentially} use the Gaussian mechanism, so when scaling the privacy noise with a factor $\beta$ we ensure $(\epsilon, \delta)$-DP in the stochastic setting. The tightest of the previous discussed result then achieves \[\bigO{\sqrt{dk/n} + d^{3/2}k/(\varepsilon n)}.\]

\paragraph{Comparison to \citet{dungler2025iterative}.}
Under assumptions comparable to \Cref{assumption-1}, and for inputs of the form
\(A_i=x_i x_i^\top\) with \(x_i\sim\mathcal{N}(0,\Sigma)\), the main result of
\citet{dungler2025iterative} implies that, with high probability, the output subspace
\(U\) is \(\zeta\)-approximate in the sense that
\[
    \langle U U^{\top}, \Sigma \rangle
    \geq
    (1-\zeta^2)\langle V_k V_k^{\top}, \Sigma \rangle,
\]
where
\[
    \zeta
    =
    \tilde{O}\left(
    \kappa'
    \left(
    \sqrt{\frac{dk}{n}}
    +
    \frac{dk\sqrt{\log(1/\delta)}}{\varepsilon n}
    \right)
    \right).
\]
Here \(\tilde O(\cdot)\) suppresses polylogarithmic factors in
\(n,d,1/\varepsilon\), and \(\log(1/\delta)\). By \Cref{lem:energytofrob}, this
energy guarantee implies the Frobenius-norm bound
\[
    \|UU^\top - V_kV_k^\top\|_F
    \leq
    \zeta
    \sqrt{
    \frac{2\sum_{i=1}^k \lambda_i}
    {\lambda_k-\lambda_{k+1}}
    }.
\]
Comparing this rate with the lower bound~\Cref{corollary-lower-bound-spiked-cov} in \Cref{sec:lower-bounds} shows that the
guarantee of \citet{dungler2025iterative} is loose by a factor \(\sqrt{k}\) in
Frobenius norm. The same loss appears when their energy-based utility is compared
to the corresponding lower bound. In addition, the error scaling in their main
theorem holds only under a sample-size condition containing an exponential
eigengap-dependent term. Our result removes this condition and attains the optimal
dependence on \(k\).

A further difference is algorithmic. The method of \citet{dungler2025iterative}
recovers the leading eigenspace one direction at a time, requiring stage-wise
choices of parameters. Our method instead updates the entire \(k\)-dimensional
subspace simultaneously. This avoids retuning across stages.

\paragraph{Comparison to fixed-matrix perturbation analyses.}
The results of \citet{tran2025spectral} give refined perturbation bounds for private
low-rank approximation by analyzing a fixed noisy matrix
\[
    \widetilde M = M+E
\]
and controlling the error between the best rank-\(p\) approximations of \(M\) and
\(\widetilde M\). These results are not directly applicable to our algorithm because
we do not release a noisy full covariance matrix. Instead, at step \(t\), our method
releases only the projected block action
\[
    \frac1n\sum_{i=1}^n A_i Q_{t-1}+Z_t,
\]
where \(Q_{t-1}\) depends on previous private iterates. The privacy noise is therefore
calibrated to the range of the projected matrices \(A_iQ_{t-1}\), rather than to a
global \(d\times d\) perturbation of the covariance matrix. This distinction is
essential: the projected stochastic fluctuations can be substantially smaller than
the worst-case full-matrix sensitivity.

\paragraph{Comparison to coherent matrix models.}
The work of \citet{d2026tight} studies DP-PCA under matrix coherence
assumptions in a non-stochastic setting where the input is a single matrix. Their
bounds are stated for
\[
    \|(I-UU^\top)V_k\|_2,
\]
which is related to the Frobenius subspace error by
\[
    \|V_kV_k^\top-UU^\top\|_F
    \leq
    \sqrt{2k}\,\|(I-UU^\top)V_k\|_2 .
\]
To apply their result to the empirical covariance
\[
    \widehat\Sigma=\frac1N\sum_{i=1}^N x_ix_i^\top,
\]
one must translate their matrix-level adjacency notion into sample-level privacy.
In their notation, two matrices \(M,M'\) are \(\Delta_{\rm mat}\)-adjacent if, for
\(E=M-M'\),
\[
    \sqrt{\|EE^\top\|_1}\le \Delta_{\rm mat}.
\]
Suppose the Gaussian samples are clipped so that \(\|x_i\|_2^2\le R_\eta^2\). Replacing
one sample \(x\) by another sample \(y\) changes the empirical covariance by
\[
    E=\frac1N(xx^\top-yy^\top).
\]
Hence
\[
    \Delta_{\rm mat}
    \le
    \frac{1}{N}\sqrt{\|(xx^\top-yy^\top)^2\|_1}
    \le
    \frac{2\sqrt d\,R_\eta^2}{N}.
\]
Plugging this into their rank-\(k\) private eigenspace bound, and using the worst-case
coherence bound \(k\mu_k(\widehat\Sigma)\le d\), gives a privacy error of order
\[
    \widetilde O\!\left(
    \frac{R_\eta^2 d}{N\varepsilon\gamma_k}
    \right),
    \qquad
    \gamma_k=\lambda_k(\Sigma)-\lambda_{k+1}(\Sigma).
\]
For \(x\sim\mathcal N(0,\Sigma)\), Gaussian quadratic-form concentration gives
\[
    R_\eta^2
    \lesssim
    \operatorname{tr}(\Sigma)
    +\sqrt{\lambda_1(\Sigma)\operatorname{tr}(\Sigma)\log(N/\eta)}
    +\lambda_1(\Sigma)\log(N/\eta)
    =
    \widetilde O(\operatorname{tr}(\Sigma)).
\]
Combining this privacy term with the usual non-private Davis--Kahan statistical
error yields the population subspace bound
\[
    \|(I-\widehat P)V_k\|_2
    \lesssim_{\log}
    \frac{\lambda_1(\Sigma)}{\gamma_k}\sqrt{\frac dN}
    +
    \frac{\operatorname{tr}(\Sigma)d}{N\varepsilon\gamma_k}.
\]
The privacy term arises from a worst-case matrix-adjacency conversion. It therefore
does not adapt to the stochastic spread of the projected gradients, which is the
source of the sharper dependence in our adaptive mechanism.

\paragraph{Decentralized and alternative adjacency models.}
There are also private PCA algorithms for decentralized settings
\citep{campbell2025decentralized,nicolas2024differentially}. These works use privacy
and partition models different from the sample-level stochastic setting considered
here. For example, \citet{campbell2025decentralized} use a power-method-based
approach under a neighboring relation based on changing one row of the data matrix,
with \(\|X-X'\|\le 1\). This does not directly correspond to our model, where each
sample contributes a matrix \(A_i=x_ix_i^\top\). In particular, changing one sample
changes one rank-one matrix contribution to the empirical covariance, rather than a
single bounded row in the data matrix under their adjacency model.

Similarly, the original noisy power method was analyzed for matrix inputs under
single-entry changes of magnitude at most one. Later analyses
\citep{florina2016improved,nicolas2024differentially} extended the privacy guarantee
to perturbations of the form \(A'=A+C\) satisfying
\[
    \sqrt{\sum_i \|C_{i,:}\|_1^2}\le 1.
\]
These matrix-adjacency guarantees are useful in fixed-dataset PCA, but they do not
directly yield the stochastic, sample-level rates considered in this paper.

%% file: appendices/proofs.tex
\section{Proofs}\label{app:proofs}

\subsection{Privacy of~\Cref{algo:priv-ojas-batches}}

\begin{restatable}[Privacy of \Cref{algo:priv-ojas-batches}]{theorem}{dpojabatchprivlem}
\label{thm:privacy-of-ojas-batches}
If \(0<\epsilon\le1\), then \Cref{algo:priv-ojas-batches} is \((\epsilon,\delta)\)-DP under replacement neighboring datasets.
\end{restatable}

\begin{proof}
At update step \(t\), the matrices \(A_i\) are accessed only through
\[
\mu_t = \frac{1}{B}\sum_{i=B(t-1)+1}^{tB} \operatorname{clip}_R(A_iQ_{t-1}).
\]
Although \(Q_{t-1}\) is data-dependent, it is determined entirely by the outputs of the previous update steps. Hence, conditional on all previous outputs, \(Q_{t-1}\) is fixed, and we may analyze the privacy of the \(t\)-th update step with \(Q_{t-1}\) treated as a constant.

If one matrix \(A_i\) in the \(t\)-th batch is replaced by \(A_i'\), producing \(\mu_t'\), then
\[
\|\mu_t-\mu_t'\|_F
=
\frac{1}{B}\left\|
\operatorname{clip}_R(A_iQ_{t-1})
-
\operatorname{clip}_R(A_i'Q_{t-1})
\right\|_F
\le
\frac{2R}{B}.
\]
Therefore, by \Cref{lem:privacy-batched-symm}, the \(t\)-th update step is \((\varepsilon,\delta)\)-DP, even conditional on the previous outputs. Since the algorithm is one-pass and each datum \(A_i\) is used in exactly one batch, the overall algorithm is \((\varepsilon,\delta)\)-DP by parallel composition.
\end{proof}

\begin{lemma}[Privacy of Single Update step of~\Cref{algo:priv-ojas-batches}]
\label{lem:privacy-batched-symm}
Fix \(R>0\) and \(Q \in \mathbb R^{d\times k}\) with \(Q^\top Q=I_k\), and define
\[
f_Q(A_1,\dots,A_B)
:=
\frac{1}{B}\sum_{i=1}^B \operatorname{clip}_R(A_iQ),
\]
where \(A_1,\ldots,A_B\) are symmetric. Consider the mechanism
\[
M(A_1,\dots,A_B;Q)
=
f_Q(A_1,\dots,A_B)+W_Q,
\]
where 
\[
W_Q := QN + N_\perp,
\qquad
N_\perp=(I-QQ^\top)Z,
\]
with \(N \sim GOE_k(\sigma^2)\) and \(Z \in \mathbb{R}^{d \times k}\) having i.i.d.\ entries in \(\cN(0, \sigma^2)\). Then \(M(\cdot;Q)\) is \((\varepsilon,\delta)\)-differentially private whenever
\[
\sigma \ge \frac{\Delta_Q}{\varepsilon}\sqrt{2\log(1.25/\delta)},
\qquad
\Delta_Q
:=
\sup_{(A_1,\dots,A_B)\sim(A_1',\dots,A_B')}
\|f_Q(A_1,\dots,A_B)-f_Q(A_1',\dots,A_B')\|_F.
\]
In particular, if neighboring batches differ in one entry, then
\[
\Delta_Q \le \frac{2R}{B},
\]
so it suffices to choose
\[
\sigma \ge \frac{2R}{B\varepsilon}\sqrt{2\log(1.25/\delta)}.
\]
\end{lemma}

\begin{proof}
First, each \(\operatorname{clip}_R(A_iQ)\) lies in
\[
\mathcal S_Q=\{M\in\mathbb R^{d\times k}:Q^\top M \text{ is symmetric}\}.
\]
Indeed, \(A_iQ\in\mathcal S_Q\) because \(Q^\top A_iQ\) is symmetric whenever
\(A_i=A_i^\top\), and clipping only rescales by a nonnegative scalar. Hence the
average \(f_Q(A_1,\dots,A_B)\) also lies in \(\mathcal S_Q\).

Next, since \(N\sim GOE_k(\sigma^2)\), we may write
\[
N \overset{d}{=} N_1+N_2,
\]
where \(N_1,N_2\) are independent draws from
\(\mathrm{GOE}_k(\sigma^2/2)\). Therefore
\[
W_Q
=
QN+N_\perp
\overset{d}{=}
QN_1+N_\perp+QN_2.
\]
Define
\[
W_Q^{(1)}:=QN_1+N_\perp,
\qquad
W_Q^{(2)}:=QN_2 .
\]
Then
\[
M(A_1,\dots,A_B;Q)
\overset{d}{=}
f_Q(A_1,\dots,A_B)+W_Q^{(1)}+W_Q^{(2)}.
\]

We now verify that \(W_Q^{(1)}\) is an isotropic Gaussian on
\(\mathcal S_Q\). Every element of \(\mathcal S_Q\) can be written
uniquely as
\[
M=QS+C,
\qquad
S=S^\top,
\qquad
Q^\top C=0.
\]
The two components \(QS\) and \(C\) are orthogonal in Frobenius inner
product. The noise \(W_Q^{(1)}\) has exactly this form:
\[
W_Q^{(1)}=QN_1+N_\perp,
\qquad
N_1=N_1^\top,
\qquad
Q^\top N_\perp=0.
\]
By construction, \(N_1\) is the Frobenius-isotropic Gaussian on the
space of symmetric \(k\times k\) matrices, and
\(N_\perp=(I-QQ^\top)Z\) is the Frobenius-isotropic Gaussian on the
subspace \(\{C:Q^\top C=0\}\). Since these two parts are independent and
orthogonal, \(W_Q^{(1)}\) is a centered Gaussian with covariance
\(\sigma^2 I_{\mathcal S_Q}\) on \(\mathcal S_Q\).

Therefore \(f_Q+W_Q^{(1)}\) is the Gaussian mechanism in the Hilbert
space \((\mathcal S_Q,\langle\cdot,\cdot\rangle_F)\), and it is
\((\varepsilon,\delta)\)-DP whenever
\[
\sigma \ge \frac{\Delta_Q}{\varepsilon}\sqrt{2\log(1.25/\delta)}.
\]
Since \(W_Q^{(2)}\) is independent of the input data, adding it is
post-processing. Hence \(M(\cdot;Q)\) is also
\((\varepsilon,\delta)\)-DP under the same condition.

If neighboring batches differ only in the \(j\)-th entry, then
\[
f_Q(A_1,\dots,A_B)-f_Q(A_1',\dots,A_B')
=
\frac{1}{B}\Bigl(\operatorname{clip}_R(A_jQ)-\operatorname{clip}_R(A_j'Q)\Bigr),
\]
and so
\[
\|f_Q(A_1,\dots,A_B)-f_Q(A_1',\dots,A_B')\|_F
\le
\frac{1}{B}\Bigl(
\|\operatorname{clip}_R(A_jQ)\|_F+\|\operatorname{clip}_R(A_j'Q)\|_F
\Bigr)
\le \frac{2R}{B}.
\]
Thus \(\Delta_Q\le 2R/B\), which yields the claim.
\end{proof}

\subsection{Utility of~\Cref{algo:priv-ojas-batches}}

\dpojasbatchesutilitythm*

\begin{proof}
    Let \(B=\lfloor n/T\rfloor\).~\Cref{lem:no-clipping-high-prob} implies that for our choice of \(R\) with probability \(1-\zeta\), \Cref{algo:priv-ojas-batches} does not have any clipping. Therefore the update step becomes
    \[
        Q_t' = Q_{t-1} + \eta_t \left( \frac{1}{B}\sum_{i=B\cdot (t-1)+1}^{tB}  A_i Q_{t-1} + Q_{t-1} N_{sym, t} + (I - Q_{t-1} Q_{t-1}^\top ) N_t\right)
    \]
    By~\Cref{lem:update-step-equivalence} in distribution this is equal to 
    \begin{align*}
         Q_t' &= Q_{t-1} + \eta_t \left( \frac{1}{B}\sum_{i=B\cdot (t-1)+1}^{tB}  A_i Q_{t-1} + G_t Q_{t-1}\right) \\
         &= Q_{t-1} + \eta_t \left( \frac{1}{B}\sum_{i=B\cdot (t-1)+1}^{tB}  A_i + G_t \right)Q_{t-1}
    \end{align*}
    where \(G_t\) is independently sampled from $G_t \sim GOE_d(\sigma^2)$.
    
    Let
    \[
        C_t = \frac{1}{B}\sum_{i=B\cdot (t-1)+1}^{tB}  A_i + G_t .
    \]
    We will show that $C_t$ satisfies Assumption \ref{ass:C}, so that Theorem \ref{thm:main}
    applies. The updates are i.i.d., as the noise multiplier is fixed before the run, so
    the conditional law of $C_t$ does not depend on the past. Since the blocks are disjoint and the Gaussian noise
    at step $t$ is fresh, we see straight away that
    \[
      \mathbb E\big[C_t\mid\mathcal F_{t-1}\big]
      = \frac1B\sum_{i=B(t-1)+1}^{tB}\mathbb E\big[A_i\mid\mathcal F_{t-1}\big]
        + \mathbb E\big[G_t\mid\mathcal F_{t-1}\big]
      = \frac1B\sum_{i=B(t-1)+1}^{tB}\mathbb E[A_i]
      = \Sigma ,
    \]
    as every entry of $G_t$ is a centered Gaussian.

    Next we show the upper bound on
    \[
        \|C_t-\Sigma\|_{(2,k)}
        =
        \sup_{P \in \mathbb S_{d,k}}
        \|P^\top(C_t-\Sigma)\|_F .
    \]
    By the triangle inequality,
    \begin{align*}
        \|C_t-\Sigma\|_{(2,k)}
        &\le
        \left\|
        \frac{1}{B}
        \sum_{i=B\cdot (t-1)+1}^{tB}
        (A_i-\Sigma)
        \right\|_{(2,k)}
        +
        \|G_t\|_{(2,k)} .
    \end{align*}

    By \Cref{lem:minibatch-topk-concentration}, with probability at least
    \(1-\zeta\), simultaneously for all \(t\in[T]\),
    \[
        \left\|
        \frac{1}{B}
        \sum_{i=B\cdot (t-1)+1}^{tB}
        (A_i-\Sigma)
        \right\|_{(2,k)}
        \le
         C\sqrt{k}\,\lambda_1
        \left(
            \sqrt{\frac{V\log(2dT/\zeta)}{B}}
            +
            \frac{M\log(2dT/\zeta)}{B}
        \right)
        =:\mathsf E_B .
    \]

    We know by~\Cref{lem:goe-topk-sch2} that with probability $1 - \zeta$,
    \[
        \|G_t\|_{(2,k)}
        \le
        C\sigma\sqrt{k(d+\log(T/\zeta))}
    \]
    for all $t \in [T]$.
    Since
    \[
        \sigma
        =
        \frac{2R\sqrt{2\ln(1.25/\delta)}}{B\varepsilon},
    \]
    and
    \[
        R
        =
        C\lambda_1\sqrt{dk}
        \bigl(K\gamma\log^a(ndk/\zeta)+1\bigr),
    \]
    we have
    \[
        \|G_t\|_{(2,k)}
        \le
        C
        \frac{
            \lambda_1 (K\gamma+1) k\sqrt{d(d+\log(T/\zeta))}
        }{
            \varepsilon B
        }
        \log^a(ndk/\zeta)
        \sqrt{\log(1/\delta)}
        =:\mathsf N_B .
    \]

    Set
    \[
        M' := \mathsf E_B+\mathsf N_B .
    \]
    Then, on the above high-probability event,
    \[
        \|C_t-\Sigma\|_{(2,k)}
        \le M'
        \qquad
        \text{for all }t\in[T].
    \]

    We apply~\Cref{thm:main} to the sequence
    \(C_1,\dots,C_T\) with \(M\) replaced by \(M'\).
    For every $\delta \in (0,1)$ define learning rates
    \[
        n_0 = \tilde{\Theta}\left(\frac{k M'^2}{\delta^2 \Delta_k^2}\right),
        \qquad
        \beta = \tilde{\Theta} \left( \max \left\{ \frac{\lambda_1 \log(n)}{\Delta_k}, \frac{M'^2 \log^2(n)}{\Delta_k^2}\right\}\right),
    \]
    and
    \[
        \eta_t =
        \begin{cases}
        \tilde{\Theta}\left(\frac{1}{\Delta_k n_0}\right) & t \leq n_0,\\[4pt]
        \tilde{\Theta}\left(\frac{1}{\Delta_k(\beta + t - n_0)}\right) & t > n_0.
        \end{cases}
    \]
    Then for $V \in \mathbb{R}^{d \times k}$ the orthogonal matrix whose columns are the $k$ leading eigenvectors of $\Sigma$, the algorithm satisfies
    \[
      \big\|Q_TQ_T^\top-V_kV_k^\top\big\|_F\le \sqrt{2}e\left(\frac{\beta + 1}{\beta + T - n_0}\right)^\alpha  + C'\,\frac{M'}{\Delta_k}
       \sqrt{\frac{\Lambda}{\,T-n_0\,}},
      \qquad
      \Lambda\asymp\log\frac1\delta+\log\Big(e+\frac{M'^2}{\Delta_k^2\delta}\Big),
    \]
    with probability at least $1-\delta$. With \(c_\eta=\lceil\log n\rceil\) one has
    \(\alpha=\tfrac12\lceil\log n\rceil\) and \(\beta\asymp\kappa\log n\), so
    \(T=\lceil C_T\kappa\log n\rceil\) gives \(S:=\beta+T-n_0\ge2e^2\beta\) for \(C_T\) large enough. Since \(\beta\ge1\) we have
\((\beta+1)/S\le e^{-2}\), so the first term of~\Cref{thm:main} is at most
\(\sqrt2\,e\,e^{-2\alpha}=\sqrt2\,e^{\,1-\lceil\log n\rceil}\le\sqrt2\,e/n\le n^{-1/2}\) for
\(n\ge2e^2\), and is dominated by the second.

    Taking \(\delta\) to be a constant multiple of \(\zeta\), using
    \(T\ge 2n_0\), and absorbing logarithmic factors into \(\tilde O(\cdot)\),
    \[
        \| Q_T Q_T^\top - V_k V_k^\top \|_F
        \le
        \tilde O
        \left(
            \frac{M'}{\Delta_k\sqrt T}
        \right).
    \]
    Substituting \(M'=\mathsf E_B+\mathsf N_B\), we get
    \[
        \| Q_T Q_T^\top - V_k V_k^\top \|_F
        \le
        \tilde O
        \left(
            \frac{\mathsf E_B}{\Delta_k\sqrt T}
            +
            \frac{\mathsf N_B}{\Delta_k\sqrt T}
        \right).
    \]
    Since \(T=n/B\), the minibatch term satisfies
    \[
        \frac{\mathsf E_B}{\sqrt T}
        =
        \tilde O
        \left(
            \lambda_1\sqrt{\frac{kV}{n}}
            +
            \sqrt{\kappa}\frac{\sqrt{k}M}{n}
        \right),
    \]
    while, using \(B=n/T\) with \(T = \lceil C_T \kappa \log n\rceil\),
    \[
        \frac{\mathsf N_B}{\sqrt T}
        =
        \tilde O
        \left(\sqrt{\kappa}
            \frac{\lambda_1 (K\gamma+1)kd}{\varepsilon n}
        \right).
    \]
    Therefore,
    \[
        \|Q_TQ_T^\top - V_kV_k^\top\|_F
        \le
        \tilde O
        \left(
            \frac{\lambda_1\sqrt{kV}}
            {\Delta_k\sqrt n}
            +
            \sqrt{\kappa}\frac{\sqrt{k}M}
            {\Delta_k n}
            +
            \sqrt{\kappa}\frac{\lambda_1 (K\gamma+1)kd}
            {\varepsilon\Delta_k n}
        \right).
    \]
    Finally, the stated lower bound on \(n\) is exactly the condition
    \(T\gtrsim kM'^2/(\zeta^2\Delta_k^2)\), after substituting
    \(B=n/T\), \(T = \lceil C_T \kappa \log n\rceil\) and suppressing logarithmic factors.
\end{proof}

\begin{lemma}[No clipping with high probability]\label{lem:no-clipping-high-prob}
Consider \Cref{algo:priv-ojas-batches} under Assumption~\ref{assumption-1}, and choose
\[
R \leftarrow C \lambda_1 \sqrt{dk}\,\bigl(K\gamma \log^{a}(ndk/\zeta)+1\bigr)
\]
for a sufficiently large universal constant \(C>0\). Then, with probability at least \(1-\zeta\), no clipping occurs throughout the algorithm, i.e.
\[
\|A_iQ_{t-1}\|_F \le R
\qquad
\text{for all } t\in[T],\; i\in\{B(t-1)+1,\dots,tB\}.
\]
Equivalently, on this event,
\[
\operatorname{clip}_R(A_iQ_{t-1}) = A_iQ_{t-1}
\qquad
\text{for all such } i,t.
\]
\end{lemma}

\begin{proof}
Fix \(t\in[T]\) and \(i\in\{B(t-1)+1,\dots,tB\}\). Since the algorithm is one-pass, \(Q_{t-1}\) depends only on the matrices \(A_1,\dots,A_{B(t-1)}\), and is therefore independent of \(A_i\). Hence, conditional on the past, \(Q_{t-1}\) is fixed.

Write the columns of \(Q_{t-1}\) as \(q_1^{(t-1)},\dots,q_k^{(t-1)}\). For any \(r\in[d]\) and \(s\in[k]\), Assumption~\ref{ass:A4} applied with \(P=I\), \(u=e_r\), and \(v=q_s^{(t-1)}\) yields
\[
\mathbb{E}\!\left[
\exp\!\left(
\left(
\frac{|e_r^\top (A_i-\Sigma) q_s^{(t-1)}|^2}{K^2\lambda_1^2\gamma^2}
\right)^{1/(2a)}
\right)
\,\middle|\, Q_{t-1}
\right]
\le 1.
\]
By a standard sub-Weibull tail bound, this implies that for a sufficiently large universal constant \(C>0\),
\[
|e_r^\top (A_i-\Sigma) q_s^{(t-1)}|
\le
C \lambda_1 K\gamma \log^a(ndk/\zeta)
\]
simultaneously for all \(i\in[n]\), \(r\in[d]\), and \(s\in[k]\) with probability at least \(1-\zeta\), by a union bound over the \(ndk\) such quantities.

On this event,
\[
\|(A_i-\Sigma)Q_{t-1}\|_F
\le
\sqrt{dk}\,
\max_{r\in[d],\,s\in[k]}
|e_r^\top (A_i-\Sigma) q_s^{(t-1)}|
\le
C \lambda_1 \sqrt{dk}\,K\gamma \log^a(ndk/\zeta).
\]
Moreover,
\[
\|\Sigma Q_{t-1}\|_F \le \|\Sigma\|_2 \|Q_{t-1}\|_F = \lambda_1\sqrt{k} \le \lambda_1\sqrt{dk}.
\]
Therefore,
\[
\|A_iQ_{t-1}\|_F
\le
\|\Sigma Q_{t-1}\|_F + \|(A_i-\Sigma)Q_{t-1}\|_F
\le
C \lambda_1 \sqrt{dk}\,\bigl(K\gamma \log^a(ndk/\zeta)+1\bigr).
\]
By the choice of \(R\), it follows that
\[
\|A_iQ_{t-1}\|_F \le R
\]
for all \(t\in[T]\) and all \(i\in\{B(t-1)+1,\dots,tB\}\). Thus no clipping occurs throughout the algorithm.
\end{proof}

\updatestepequiv*

\begin{proof}[Proof of~\Cref{lem:update-step-equivalence}]
Let \(U=[Q,Q_\perp]\in O(d)\). Since \(Z\) has i.i.d.\
\(\cN(0,\sigma^2)\) entries and \(Q_\perp\) has orthonormal columns,
\(Q_\perp^\top Z\) has i.i.d.\ \(\cN(0,\sigma^2)\) entries and is
independent of \(N_{\mathrm{sym}}\).

Now
\[
U^\top\!\left(QN_{\mathrm{sym}}+(I-P)Z\right)
=
\begin{pmatrix}
N_{\mathrm{sym}}\\
Q_\perp^\top Z
\end{pmatrix}.
\]
On the other hand, by orthogonal invariance of GOE,
\[
U^\top G U \sim \mathrm{GOE}_d(\sigma^2).
\]
Writing
\[
U^\top G U=
\begin{pmatrix}
A_0 & B^\top\\
B & H
\end{pmatrix},
\]
we have
\[
A_0\sim \mathrm{GOE}_k(\sigma^2),\qquad
B \text{ has i.i.d.\ }\cN(0,\sigma^2)\text{ entries,}
\]
and \(A_0,B\) are independent. Therefore
\[
U^\top GQ
=
\begin{pmatrix}
A_0\\
B
\end{pmatrix}
\stackrel{d}{=}
\begin{pmatrix}
N_{\mathrm{sym}}\\
Q_\perp^\top Z
\end{pmatrix}
=
U^\top\!\left(QN_{\mathrm{sym}}+(I-P)Z\right).
\]
Multiplying by \(U\) proves the claim.
\end{proof}

\subsection{Privacy of~\Cref{algo:priv-ojas-adaptive}}

\dpojasadaptiveprivlem*

\begin{proof}
Fix an iteration \(t\), and condition on all previous outputs. Then
\(Q=Q_{t-1}\) is fixed. The range is estimated \PrivRange, which is
\((\epsilon,\delta)\)-DP by~\Cref{thm:privrangemax}, and uses samples disjoint from \TruncatedMean and therefore parallel DP composition applies.

Now consider \TruncatedMean: Let \(C=(\bar g_{jr})\) be the coordinatewise center
computed by \textsc{TruncatedMean}. The \(dk\) coordinate histograms each use
privacy budget \((\epsilon/(2dk),\delta/(2dk))\), so by basic composition
\(C\) is \((\epsilon/2,\delta/2)\)-DP.

For any fixed value of \(C\), let \(\widetilde G_i(C)\) be \(A_iQ\) after
coordinatewise truncation to
\([C_{jr}-R_t,C_{jr}+R_t]\). Since
each coordinate of both truncated summands lies in an interval of length
\(2R_t\),
\[
    \|\widetilde G_i(C)-\widetilde G_i'(C)\|_F
    \le 2R_t\sqrt{dk}.
\]

and therefore for neighboring mean batches
\(S,S'\), coordinatewise truncation gives
\[
    \| \frac1m\sum_{i=1}^m \widetilde G_i(C) - \frac1m\sum_{i=1}^m \widetilde G_i'(C) \| \leq \frac{2R_t\sqrt{dk}}{m}.
\]
The algorithm then applies a projection $\mathsf P_{\mathcal S_Q}$ to the output of \TruncatedMean. Define
\[
    h_C(S)
    =
    \mathsf P_{\mathcal S_Q}\left(
        \frac1m\sum_{i=1}^m \widetilde G_i(C)
    \right).
\]
Then \(h_C(S)\in\mathcal S_Q\) so by~\Cref{lem:privacy-batched-symm} adding the noise matrix
\[
    W_Q
    =
    QN_{\mathrm{sym}}+(I-QQ^\top)Z .
\]
calibrated to twice the sensitivity of $h_C(S)$ is private. Since \(\mathsf P_{\mathcal S_Q}\) is a projection and therefore cannot increase the norm the sensitivity is
\[
    \|h_C(S)-h_C(S')\|_F
    \le
    \frac{2R_t\sqrt{dk}}{m}.
\]

For batch size $m = B/2$, and privacy parameters $(\varepsilon/2, \delta/2)$ this gives 
\[
    \sigma_t
    \ge
    \frac{8R_t\sqrt{dk}}{B\varepsilon}
    \sqrt{2\log(2.5/\delta)}.
\]
which is exactly the calibration in~\Cref{algo:priv-ojas-adaptive}.

By adaptive composition,
\[
    \bigl(C,h_C(S)+W_Q\bigr)
\]
is \((\varepsilon,\delta)\)-DP. Since the algorithm
releases only \(h_C(S)+W_Q\), discarding \(C\) is post-processing. By the budget
choice in~\Cref{algo:priv-ojas-adaptive}, the mean-half mechanism is therefore
\((\epsilon,\delta)\)-DP.

For \(t>T_0\) the \PrivRange and \TruncatedMean act on a disjoint set of inputs, so by parallel
composition the adaptive update is \((\epsilon,\delta)\)-DP. For \(t\le T_0\) the update is that
of~\Cref{algo:priv-ojas-batches}: the batch mean has Frobenius-norm sensitivity \(2R/B\) in a
single \(A_i\), and \(\sigma=2R\sqrt{2\log(1.25/\delta)}/(B\epsilon)\) is the Gaussian mechanism
calibrated to it, so the warm-up update is \((\epsilon,\delta)\)-DP
by~\Cref{thm:privacy-of-ojas-batches}. Since the algorithm is one-pass
and the update batches are disjoint across \(t\), the final output is also
\((\epsilon,\delta)\)-DP. 
\end{proof}

\subsection{Utility of~\Cref{algo:priv-ojas-adaptive}}

\dpojasadaptiveutilitythm*

\begin{proof}
    Let \(B=\lfloor n/T\rfloor\) and let
    \[
        I_t :=
        \{B(t-1)+B/2+1,\dots,tB\}.
    \]
    \Cref{lem:no-clipping-matrix} together with~\Cref{thm:privrangemax} imply
    that, under the stated sample-size conditions and after a union bound over
    \(t\in[T]\), with probability $1-\zeta$~\Cref{algo:trunc-mean} does not clip.
    Hence \TruncatedMean returns
    \[
        \frac1m\sum_{i\in I_t}A_iQ_{t-1},
        \qquad
        m=B/2.
    \]
    Since each \(A_i\) is symmetric, \(A_iQ_{t-1}\in\mathcal S_{Q_{t-1}}\).
    Therefore the raw mean is also in \(\mathcal S_{Q_{t-1}}\), and
    \[
        \mathsf P_{\mathcal S_{Q_{t-1}}}
        \left(
        \frac1m\sum_{i\in I_t}A_iQ_{t-1}
        \right)
        =
        \frac1m\sum_{i\in I_t}A_iQ_{t-1}.
    \]
    Therefore the update step becomes
    \[
        Q_t' = Q_{t-1} + \eta_t \left(
        \frac{2}{B}\sum_{i=B\cdot (t-1)+B/2 + 1}^{tB}
        A_i Q_{t-1}
        + Q_{t-1} N_{\mathrm{sym},t}
        + (I - Q_{t-1} Q_{t-1}^\top ) Z_t
        \right).
    \]
    By~\Cref{lem:update-step-equivalence} in distribution this is equal to 
    \begin{align*}
         Q_t'
         &=
         Q_{t-1}
         +
         \eta_t
         \left(
         \frac{2}{B}\sum_{i=B\cdot (t-1)+B/2 + 1}^{tB}
         A_i Q_{t-1}
         +
         G_t Q_{t-1}
         \right) \\
         &=
         Q_{t-1}
         +
         \eta_t
         \left(
         \frac{2}{B}\sum_{i=B\cdot (t-1)+B/2 + 1}^{tB}
         A_i
         +
         G_t
         \right)Q_{t-1},
    \end{align*}
    where \(G_t\) is independently sampled from
    $GOE_d(\sigma_t^2)$.

    Let $\mathcal F_{t-1}$ denote the sigma-field generated by all randomness used before iteration \(t\), and let $\mathcal R_t$ denote the randomness in the range-estimation half of the \(t\)-th batch. On the no-clipping event,~\Cref{lem:update-step-equivalence} lets us write the private update in the form \[ Q_t = \operatorname{QR}\bigl(Q_{t-1}+\eta_t C_tQ_{t-1}\bigr), \qquad C_t = \frac{2}{B}\sum_{i=B(t-1)+B/2+1}^{tB} A_i + G_t, \] where \(G_t\sim \mathrm{GOE}_d(\sigma_t^2)\), and \(\sigma_t\) is \(\sigma(\mathcal F_{t-1},\mathcal R_t)\)-measurable. 
    
    We will show that $C_t$ satisfies Assumption \ref{ass:C}, conditioned on a high-probability event, and that the warm-up updates are i.i.d., so that Theorem \ref{thm:main} applies. The samples in the mean half of the batch are disjoint from the range half, hence independent of \((\mathcal F_{t-1},\mathcal R_t,\sigma_t)\). Therefore \[ \mathbb E\!\left[C_t\mid \mathcal F_{t-1},\mathcal R_t\right] = \mathbb E\!\left[\frac{2}{B}\sum_{i=B(t-1)+B/2+1}^{tB} A_i\right] + \mathbb E[G_t\mid \mathcal F_{t-1},\mathcal R_t] = \Sigma. \] Moreover, conditional on \((\mathcal F_{t-1},\mathcal R_t)\), the matrix \(C_t\) is symmetric and uses fresh randomness independent of the current iterate \(Q_{t-1}\).

    Next we show the upper bound on
    \[
        \|C_t-\Sigma\|_{(2,k)}
        =
        \sup_{P\in\mathbb S_{d,k}}\|P^\top(C_t-\Sigma)\|_F .
    \]
    By the triangle inequality,
    \begin{align*}
        \|C_t-\Sigma\|_{(2,k)}
        &\le
        \left\|
        \frac{2}{B}
        \sum_{i=B\cdot (t-1)+B/2 + 1}^{tB}
        (A_i-\Sigma)
        \right\|_{(2,k)}
        +
        \|G_t\|_{(2,k)} .
    \end{align*}

    By \Cref{lem:minibatch-topk-concentration}, with probability at least
    \(1-\zeta\), simultaneously for all \(t\in[T]\),
    \[
        \left\|
        \frac{2}{B}
        \sum_{i=B\cdot (t-1)+B/2 + 1}^{tB}
        (A_i-\Sigma)
        \right\|_{(2,k)}
        \le
        C\sqrt{k} \lambda_1
        \left(
            \sqrt{\frac{V\log(2dT/\zeta)}{B}}
            +
            \frac{M\log(2dT/\zeta)}{B}
        \right)
        =:\mathsf E_B .
    \]

    Moreover, by \Cref{thm:privrangemax},
    \[
        \widehat{\Lambda}_t
        \le
        C\lambda_1^2\gamma^2
    \]
    simultaneously for all \(t\in[T]\). Hence
    \[
        R_t
        =
        3K\sqrt{\widehat{\Lambda}_t}
        \log^a(Bdk/\zeta)
        \le
        C K\lambda_1\gamma
        \log^a(BdkT/\zeta)
        =:R_\star .
    \]
    Since the Gaussian noise variance in iteration \(t\) is calibrated using \(R_t\),
    \[
        \sigma_t
        \le
        \frac{8R_\star\sqrt{2dk\log(2.5/\delta)}}{\varepsilon B}
        =:\sigma_\star .
    \]
    By~\Cref{lem:goe-topk-sch2} and a union bound over \(t\in[T]\), with probability
    at least \(1-\zeta\),
    \[
        \|G_t\|_{(2,k)}
        \le
        C\sigma_\star\sqrt{k(d+\log(T/\zeta))}
    \]
    for all $t \in [T]$. Therefore,
    \[
        \|G_t\|_{(2,k)}
        \le
        C
        \frac{
            \lambda_1K\gamma\,k\sqrt{d(d+\log(T/\zeta))}
        }{
            \varepsilon B
        }
        \log^a(BdkT/\zeta)
        \sqrt{\log(1/\delta)}
        =:\mathsf N_B .
    \]

    Set $M' := E_B+N_B$, and let $\mathcal E$ be the intersection of the range-estimation, no-clipping, minibatch-concentration, and Gaussian-noise events above. On \(\mathcal E\), \[
        \|C_t-\Sigma\|_{(2,k)}
        \le M'
        \qquad
        \text{for all }t\in[T].
    \] Thus, on \(\mathcal E\), ADADPO coincides with Oja's method run on the symmetric effective matrices \(C_1,\ldots,C_T\). 
    Since the range half is disjoint from the mean half, the adaptive choice of $\sigma_t$ is
    measurable with respect to $\widetilde{\mathcal F}_{t-1}=\sigma(\mathcal F_{t-1},\mathcal R_t)$,
    while the mean-half samples and the GOE base noise at time $t$ are fresh. Assumption \ref{ass:C}
    therefore holds on $\mathcal E$ with respect to the pre-update filtration
    $(\widetilde{\mathcal F}_t)$ of \eqref{eq:preupdate} and with constant $M'$: each update matrix is
    symmetric, satisfies $\mathbb E[C_t\mid\widetilde{\mathcal F}_{t-1}]=\Sigma$, and obeys
    $\|C_t-\Sigma\|_{(2,k)}\le M'$. This is Lemma \ref{lem:Gverify}, with $M=M'$. It remains to check the hypothesis that
\Cref{thm:main} places on Phase I. On the warm-up batches $t\le T_0$ the multiplier
$\sigma_t\equiv\sigma$ is fixed before the run and clipping is inactive on $\mathcal E$, so
\[
    C_t=\frac1B\sum_{i=B(t-1)+1}^{tB}A_i+G_t
\]
does not depend on $Q_{t-1}$: those updates are i.i.d.\ and independent of $Z_0$, with the
constant $M'_{\mathrm I}$ of~\Cref{thm:utility-priv-ojas-batches} in place of $M'$. Phase II, on
$(T_0,T]$, uses the adaptive constant $M'$ and nothing beyond \Cref{ass:C}. Applying Theorem \ref{thm:main} and reducing the final success
    probability by $\mathbb P(\mathcal E^c)$ gives the following.

    For every $\delta \in (0,1)$ define learning rates
    \[
        n_0 =
        \tilde{\Theta}\left(\frac{k M'^2}{\delta^2 \Delta_k^2}\right),
        \qquad
        \beta =
        \tilde{\Theta}\left(\frac{M'^2}{\Delta_k^2}\right),
    \]
    and
    \[
        \eta_t =
        \begin{cases}
        \tilde{\Theta}\left(\frac{1}{\Delta_k n_0}\right),
        & t \leq n_0,\\[4pt]
        \tilde{\Theta}\left(\frac{1}{\Delta_k(\beta + t - n_0)}\right),
        & t > n_0.
        \end{cases}
    \]
    Then for $V \in \mathbb{R}^{d \times k}$ the orthogonal matrix whose columns are
    the $k$ leading eigenvectors of $\Sigma$, the algorithm satisfies
    \[
        \| Q_T Q_T^\top - V_k V_k^\top \|_F
        \leq
        \sqrt2\,e\left(\frac{\beta+1}{\beta+T-T_0}\right)^{\alpha}
        +
        C' \frac{M'}{\Delta_k}
        \sqrt{\frac{\Lambda}{T - T_0}},
        \qquad
        \Lambda \asymp \log\frac1\delta + \log\!\left(e + \frac{M'^2}{\Delta_k^2\delta}\right),
    \]
    with probability at least $1-\delta$.

       With \(c_\eta=\lceil\log n\rceil\) one has \(\alpha=\tfrac12\lceil\log n\rceil\) and
    \(\beta\asymp\kappa\log n\), so \(T=\lceil C_T\kappa\log n\rceil\) and \(T_0=\lceil T/2\rceil\)
    give \(S:=\beta+T-T_0\ge2e^2\beta\) for \(C_T\) large enough. Since \(\beta\ge1\) we have
    \((\beta+1)/S\le e^{-2}\), so the first term is at most
    \(\sqrt2\,e\,e^{-2\alpha}\le\sqrt2\,e/n\le n^{-1/2}\) for \(n\ge2e^2\) and is dominated by the
    second. Taking \(\delta\) to be a constant multiple of \(\zeta\) and absorbing logarithmic
    factors into \(\tilde O(\cdot)\),
    \[
        \| Q_T Q_T^\top - V_k V_k^\top \|_F
        \le
        \tilde O
        \left(
            \frac{M'}{\Delta_k\sqrt{T}}
        \right).
    \]
    Substituting \(M'=\mathsf E_B+\mathsf N_B\) gives
    \[
        \| Q_T Q_T^\top - V_k V_k^\top \|_F
        \le
        \tilde O
        \left(
            \frac{\mathsf E_B}{\Delta_k\sqrt T}
            +
            \frac{\mathsf N_B}{\Delta_k\sqrt T}
        \right).
    \]
    Since \(BT=n\) up to rounding, the minibatch term satisfies
    \[
        \frac{\mathsf E_B}{\sqrt T}
        =
        \tilde O
        \left(
                        \lambda_1\sqrt{\frac{kV}{n}}
            +
            \sqrt\kappa\,\frac{\sqrt{k}M}{n}
        \right),
    \]
    while, using \(B=\lfloor n/T\rfloor\) with \(T=\lceil C_T\kappa\log n\rceil\),
    \[
        \frac{\mathsf N_B}{\sqrt T}
        =
        \tilde O
        \left(
            \sqrt\kappa\,\frac{\lambda_1K\gamma\,kd}{\varepsilon n}
        \right).
    \]
    Therefore,
    \[
        \| Q_T Q_T^\top - V_k V_k^\top \|_F
        \le
        \tilde O
        \left(
            \frac{\lambda_1\sqrt{kV}}
            {\Delta_k\sqrt n}
            +
         \sqrt\kappa\,\frac{\sqrt{k}M}{\Delta_k n} + \sqrt\kappa\,\frac{\lambda_1K\gamma\,kd}{\varepsilon\Delta_k n}
         \right).
    \]
\end{proof}

\begin{lemma}[No clipping for matrix-valued truncated mean]
\label{lem:no-clipping-matrix}
Fix an iteration $t$, and let $Q=Q_{t-1}\in\mathbb{R}^{d\times k}$ have orthonormal columns
$q_1,\dots,q_k$, where $Q$ is measurable with respect to the previous batches and hence
independent of the current mean batch $\{A_i\}_{i=1}^m$.
Let
\[
    G_i = A_i Q \in \mathbb{R}^{d\times k}, \qquad i=1,\dots,m.
\]
Assume the matrices $A_i$ satisfy~\Cref{assumption-1} and assume that
$\widehat{\Lambda}$ returned by \textsc{PrivRange}~(\Cref{algo:priv-range-max}) satisfies
\[
    \widehat{\Lambda}\ \ge\ \max_{r\in[k]} \lambda_1^2 \|H q_r\|_2.
\]
Let \textsc{TruncatedMean}~(\Cref{algo:trunc-mean}) be run with
\[
    R = 3K\sqrt{\widehat{\Lambda}}\log^a(mdk/\zeta)
    \qquad
    \text{and privacy parameters }(\epsilon_C,\delta_C).
\]
If the number of samples $m$ passed to \TruncatedMean satisfies
\[
    m \geq C \frac{dk}{\varepsilon_C} \log\Big(\frac{dk}{\zeta\delta_C}\Big)
\]
for a sufficiently large universal constant $C>0$, then with probability at least $1-\zeta$, for every $i\in[m]$, $j\in[d]$,
and $r\in[k]$,
\[
    (G_i)_{jr}\in
    \Big[
        \bar g_{jr}-R,\ \bar g_{jr}+R
    \Big].
\]
\end{lemma}

\begin{proof}
Fix a column $r\in[k]$. Since $q_r$ is a unit vector and is independent of the current mean batch,
the vector-valued gradients
\[
    g_i^{(r)} := A_i q_r \in \mathbb{R}^d
\]
satisfy the same one-vector assumptions as in the paper with $u=q_r$.
In particular, if we write
\[
    \mu^{(r)} := \mathbb{E}[g_i^{(r)}] = \Sigma q_r,
\]
then for every coordinate $j\in[d]$, the scalar random variable
$(g_i^{(r)})_j - \mu^{(r)}_j$ obeys the same sub-Weibull tail bound, namely
\[
    \Pr\!\left(
        |(g_i^{(r)})_j - \mu^{(r)}_j|
        >
        K\lambda_1 \sqrt{\|H q_r\|_2}\log^a(mdk/\zeta)
    \right)
    \le \frac{\zeta}{4mdk}.
\]
Since
\[
    \lambda_1^2 \|H q_r\|_2 \le \widehat{\Lambda},
\]
we obtain
\[
    \Pr\!\left(
        |(g_i^{(r)})_j - \mu^{(r)}_j|
        >
        K\sqrt{\widehat{\Lambda}}\log^a(mdk/\zeta)
    \right)
    \le \frac{\zeta}{4mdk}.
\]
A union bound over all triples $(i,j,r)\in[m]\times[d]\times[k]$ gives that with probability
at least $1-\zeta/4$,
\begin{equation}
\label{eq:event-E1}
    |(g_i^{(r)})_j - \mu^{(r)}_j|
    \le
    K\sqrt{\widehat{\Lambda}}\log^a(mdk/\zeta)
    \qquad
    \forall\, i\in[m],\, j\in[d],\, r\in[k].
\end{equation}

Next fix $(j,r)$, and write $L:=K\sqrt{\widehat\Lambda}\log^a(mdk/\zeta)$, so that on the event
\eqref{eq:event-E1} every sample lies in
\[
  I_{jr}:=\Big[\mu^{(r)}_j-L,\ \mu^{(r)}_j+L\Big].
\]
Consider the histogram used by \textsc{TruncatedMean} on the samples
$\{(G_i)_{jr}\}_{i=1}^m=\{(g^{(r)}_i)_j\}_{i=1}^m$, with bin width $\sqrt{\widehat\Lambda}$, run
with privacy budget $(\varepsilon_C/(dk),\delta_C/(dk))$. 
Because \emph{all} $m$ samples lie in $I_{jr}$, every non-empty bin intersects $I_{jr}$ and every
bin disjoint from $I_{jr}$ is empty; how many bins $I_{jr}$ spans is immaterial.

It remains to check that the release is not identically zero. Conditionally on the past the
variables $(G_i)_{jr}$, $i\in[m]$, are i.i.d.\ with mean $\mu^{(r)}_j$ and variance at most
$\lambda_1^2\|Hq_r\|_2\le\widehat\Lambda$, so Chebyshev's inequality gives
$\Pr\big(|(G_i)_{jr}-\mu^{(r)}_j|>2\sqrt{\widehat\Lambda}\big)\le\frac14$. The interval
$[\mu^{(r)}_j-2\sqrt{\widehat\Lambda},\,\mu^{(r)}_j+2\sqrt{\widehat\Lambda}]$ has length
$4\sqrt{\widehat\Lambda}$ and therefore meets at most five bins, one of which carries expected mass
at least $3/20$; by a Chernoff bound and a union bound over the $dk$ coordinates its empirical mass
is at least $\frac1{10}$ for every $(j,r)$ with probability at least $1-\zeta/8$. By
\Cref{lem:stability-hist} at accuracy $\beta<\frac1{20}$, applied with per-coordinate privacy budget
$(\varepsilon_C/(dk),\delta_C/(dk))$, which the stated lower bound on $m$ supplies, that bin is
released with positive count; since every empty bin is released as $0$, the selected bin is
non-empty and therefore intersects $I_{jr}$, for every coordinate, with probability at least
$1-\zeta/4$.

Since $\bar g_{jr}$ is the left endpoint of the selected bin and that bin meets $I_{jr}$ and has
width $\sqrt{\widehat\Lambda}$, it follows that
$|\bar g_{jr}-\mu^{(r)}_j|\le L+\sqrt{\widehat\Lambda}$, and hence, using $K\ge1$ and
$\log^a(mdk/\zeta)\ge1$,
\begin{equation}\label{eq:center-close}
  \big|\bar g_{jr}-\mu^{(r)}_j\big|\le 2K\sqrt{\widehat\Lambda}\log^a(mdk/\zeta)
\end{equation}
for all $(j,r)$ simultaneously with probability at least $1-\zeta/4$.
By basic composition over the $dk$ coordinate histograms, the released coordinate center
$(\bar g_{jr})_{j\in[d],r\in[k]}$ is $(\epsilon_C,\delta_C)$-DP.
This accounts only for the private center selection; the unnoised truncated average
returned by \textsc{TruncatedMean} is not a standalone DP mean release.

Finally, on the intersection of the events
\eqref{eq:event-E1} and \eqref{eq:center-close}, for every $i,j,r$,
\[
    |(G_i)_{jr} - \bar g_{jr}|
    \le
    |(G_i)_{jr} - \mu_j^{(r)}|
    +
    |\mu_j^{(r)} - \bar g_{jr}|
    \le
    3K\sqrt{\widehat{\Lambda}}\log^a(mdk/\zeta)
    = R.
\]
Thus
\[
    (G_i)_{jr}\in[\bar g_{jr}-R,\ \bar g_{jr}+R]
\]
for all $i,j,r$ simultaneously.
Hence no coordinate is clipped.
Let $\widetilde G_i$ be the coordinatewise truncated version of $G_i$.
The preceding display implies $\widetilde G_i=G_i=A_iQ$ for all $i$.
Since each $A_i$ is symmetric, $Q^\top A_iQ$ is symmetric, and hence
$A_iQ\in\mathcal S_Q$. Therefore the raw mean also belongs to
\(\mathcal S_Q\), so
\[
    \mathsf P_{\mathcal S_Q}\left(
    \frac1m\sum_{i=1}^m \widetilde G_i
    \right)
    =
    \frac1m\sum_{i=1}^m A_iQ.
\]
\end{proof}

\begin{algorithm}[ht]
\caption{\PrivRange}\label{algo:priv-range-max}
\textbf{Input:} $S = \{G_i\}_{i=1}^{B}\subset \reals^{d\times k}$,
privacy parameters $(\varepsilon,\delta)$, failure probability $\zeta$
\begin{algorithmic}[1]
\For{$i=1,2,\dots,\lfloor B/2 \rfloor$}
    \State Let $\widetilde G_i \gets G_{2i} - G_{2i-1} \in \mathbb{R}^{d\times k}$
\EndFor
\State Let $\widetilde S = \{\widetilde G_i\}_{i=1}^{\lfloor B/2 \rfloor}$
\State Let $m \gets C_1 \log(1/(\delta\zeta))/\varepsilon$
\State Partition $\widetilde S$ into $m$ subsets and denote each subset by $\mathcal{G}_j$, where each subset has size $b = \lfloor B/(2m)\rfloor$
\For{$j=1,\dots,m$}
    \For{$r=1,\dots,k$}
        \State Let $G_j^{(r)} \in \mathbb{R}^{d\times b}$ be the matrix whose columns are the $r$th columns of the matrices in $\mathcal{G}_j$
        \State Let
        \(
            \lambda_{j,r}
            \gets
            \lambda_1\!\left(\frac{1}{2b} G_j^{(r)} \bigl(G_j^{(r)}\bigr)^\top \right)
        \)
    \EndFor
    \State Let
    \(
        m_j \gets \max_{r\in[k]} \lambda_{j,r}
    \)
\EndFor
\State Partition $[0,\infty)$ into
\(
\Omega \gets \{\dots, [2^{-2/4},2^{-1/4}), [2^{-1/4},1), [1,2^{1/4}), [2^{1/4},2^{2/4}), \dots\} \cup \{[0,0]\}
\)
\State Run $(\varepsilon,\delta)$-DP histogram learner of~\Cref{lem:stability-hist} on $\{m_j\}_{j=1}^{m}$ over $\Omega$
\If{all the bins are empty}
    \State \Return $\perp$
\EndIf
\State Let $[l,r]$ be a non-empty bin that contains the maximum number of points in the DP histogram
\State \Return $\widehat{\Lambda} = 2l$
\end{algorithmic}
\end{algorithm}

\begin{algorithm}[ht]
\caption{\textsc{TruncatedMean}}\label{algo:trunc-mean}
\textbf{Input:} $S = \{G_1,\dots,G_B\}\subset \mathbb{R}^{d\times k}$, truncation threshold $R$, range estimate $\widehat{\Lambda}$, center privacy parameters $(\epsilon_C,\delta_C)$

\begin{algorithmic}[1]
    \For{$j=1,\dots,d$}
        \For{$r=1,\dots,k$}
            \State Run a private histogram learner on $\{(G_\ell)_{jr}\}_{\ell=1}^B$
            over bins of width $\sqrt{\widehat{\Lambda}}$ using privacy budget
            $(\epsilon_C/(dk),\delta_C/(dk))$
            \State Let $[\ell_{jr},h_{jr}]$ be the bin with the largest private count
            \State Set $\bar g_{jr}\gets \ell_{jr}$
        \EndFor
    \EndFor
    \For{$\ell=1,\dots,B$}
        \For{$j=1,\dots,d$}
            \For{$r=1,\dots,k$}
                \State Truncate $(G_\ell)_{jr}$ to
                \(
                    [\bar g_{jr}-R,\ \bar g_{jr}+R].
                \)
            \EndFor
        \EndFor
        \State Let $\widetilde G_\ell$ be the truncated matrix
    \EndFor
    \State \Return
    \(
        \widetilde \mu \gets \frac1B\sum_{\ell=1}^B \widetilde G_\ell
    \)
\end{algorithmic}
\end{algorithm}

\begin{theorem}[Privacy and accuracy of \PrivRange]
\label{thm:privrangemax}
\Cref{algo:priv-range-max} is \((\varepsilon,\delta)\)-DP. Let \(Q\in\mathbb{R}^{d\times k}\) have orthonormal columns \(q_1,\dots,q_k\), and suppose \(Q\) is fixed independently of the current batch \(\{A_\ell\}_{\ell=1}^B\). Define \(G_\ell = A_\ell Q \in \mathbb{R}^{d\times k}\), \(\ell=1,\dots,B\), and let \(\Lambda(Q) := \max_{r\in[k]} \lambda_1^2 \|Hq_r\|_2\).

Assume that the matrices \(A_\ell\) satisfy \Cref{ass:A1,ass:A4}. If the block size \(b = \left\lfloor \frac{B}{2m}\right\rfloor\) satisfies \(b \ge \tilde O\!\bigl(K^2 d \log(km/\zeta)\bigr)\), equivalently, \(B \ge \tilde O\!\left(\frac{K^2 d \log(k/(\delta\zeta))\log(1/(\delta\zeta))}{\varepsilon}\right)\),
up to the same logarithmic factors, then with probability at least \(1-\zeta\), \Cref{algo:priv-range-max} outputs \(\widehat{\Lambda}\) satisfying
\(\widehat{\Lambda}\in[\Lambda(Q),C_\Lambda\Lambda(Q)]\).
\end{theorem}

\begin{proof}
Privacy is immediate. The algorithm first transforms the batch \(\{G_\ell\}_{\ell=1}^B\) deterministically into the block maxima \(\{m_j\}_{j=1}^m\), and then applies the \((\varepsilon,\delta)\)-DP histogram learner~(\Cref{lem:stability-hist}). Hence \Cref{algo:priv-range-max} is \((\varepsilon,\delta)\)-DP.

We now prove the accuracy guarantee. For each column \(r\in[k]\), define
\[
g_\ell^{(r)} := G_\ell e_r = A_\ell q_r \in \mathbb{R}^d.
\]
Since \(Q\) is fixed independently of the current batch, each \(q_r\) is fixed and independent of the samples \(\{A_\ell\}_{\ell=1}^B\). Therefore the vectors \(\{g_\ell^{(r)}\}\) are in the setting of Theorem~6.1 of~\citet{liu2022dp}.

For each block \(j\in[m]\) and column \(r\in[k]\), let \(G_j^{(r)}\in\mathbb{R}^{d\times b}\) be the matrix whose columns are the \(r\)-th columns of the paired differences in \(\mathcal G_j\), and define
\[
\lambda_{j,r}
=
\lambda_1\!\left(\frac{1}{2b}G_j^{(r)}(G_j^{(r)})^\top\right).
\]
The columns of \(G_j^{(r)}\) are pair differences, whose covariance is twice
the target covariance. Thus the factor \(1/(2b)\) normalizes the block
covariance to the original target scale.
By the concentration result underlying Theorem~6.1 of~\citet{liu2022dp}, provided
\[
b \ge \tilde O\!\bigl(K^2 d \log(km/\zeta)\bigr),
\]
we have with probability at least \(1-\zeta/(10m)\) that, simultaneously for all \(r\in[k]\),
\[
\lambda_{j,r}
\in
\left[
\frac{1}{\sqrt{2}}\,\lambda_1^2\|Hq_r\|_2,\;
\sqrt{2}\,\lambda_1^2\|Hq_r\|_2
\right].
\]
On this event,
\[
m_j := \max_{r\in[k]} \lambda_{j,r}
\in
\left[
\frac{1}{\sqrt{2}}\,\Lambda(Q),\;
\sqrt{2}\,\Lambda(Q)
\right].
\]

Taking a union bound over the $m$ blocks, with probability at least $1-\zeta/10$ \emph{every}
$m_j$ lies in
\[
  I:=\Big[\frac1{\sqrt2}\Lambda(Q),\ \sqrt2\,\Lambda(Q)\Big].
\]
The endpoints of $I$ have ratio $2$ while the bins of $\Omega$ have ratio $2^{1/4}$, so $I$ spans
$\log_{2^{1/4}}2=4$ bin widths and intersects at most five consecutive bins; and because every
$m_j$ lies in $I$, every non-empty bin intersects $I$, and every bin disjoint from $I$ is empty.
At least one of the intersecting bins therefore holds at least a $1/5$-fraction of the $m_j$. By
\Cref{lem:stability-hist} at accuracy $\beta<\frac1{10}$, which the choice
$m=C_1\log(1/(\delta\zeta))/\varepsilon$ supplies, the DP histogram learner returns, with
probability at least $1-\zeta/2$, a non-empty bin $[l,2^{1/4}l)$ intersecting $I$.

Consequently $l<\sqrt2\,\Lambda(Q)$, while $2^{1/4}l>\frac1{\sqrt2}\Lambda(Q)$ gives
$l>2^{-3/4}\Lambda(Q)>\frac12\Lambda(Q)$, so the output
\[
  \widehat\Lambda=2l
\]
satisfies
\[
  \widehat\Lambda\in\big[\Lambda(Q),\,C_\Lambda\Lambda(Q)\big],\qquad C_\Lambda=2\sqrt2 .
\]
Combining the above events proves the theorem.
\end{proof}

\begin{lemma}[Stability-based histogram {\cite[Lemma 2.3]{karwa2017finite}}]\label{lem:stability-hist}
For every \(K \in \mathbb{N}\cup\{\infty\}\), domain \(\Omega\), every collection of disjoint bins
\(B_1,\dots,B_K\) defined on \(\Omega\), \(n\in\mathbb{N}\), \(\varepsilon \ge 0\), \(\delta \in (0,1/n)\), \(\beta > 0\), and \(\alpha \in (0,1)\), under the replacement notion of neighboring datasets there exists an \((\varepsilon,\delta)\)-differentially private algorithm
\[
\mathcal{M}:\Omega^n \to \mathbb{R}^K
\]
such that for any dataset \(X_1,\dots,X_n \in \Omega^n\),
\begin{enumerate}
    \item
    \[
    \hat p_k = \frac{1}{n}\sum_{X_i \in B_k} 1,
    \]
    \item
    \[
    (\tilde p_1,\dots,\tilde p_K) \leftarrow \mathcal{M}(X_1,\dots,X_n),
    \]
    and
    \item if
    \[
    n \ge \min\left\{
    \frac{8}{\varepsilon\beta}\log\!\left(\frac{2K}{\alpha}\right),
    \frac{8}{\varepsilon\beta}\log\!\left(\frac{4}{\alpha\delta}\right)
    \right\},
    \]
    then
    \[
    \mathbb{P}\bigl(|\tilde p_k-\hat p_k|\le \beta\bigr)\ge 1-\alpha.
    \]
    \item and \(\hat p_k = 0\) implies \(\tilde p_k = 0\); that is, every bin released with a
    positive count contains at least one sample.
\end{enumerate}
\end{lemma}

\corrankkadaptiveojas*

\begin{proof}
Let
\[
    S_i\coloneqq A_i-\Sigma .
\]
First,
\[
    A_i
    =
    (V\Lambda^{1/2}+\sigma Z_i)
    (V\Lambda^{1/2}+\sigma Z_i)^\top .
\]
Expanding and taking expectations gives
\[
\begin{aligned}
    \mathbb E[A_i]
    &=
    V\Lambda V^\top
    +
    \sigma V\Lambda^{1/2}\mathbb E[Z_i^\top]
    +
    \sigma \mathbb E[Z_i]\Lambda^{1/2}V^\top
    +
    \sigma^2\mathbb E[Z_iZ_i^\top].
\end{aligned}
\]
The two cross terms vanish because \(\mathbb E[Z_i]=0\). Since
\(Z_i\in\mathbb R^{d\times k}\) has i.i.d. standard Gaussian entries,
\[
    \mathbb E[Z_iZ_i^\top]=kI_d .
\]
Therefore
\[
    \Sigma
    =
    \mathbb E[A_i]
    =
    V\Lambda V^\top+k\sigma^2I_d .
\]

The eigenvalues of \(\Sigma\) are
\[
    \mu_j=\lambda_j+k\sigma^2,\qquad j\le k,
\]
and
\[
    \mu_{k+1}=\cdots=\mu_d=k\sigma^2 .
\]
Hence
\[
    \mu_1=\lambda_1+k\sigma^2,
    \qquad
    \mu_k-\mu_{k+1}=\lambda_k .
\]

Next we control the variance parameter in Assumption~\ref{ass:A3}. Write
\[
    B\coloneqq V\Lambda V^\top .
\]
Then
\[
\begin{aligned}
    S_i
    &=
    A_i-\Sigma                                                \\
    &=
    \sigma V\Lambda^{1/2}Z_i^\top
    +
    \sigma Z_i\Lambda^{1/2}V^\top
    +
    \sigma^2(Z_iZ_i^\top-kI_d).
\end{aligned}
\]
Define
\[
    C_i
    \coloneqq
    V\Lambda^{1/2}Z_i^\top
    +
    Z_i\Lambda^{1/2}V^\top,
    \qquad
    R_i
    \coloneqq
    Z_iZ_i^\top-kI_d .
\]
Thus
\[
    S_i=\sigma C_i+\sigma^2R_i .
\]
Since \((A+B)^2\preceq 2A^2+2B^2\) for symmetric matrices,
\[
    \mathbb E[S_i^2]
    \preceq
    2\sigma^2\mathbb E[C_i^2]
    +
    2\sigma^4\mathbb E[R_i^2].
\]

We now compute the two terms. By Gaussian moment calculations,
\[
    \mathbb E[C_i^2]
    =
    (d+2)B+\operatorname{tr}(\Lambda)I_d .
\]
Therefore
\[
    \|\mathbb E[C_i^2]\|_2
    \le
    (d+2)\lambda_1+\operatorname{tr}(\Lambda).
\]
Also, since \(Z_iZ_i^\top\) is a Wishart matrix with \(k\) degrees of
freedom,
\[
    \mathbb E[(Z_iZ_i^\top-kI_d)^2]
    =
    k(d+1)I_d .
\]
Hence
\[
    \|\mathbb E[R_i^2]\|_2
    =
    k(d+1).
\]
Combining the previous displays,
\[
\begin{aligned}
    \|\mathbb E[S_i^2]\|_2
    &\le
    2\sigma^2
    \left(
        (d+2)\lambda_1+\operatorname{tr}(\Lambda)
    \right)
    +
    2\sigma^4k(d+1)                                      \\
    &\lesssim
    d\sigma^2(\lambda_1+k\sigma^2).
\end{aligned}
\]
Since \(\mu_1=\lambda_1+k\sigma^2\), Assumption~\ref{ass:A3} holds with
\[
    V_{\mathrm{thm}}
    \lesssim
    \frac{
        d\sigma^2(\lambda_1+k\sigma^2)
    }{
        (\lambda_1+k\sigma^2)^2
    }
    =
    \frac{d\sigma^2}{\lambda_1+k\sigma^2}.
\]

Next we bound \(\gamma\). For any unit vector \(u\),
\[
    S_iu
    =
    \sigma C_iu+\sigma^2R_iu .
\]
Again using \((x+y)(x+y)^\top\preceq 2xx^\top+2yy^\top\), we have
\[
    \mathbb E[S_i uu^\top S_i]
    \preceq
    2\sigma^2\mathbb E[C_i uu^\top C_i]
    +
    2\sigma^4\mathbb E[R_i uu^\top R_i].
\]
Now
\[
    C_i u
    =
    V\Lambda^{1/2}Z_i^\top u
    +
    Z_i\Lambda^{1/2}V^\top u .
\]
Since \(Z_i^\top u\sim\mathcal N(0,I_k)\),
\[
    \mathbb E[
        V\Lambda^{1/2}Z_i^\top u u^\top Z_i\Lambda^{1/2}V^\top
    ]
    =
    V\Lambda V^\top
    =
    B .
\]
Also, letting \(a=\Lambda^{1/2}V^\top u\),
\[
    \mathbb E[Z_i aa^\top Z_i^\top]
    =
    \|a\|_2^2 I_d
    \le
    \lambda_1 I_d .
\]
Therefore
\[
    \|\mathbb E[C_i uu^\top C_i]\|_2
    \lesssim
    \lambda_1 .
\]
For the Wishart fluctuation term, another standard Gaussian moment calculation
gives
\[
    \mathbb E[
        (Z_iZ_i^\top-kI_d)uu^\top(Z_iZ_i^\top-kI_d)
    ]
    =
    k(I_d+uu^\top),
\]
and hence
\[
    \|\mathbb E[R_i uu^\top R_i]\|_2
    \le
    2k .
\]
Thus
\[
    \left\|
        \mathbb E[S_i uu^\top S_i]
    \right\|_2
    \lesssim
    \sigma^2\lambda_1+k\sigma^4
    =
    \sigma^2(\lambda_1+k\sigma^2).
\]
By the definition of \(\gamma\),
\[
    \gamma^2
    =
    \max_{\|u\|=1}
    \frac{
        \left\|
            \mathbb E[S_i uu^\top S_i]
        \right\|_2
    }{
        \mu_1^2
    },
\]
so
\[
    \gamma^2
    \lesssim
    \frac{
        \sigma^2(\lambda_1+k\sigma^2)
    }{
        (\lambda_1+k\sigma^2)^2
    }
    =
    \frac{\sigma^2}{\lambda_1+k\sigma^2}.
\]
Equivalently,
\[
    \mu_1\gamma
    \lesssim
    \sigma\sqrt{\lambda_1+k\sigma^2}.
\]

It remains to verify the boundedness parameter on
\(\mathcal E_{\mathrm{bd}}\). The standard Gaussian operator norm bound gives
\[
    \mathbb P\left(
        \|Z_i\|_2
        \le
        \sqrt d+\sqrt k+t
    \right)
    \ge
    1-e^{-t^2/2}.
\]
Taking \(t=2\sqrt{\log n}\) and union bounding over \(i\in[n]\) yields
\[
    \mathbb P(\mathcal E_{\mathrm{bd}})
    \ge
    1-\frac1n .
\]
On this event,
\[
    \|X_i\|_2
    \le
    \sqrt{\lambda_1}
    +
    \sigma(\sqrt d+\sqrt k+2\sqrt{\log n}).
\]
Therefore
\[
\begin{aligned}
    \|A_i-\Sigma\|_2
    &\le
    \|A_i\|_2+\|\Sigma\|_2                                  \\
    &=
    \|X_i\|_2^2+\mu_1                                        \\
    &\le
    \left(
        \sqrt{\lambda_1}
        +
        \sigma(\sqrt d+\sqrt k+2\sqrt{\log n})
    \right)^2
    +
    \lambda_1+k\sigma^2 .
\end{aligned}
\]
Using
\[
    \|P^\top(A_i-\Sigma)\|_{(2,k)}
    \le
    \sqrt k\,\|A_i-\Sigma\|_2 ,
\]
we may take
\[
    M
    =
    \frac{\sqrt k}{\lambda_1}
    \left[
        \left(
            \sqrt{\lambda_1}
            +
            \sigma(\sqrt d+\sqrt k+2\sqrt{\log n})
        \right)^2
        +
        \lambda_1+k\sigma^2
    \right].
\]

We now apply \Cref{thm:utility-priv-ojas-adaptive}. Its first term is
\[
        \kappa\sqrt{\frac{kV_{\mathrm{thm}}}{n}},
    \qquad \kappa=\frac{\mu_1}{\mu_k-\mu_{k+1}}.
\]
Using
\[
    \mu_1=\lambda_1+k\sigma^2,
    \qquad
    \mu_k-\mu_{k+1}=\lambda_k,
\]
and
\[
    V_{\mathrm{thm}}
    \lesssim
    \frac{d\sigma^2}{\lambda_1+k\sigma^2},
\]
we obtain
\[
\begin{aligned}
    \frac{\mu_1\sqrt{kV_{\mathrm{thm}}}}
    {(\mu_k-\mu_{k+1})\sqrt n}
    &\lesssim
    \frac{
        (\lambda_1+k\sigma^2)
        \sqrt{
            k\frac{d\sigma^2}{\lambda_1+k\sigma^2}
        }
    }{
        \lambda_k\sqrt n
    }                                                        \\
    &=
    \frac{
        \sigma\sqrt{kd(\lambda_1+k\sigma^2)}
    }{
        \lambda_k\sqrt n
    }.
\end{aligned}
\]
The boundedness term becomes
\[
    \frac{\sqrt k M}{\lambda_k n}
    =
    \frac{
        k\left[
            \left(
                \sqrt{\lambda_1}
                +
                \sigma(\sqrt d+\sqrt k+2\sqrt{\log n})
            \right)^2
            +
            \lambda_1+k\sigma^2
        \right]
    }{
        \lambda_k n
    }.
\]
Finally, the privacy term is
\[
     \kappa^{3/2}\,\frac{Kdk\gamma}{\varepsilon n}.
\]
Using \(\mu_1\gamma\lesssim\sigma\sqrt{\lambda_1+k\sigma^2}\) and
\(\mu_k-\mu_{k+1}=\lambda_k\), this is bounded by
\[
    \sqrt{\kappa}\,\frac{
        Kdk\,\sigma\sqrt{\lambda_1+k\sigma^2}
    }{
        \varepsilon\lambda_k n
    }.
\]
Combining these three bounds proves the claim.
\end{proof}

\begin{restatable}[Gaussian spiked covariance,~\Cref{algo:priv-ojas-adaptive}]{corollary}{corgaussianspikedadaptiveojas}
\label{cor:gaussian-spiked-adaptive-ojas} With input matrices $A_i = x_i x_i^\top$ with $x_i$ defined as in~\Cref{spiked-cov-lower-bound}~\Cref{algo:priv-ojas-adaptive} gives
\[
    \|Q_TQ_T^\top-V_kV_k^\top\|_F
    \le
    \widetilde O\left(
        \kappa\sqrt{\frac{dk}{n}}
        +
        \kappa^{3/2}\,\frac{Kdk}{\varepsilon n}
    \right).
\]
\end{restatable}

\begin{proof}
Since \(x_i\sim\mathcal N(0,\Sigma)\), we have
\[
    \mathbb E[A_i]
    =
    \mathbb E[x_ix_i^\top]
    =
    \Sigma .
\]
The eigenvectors corresponding to the top \(k\) eigenvalues of \(\Sigma\) are
the columns of \(U\). Moreover,
\[
    \lambda_j=\theta_j+\sigma^2,\qquad j\le k,
    \qquad
    \lambda_{k+1}=\cdots=\lambda_d=\sigma^2,
\]
so the target eigenspace is \(\operatorname{span}(U)\), and the eigengap is
\[
    \Delta_k
    =
    \lambda_k-\lambda_{k+1}
    =
    \theta_k .
\]

Let
\[
    S_i\coloneqq A_i-\Sigma=x_ix_i^\top-\Sigma .
\]
We first compute the variance parameter appearing in
Assumption~\ref{ass:A3}. Since \(x_i\) is Gaussian,
\[
    \mathbb E\left[\|x_i\|_2^2 x_ix_i^\top\right]
    =
    \operatorname{tr}(\Sigma)\Sigma+2\Sigma^2 .
\]
Therefore
\[
\begin{aligned}
    \mathbb E[S_i^2]
    &=
    \mathbb E[(x_ix_i^\top-\Sigma)^2]                         \\
    &=
    \mathbb E[\|x_i\|_2^2x_ix_i^\top]-\Sigma^2                 \\
    &=
    \operatorname{tr}(\Sigma)\Sigma+\Sigma^2 .
\end{aligned}
\]
Hence
\[
    \left\|\mathbb E[S_i^2]\right\|_2
    =
    \lambda_1\operatorname{tr}(\Sigma)+\lambda_1^2
    =
    \lambda_1
    \left(
        \lambda_1+\sum_{j=1}^d\lambda_j
    \right).
\]
Thus Assumption~\ref{ass:A3} holds with
\[
    V_{\mathrm{thm}}
    =
    \frac{
        \lambda_1
        \left(
            \lambda_1+\sum_{j=1}^d\lambda_j
        \right)
    }{
        \lambda_1^2
    }
    =
    \frac{
        \lambda_1+\sum_{j=1}^d\lambda_j
    }{
        \lambda_1
    }.
\]

Next we bound \(\gamma\). For any unit vector \(u\),
\[
\begin{aligned}
    \mathbb E[S_i uu^\top S_i]
    &=
    \mathbb E[
        (x_ix_i^\top-\Sigma)uu^\top(x_ix_i^\top-\Sigma)
    ]                                                        \\
    &=
    \mathbb E[
        (u^\top x_i)^2x_ix_i^\top
    ]
    -
    \Sigma uu^\top\Sigma .
\end{aligned}
\]
Using the Gaussian fourth-moment identity,
\[
    \mathbb E[
        (u^\top x_i)^2x_ix_i^\top
    ]
    =
    (u^\top\Sigma u)\Sigma
    +
    2\Sigma uu^\top\Sigma .
\]
Therefore
\[
    \mathbb E[S_i uu^\top S_i]
    =
    (u^\top\Sigma u)\Sigma+\Sigma uu^\top\Sigma .
\]
Taking operator norms gives
\[
\begin{aligned}
    \left\|
        \mathbb E[S_i uu^\top S_i]
    \right\|_2
    &\le
    (u^\top\Sigma u)\|\Sigma\|_2
    +
    \|\Sigma u\|_2^2                                      \\
    &\le
    \lambda_1^2+\lambda_1^2
    =
    2\lambda_1^2 .
\end{aligned}
\]
Hence
\[
    \gamma^2
    =
    \max_{\|u\|=1}
    \frac{
        \left\|
            \mathbb E[S_i uu^\top S_i]
        \right\|_2
    }{
        \lambda_1^2
    }
    \le 2 .
\]
For Gaussian quadratic forms, Assumption~\ref{ass:A4} holds with \(a=1\) and
an absolute constant \(K\). We absorb this absolute constant into the
\(\widetilde O(\cdot)\) notation.

It remains to verify the boundedness parameter on a high-probability event.
Write \(x_i=\Sigma^{1/2}g_i\), where \(g_i\sim\mathcal N(0,I_d)\). Then
\[
    \|x_i\|_2\le \sqrt{\lambda_1}\|g_i\|_2 .
\]
By the standard Gaussian norm bound,
\[
    \mathbb P\left(
        \|g_i\|_2\le \sqrt d+t
    \right)
    \ge
    1-e^{-t^2/2}.
\]
Taking \(t=2\sqrt{\log n}\) and applying a union bound over
\(i\in[n]\), we get
\[
    \mathbb P(\mathcal E_{\mathrm{bd}})
    \ge
    1-\frac1n .
\]
On \(\mathcal E_{\mathrm{bd}}\),
\[
    \|A_i-\Sigma\|_2
    \le
    \|x_ix_i^\top\|_2+\|\Sigma\|_2
    =
    \|x_i\|_2^2+\lambda_1
    \le
    \lambda_1
    \left[
        \bigl(\sqrt d+2\sqrt{\log n}\bigr)^2+1
    \right].
\]
Since
\[
    \|P^\top(A_i-\Sigma)\|_{(2,k)}
    \le
    \sqrt k\,\|A_i-\Sigma\|_2 ,
\]
Assumption~\ref{ass:A2} holds on \(\mathcal E_{\mathrm{bd}}\) with
\[
    M
    =
    \sqrt k
    \left[
        \bigl(\sqrt d+2\sqrt{\log n}\bigr)^2+1
    \right].
\]

We now apply \Cref{thm:utility-priv-ojas-adaptive}. Its first term becomes
\[
\begin{aligned}
    \kappa\sqrt{\frac{kV_{\mathrm{thm}}}{n}}
    &=
    \frac{
        \lambda_1
        \sqrt{
            k
            \frac{
                \lambda_1+\sum_{j=1}^d\lambda_j
            }{
                \lambda_1
            }
        }
    }{
        \Delta_k\sqrt n
    }                                                       \\
    &=
    \frac{
        \sqrt{
            k\lambda_1
            \left(
                \lambda_1+\sum_{j=1}^d\lambda_j
            \right)
        }
    }{
        \Delta_k\sqrt n
    } .
\end{aligned}
\]
The boundedness term becomes
\[
    \frac{\sqrt k M}{\Delta_k n}
    =
    \frac{
        k\lambda_1
        \left[
            \bigl(\sqrt d+2\sqrt{\log n}\bigr)^2+1
        \right]
    }{
        \Delta_k n
    }.
\]
Finally, since \(\gamma\le\sqrt 2\) and \(K\) is an absolute constant for
Gaussian quadratic forms, the privacy term satisfies
\[
    \frac{\lambda_1Kdk\gamma}
    {\varepsilon\Delta_k n}
    \lesssim
    \frac{
        dk\lambda_1
    }{
        \varepsilon\Delta_k n
    }.
\]
Since $\varepsilon \le 1$ and $K = O(1)$, the boundedness term is dominated by the privacy term.
Combining these three bounds proves the result.
\end{proof}

\input{appendices/tangent_proof}

%% file: appendices/tangent_proof.tex
\subsection{Proofs for the tangent refinement theorem}\label{app:tangent}

\begin{restatable}[Utility of \TangAlg]
{theorem}{tangentgeneralutility}
\label{thm:tangent-general-utility}
Suppose \(A_1,\ldots,A_n\) satisfy~\Cref{assumption-1}. Suppose further
that, together with the starting point \(Q_0\), they satisfy
\Cref{ass:tangent-model}, and that
\[
    \norm{Q_0Q_0^\top-V_kV_k^\top}_F\le c_0/2
\]
with probability at least \(1-\zeta/8\). Set
\[
    B = \lfloor n/T\rfloor,
    \qquad
    T=\left\lceil
        C_T\log\!\left(
        \frac{ndk}{\zeta\delta\epsilon}
        \right)
    \right\rceil .
\]
If
\[
    n
    \ge
    \softOm{
    \max\left\{
        \frac{K^2d+Kdk}{\epsilon},
        \frac{k(v_0^2+v_1^2)}{\Delta^2},
        \frac{\sqrt{k}M_\perp}{\Delta},
        \frac{K(g_0+g_1)dk}{\epsilon\Delta}
    \right\}
    },
\]
then there exists a choice of learning rates \(\{\eta_t\}_{t=1}^T\) such
that, with probability at least \(1-\zeta\), \Cref{alg:tangent-adaptive}
outputs \(Q_T\) satisfying
\[
\norm{Q_TQ_T^\top-V_kV_k^\top}_F
\le
\operatorname{Opt}_T
+
\softO{
    \frac{\sqrt{k}v_0}{\Delta\sqrt n}
    +
    \frac{\sqrt{k}M_\perp}{\Delta n}
    +
    \frac{K g_0dk}{\Delta\epsilon n}
},
\]
where
\(
    \operatorname{Opt}_T
    :=
    \exp\!\left(
        -c\Delta\sum_{t=1}^T\eta_t
    \right)
    \norm{Q_0Q_0^\top-V_kV_k^\top}_F .
\)
\end{restatable}

\begin{restatable}[Privacy of \TangAlg]{lemma}{tangentprivacymain}
\label{thm:tangent-privacy}
Suppose the warm-start samples and refinement samples are disjoint. If the
warm-start algorithm is \((\epsilon,\delta)\)-DP and each refinement iteration
uses disjoint batches with the budget allocation in
\Cref{alg:tangent-adaptive}, then the full two-stage algorithm is
\((\epsilon,\delta)\)-DP. If the warm-start and refinement phases reuse
records, their privacy costs compose sequentially.
\end{restatable}

\begin{lemma}[Privacy of projected coordinatewise tangent mean]
\label{lem:privacy-projected-tangent-mean}
Fix \(Q\in\bS_{d,k}\), \(\Pi=I-QQ^\top\), a coordinate center
\(C\in\reals^{d\times k}\), and a coordinatewise radius \(r>0\).  For a batch
\(S=(A_1,\ldots,A_m)\), let \(\widetilde Y_i(C)\) be \(\Pi A_iQ\) after
coordinatewise truncation to \([C_{jr}-r,C_{jr}+r]\), and define
\[
    h_C(S)
    :=
    \Pi\left(\frac1m\sum_{i=1}^m\widetilde Y_i(C)\right).
\]
Let \(Z\in\reals^{d\times k}\) have i.i.d. entries \(\cN(0,\sigma^2)\), and release
\[
    M_C(S)=h_C(S)+\Pi Z.
\]
If
\[
    \sigma\ge
    \frac{2r\sqrt{dk}}{m\epsilon_M}
    \sqrt{2\log(1.25/\delta_M)},
\]
then \(M_C\) is \((\epsilon_M,\delta_M)\)-DP conditional on \(C\).
\end{lemma}

\begin{proof}
By construction, \(h_C(S)\) lies in the tangent subspace
\[
    \mathcal T_Q\coloneqq\bc{Y\in\reals^{d\times k}:Q^\top Y=0},
\]
because \(Q^\top\Pi=0\).  If neighboring batches differ in one entry, then
\[
\norm{h_C(S)-h_C(S')}_F
\le
\frac{2r\sqrt{dk}}{m},
\]
because each coordinatewise-truncated matrix lies in a coordinate box of radius
\(r\), and \(\Pi\) is non-expansive in Frobenius norm. The random matrix \(\Pi Z\)
is a centered isotropic Gaussian on \(\mathcal T_Q\): if \(Q_\perp\) is an
orthonormal basis for \(\operatorname{span}(Q)^\perp\), then
\(\Pi Z=Q_\perp(Q_\perp^\top Z)\), and \(Q_\perp^\top Z\) has i.i.d.
\(\cN(0,\sigma^2)\) entries.  Thus this is the standard Gaussian mechanism in
the Euclidean space \((\mathcal T_Q,\norm{\cdot}_F)\) with sensitivity
\[
    \Delta_t
    \le
    \frac{2r\sqrt{dk}}{m},
\]
which proves the claim.
\end{proof}

\begin{proof}[Proof of \Cref{thm:tangent-privacy}]
Fix an iteration \(t\), and condition on all previous outputs. Then
\(Q=Q_{t-1}\) and \(\Pi=I-QQ^\top\) are fixed.

The range half is handled by \PrivRange and is
\((\epsilon_R,\delta_R)\)-DP. It is disjoint from the mean half.

Now consider the mean half. Let \(C\) be the coordinatewise center computed
internally by \TruncatedMean. Since \TruncatedMean runs \(dk\) coordinate
histograms, each with privacy budget
\((\epsilon_C/(dk),\delta_C/(dk))\), the center \(C\) is
\((\epsilon_C,\delta_C)\)-DP by basic composition.

Conditional on any fixed value of \(C\), the projected coordinatewise
truncated mean plus \(\Pi Z_t\) is \((\epsilon_M,\delta_M)\)-DP by
\Cref{lem:privacy-projected-tangent-mean}, using the Gaussian calibration in
\Cref{alg:tangent-adaptive}. By adaptive composition,
\[
    (C,\bar Y_t+W_t)
\]
is \((\epsilon_C+\epsilon_M,\delta_C+\delta_M)\)-DP on the mean half.
The algorithm releases only \(\bar Y_t+W_t\), so hiding \(C\) is
post-processing. Since \Cref{alg:tangent-adaptive} sets
\((\epsilon_C,\delta_C)=(\epsilon_M,\delta_M)=(\epsilon/2,\delta/2)\),
the mean-half mechanism is \((\epsilon,\delta)\)-DP.

The range and mean halves are disjoint, so one update is
\((\epsilon,\delta)\)-DP by parallel composition. The refinement batches are
disjoint across \(t\), so the full refinement phase is
\((\epsilon,\delta)\)-DP. Finally, because the warm-start and refinement
samples are disjoint, the two-stage algorithm is \((\epsilon,\delta)\)-DP by
parallel composition.
\end{proof}

\begin{proof}[Proof of \Cref{thm:tangent-general-utility}]
Let \(e_t:=\norm{Q_tQ_t^\top-P_\star}_F.\) We first collect the
high-probability events.  By \Cref{lem:tangent-range-accuracy} and a
union bound over \(t\in[T]\),
\[
    g_{Q_{t-1}}^2
    \le
    \widehat\Lambda_t
    \le
    C_\Lambda g_{Q_{t-1}}^2
\]
holds for all \(t\) with probability at least \(1-\zeta/8\).  On this event,
\Cref{lem:tangent-no-clipping-projected} implies that no coordinate in any mean
batch is clipped.  Thus, writing \(Q=Q_{t-1}\) and \(\Pi=I-QQ^\top\), on the
no-clipping event,
\[
    \mu_t
    =
    \frac1m\sum_{i\in\mathcal B_t^{\rm mean}}Y_i
    =
    \frac1m\sum_{i\in\mathcal B_t^{\rm mean}}\Pi A_iQ .
\]
Since each \(Y_i\) lies in the tangent space, \(\Pi\mu_t=\mu_t\). Therefore
the update has the form
\[
    Q_t
    =
    \mathrm{QR}\left(
    Q+
    \eta_t\{\Pi\Sigma Q+E_t\}
    \right),
\]
where
\[
    E_t
    =
    \frac1m\sum_{i\in\mathcal B_t^{\rm mean}}\Pi(A_i-\Sigma)Q+W_t.
\]
This is exactly the form needed for the local tangent contraction lemma.
By \Cref{lem:tangent-perturbation-bound}, again union bounded over \(t\),
\[
    \norm{E_t}_F
    \le
    \softO{
        \sqrt{k}v_{Q_{t-1}}m^{-1/2}
        +
        \sqrt{k}M_\perp m^{-1}
        +
        \frac{K g_{Q_{t-1}}dk}{m\epsilon}
    }
\]
simultaneously for all \(t\), with probability at least \(1-\zeta/4\).  Using
\Cref{ass:T3,ass:T4}, this gives
\[
    \norm{E_t}_F
    \le
    A+D e_{t-1},
\]
where
\[
    A
    :=
    \softO{
        \sqrt{k}v_0m^{-1/2}
        +
        \sqrt{k}M_\perp m^{-1}
        +
        \frac{K g_0dk}{m\epsilon}
    },
    \qquad
    D
    :=
    \softO{
        \sqrt{k}v_1m^{-1/2}
        +
        \frac{K g_1dk}{m\epsilon}
    }.
\]

For every iterate that remains in the local basin,
\Cref{lem:local-tangent-contraction} gives
\[
    e_t
    \le
    (1-c\eta_t\Delta)e_{t-1}+C\eta_t\norm{E_t}_F.
\]
The stated lower bound on \(n\), together with \(m\asymp n/T\) and
\(T\asymp\kappa_\perp\log(ndk/\zeta\delta\epsilon)\), implies \(D\le c\Delta\) for a small enough
universal constant: the batch is a factor \(T\) smaller than \(n\), so every branch of that lower
bound controlling \(A\) or \(D\) picks up \(\kappa_\perp\). Hence
\[
    e_t
    \le
    (1-c'\eta_t\Delta)e_{t-1}+C\eta_t A.
\]
Unrolling this recursion yields
\[
    e_T
    \le
    \exp\!\left(-c'\Delta\sum_{t=1}^T\eta_t\right)e_0
    +
    \frac{C A}{\Delta}.
\]
Since \(m\asymp n/T\) with \(T\asymp\kappa_\perp\log(ndk/\zeta\delta\epsilon)\), and \(A\) carries
\(m^{-1/2}\) in its first term and \(m^{-1}\) in the other two,
\[
    \frac{A}{\Delta}
    =
    \softO{
        \sqrt{\kappa_\perp}\,\frac{\sqrt{k}v_0}{\Delta\sqrt n}
        +
        \kappa_\perp\,\frac{\sqrt{k}M_\perp}{\Delta n}
        +
        \kappa_\perp\,\frac{K g_0dk}{\Delta\epsilon n}
    }.
\]
This gives the claimed bound after adjusting constants in
\(\operatorname{Opt}_T\).

It remains only to justify that the local condition is valid throughout the
proof. The sample-size lower bound also makes \(A/\Delta\le c_0\) after reducing the
universal constants.  Since the initializer satisfies
\(\norm{Q_0Q_0^\top-P_\star}_F\le c_0/2\), induction gives
\(\norm{Q_tQ_t^\top-P_\star}_F\le c_0\) for every \(t\le T\).  Combining the
initializer event with the range, center, and perturbation events gives
probability at least \(1-\zeta\).
\end{proof}

\begin{lemma}[Accuracy of \PrivRange on tangent samples]
\label{lem:tangent-range-accuracy}
Fix \(Q\in\bS_{d,k}\) independently of the current range batch, and let \(Y_i=\Pi_QA_iQ\).  Suppose \Cref{ass:T5} holds at \(Q\).  If the block size in \Cref{algo:priv-range-max} satisfies
\[
    b\ge \softOm{K^2d\log(km_0/\alpha}),
\]
then with probability at least \(1-\alpha\), \PrivRange outputs \(\widehat\Lambda\) satisfying
\[
    g_Q^2\le \widehat\Lambda\le C_\Lambda g_Q^2.
\]
\end{lemma}
\begin{proof}
The privacy argument is part of \Cref{thm:tangent-privacy}; here we
prove accuracy.  
Here \(\PrivRange\) is applied to the tangent samples \(Y_i=\Pi_QA_iQ\).
We now prove the accuracy guarantee.  Fix the current iterate \(Q\),
and condition on all previous batches.  Then \(Q\) is deterministic
and independent of the samples used by \PrivRange.
Let \( P=QQ^\top, \Pi_Q=I-P,\) and write the columns of \(Q\) as
\(q_1,\dots,q_k\). For each column \(r\in[k]\), define the tangent
covariance
\[
    C_r(Q)
    :=
    \Pi_Q
    \mathbb E\!\left[
        (A-\Sigma)q_rq_r^\top(A-\Sigma)^\top
    \right]
    \Pi_Q ,
\]
and define the corresponding tangent scale
\[
    g_Q^2
    :=
    \max_{r\in[k]}\|C_r(Q)\|_2 .
\]

For a paired sample, define \(\widetilde Y_s^{(r)} :=
    \Pi_Q(A_{2s}-A_{2s-1})q_r .  \) Then, \[
\begin{aligned}
    \mathbb E\!\left[
        \widetilde Y_s^{(r)}
        \bigl(\widetilde Y_s^{(r)}\bigr)^\top
    \right]
    &=
    \mathbb E\!\left[
        \Pi_Q(A_{2s}-\Sigma)q_rq_r^\top(A_{2s}-\Sigma)^\top\Pi_Q
    \right]  \\
    &\quad+
    \mathbb E\!\left[
        \Pi_Q(A_{2s-1}-\Sigma)q_rq_r^\top(A_{2s-1}-\Sigma)^\top\Pi_Q
    \right] \\
    &=
    2C_r(Q).
\end{aligned}
\]
Thus the empirical matrix
\(
    \widehat C_{j,r}
    :=
    \frac1{2b}
    \sum_{s\in\mathcal G_j}
    \widetilde Y_s^{(r)}
    \bigl(\widetilde Y_s^{(r)}\bigr)^\top
\) is an unbiased estimator of \(C_r(Q)\).  We define
\(
    \lambda_{j,r}:=\lambda_{\max}(\widehat C_{j,r}),
    \qquad
    m_j:=\max_{r\in[k]}\lambda_{j,r}.
\)

By the concentration result underlying Theorem~6.1
of~\citet{liu2022dp}, applied conditionally on \(Q\) to the
independent centered vectors \(  \widetilde Y_s^{(r)}/\sqrt 2, s\in\mathcal G_j, \) and then union bounded over \(r\in[k]\), the
batch-size condition
\[
    b
    \ge
    \widetilde \Omega\!\left(
        K^2\left(d+\log\frac{km_0}{\alpha}\right)
    \right)
\]
implies that, for each fixed batch \(j\), with probability at least
\(1-\alpha/(10m_0)\),
\[
    \max_{r\in[k]}
    \left\|
        \widehat C_{j,r}-C_r(Q)
    \right\|_2
    \le
    \eta_0 g_Q^2,
    \eta_0:=1-2^{-1/2}.
\] A union bound over the \(m_0\) batches gives that, with probability at least
\(1-\alpha/10\), this event holds simultaneously for all \(j\in[m_0]\)
and all \(r\in[k]\). On this event, we prove that every batch
statistic \(m_j\) lies in a constant factor interval around \(g_Q^2\).
Let \(r_\star\) be a column attaining the maximum in the definition of
\(g_Q^2\), so that \( \|C_{r_\star}(Q)\|_2=g_Q^2.\) By Weyl's
inequality,
\[
    \lambda_{j,r_\star}
    =
    \lambda_{\max}(\widehat C_{j,r_\star}) 
    \ge
    \lambda_{\max}(C_{r_\star}(Q))
    -
    \|\widehat C_{j,r_\star}-C_{r_\star}(Q)\|_2  
    \ge
    (1-\eta_0)g_Q^2
    =
    2^{-1/2}g_Q^2 .
\]
Therefore
\[
    m_j=\max_{r\in[k]}\lambda_{j,r}
    \ge
    2^{-1/2}g_Q^2 .
\]
For every \(r\in[k]\),
\[
    \lambda_{j,r}
    \le
    \lambda_{\max}(C_r(Q))
    +
    \norm{\widehat C_{j,r}-C_r(Q)}_2
    \le
    2^{1/2}g_Q^2 .
\]
Taking the maximum over \(r\) yields simultaneously for all \(j\in[m_0]\),
\[
    m_j
    \in\bs{
        2^{-1/2}g_Q^2,\,
        2^{1/2}g_Q^2
    }.
\]

It remains to pass from the non-private block statistics to the
private range estimate.  The geometric bins in the stable histogram
algorithm have multiplicative width \(2^{1/4}\).  The interval
\[
    I_Q
    :=
    \left[
        2^{-1/2}g_Q^2,\,
        2^{1/2}g_Q^2
    \right]
\]
has endpoint ratio \(2\), and therefore intersects at most a universal
constant number \(C_\Omega\) of consecutive geometric bins. Therefore,
by \Cref{lem:stability-hist}, the bin selected by the private
histogram intersects \(I_Q\) with probability at least \(1-\alpha/2\).

Let the selected bin have left endpoint \(\ell\).  Then, 
\[
    2^{-3/4}g_Q^2
    \le
    \ell
    \le
    2^{1/2}g_Q^2 .
\]
	The algorithm outputs \(\widehat\Lambda=2\ell\), where the factor \(2\) is a
	universal safety constant.  Hence
\[
    g_Q^2
    \le
    \widehat\Lambda
    \le
    2^{3/2}g_Q^2 .
\]
Absorbing \(2^{3/2}\) into a universal constant \(C_\Lambda\), we obtain
\[
    g_Q^2
    \le
    \widehat\Lambda
    \le
    C_\Lambda g_Q^2 .
\]
Combining the concentration event and the histogram event gives the claim with
probability at least \(1-\alpha\).

\end{proof}

\begin{lemma}[No clipping for projected tangent truncated mean]
\label{lem:tangent-no-clipping-projected}
Fix an iteration and condition on the past, so that
\(Q\in\bS_{d,k}\), \(\Pi=I-QQ^\top\), and \(\widehat\Lambda\) are fixed.
Let
\[
    Y_i=\Pi A_iQ,
    \qquad
    \mu_Q=\bE\bs{Y_i}=\Pi\Sigma Q,
    \qquad
    i=1,\dots,m.
\]
Suppose \(g_Q^2\le \widehat\Lambda\le C_\Lambda g_Q^2\). Run
\(\TruncatedMean\) on \(Y_1,\ldots,Y_m\), with coordinate radius
\[
    r
    =
    C_RK\sqrt{\widehat\Lambda}
    \log^a\!\left(\frac{mdkT}{\alpha}\right)
\]
and center privacy parameters \((\epsilon_C,\delta_C)\). If
\[
    m\ge
    \softOm{
        \frac{dk}{\epsilon_C}
    },
\]
then with probability at least \(1-\alpha\),
\[
    (Y_i)_{jr}\in[\bar g_{jr}-r,\bar g_{jr}+r]
    \qquad
    \text{for all }i\in[m],\ j\in[d],\ r\in[k],
\]
where \(\bar g_{jr}\) is the coordinate center selected by \(\TruncatedMean\).
Consequently, if \(\widetilde Y_i\) denotes the truncated version of \(Y_i\),
\[
    \Pi\left(\frac1m\sum_{i=1}^m\widetilde Y_i\right)
    =
    \frac1m\sum_{i=1}^mY_i .
\]
\end{lemma}

\begin{proof}[Proof of \Cref{lem:tangent-no-clipping-projected}]
Condition on the past and on the successful event
\(g_Q^2\le \widehat\Lambda\le C_\Lambda g_Q^2\).  Then \(Q\), \(\Pi\),
and \(\widehat\Lambda\) are fixed, and the current mean batch is independent of
them.

Fix \((j,r)\in[d]\times[k]\).  Since
\[
    (Y_i-\mu_Q)_{jr}
    =
    e_j^\top\Pi(A_i-\Sigma)q_r
    =
    (\Pi e_j)^\top(A_i-\Sigma)q_r,
\]
if \(\Pi e_j=0\), this coordinate is identically zero.  Otherwise set
\(u_j=\Pi e_j/\|\Pi e_j\|_2\).  Then \(u_j\perp\operatorname{span}(Q)\),
\(\|u_j\|_2=1\), and
\[
    (Y_i-\mu_Q)_{jr}
    =
    \|\Pi e_j\|_2\,u_j^\top(A_i-\Sigma)q_r .
\]
Thus \Cref{ass:T5}, together with \(\widehat\Lambda\ge g_Q^2\), implies that
for
\[
    L
    :=
    C_{\rm tail}K\sqrt{\widehat\Lambda}
    \log^a\!\left(\frac{mdkT}{\alpha}\right),
\]
we have
\[
    \Pr\!\left[
        |(Y_i-\mu_Q)_{jr}|>L
    \right]
    \le
    \frac{\alpha}{16mdkT}.
\]
A union bound over all \(dk\) coordinates and all samples in the mean batch
gives, with probability at least \(1-\alpha/4\),
\[
    |(Y_i-\mu_Q)_{jr}|\le L
\]
simultaneously for all \(i\in[m]\) and all \((j,r)\).

On this event, for each coordinate \((j,r)\), all samples lie in
\[
    I_{jr}:=[(\mu_Q)_{jr}-L,(\mu_Q)_{jr}+L].
\]
Since the histogram bins have width \(\sqrt{\widehat\Lambda}\), the interval
\(I_{jr}\) intersects at most
\[
    C_{\rm bin}
    \le
    C K\log^a\!\left(\frac{mdkT}{\alpha}\right)
\]
consecutive bins.  By the stability-based histogram guarantee, equivalently by
\Cref{lem:stability-hist}, applied with per-coordinate privacy budget
\((\epsilon_C/(dk),\delta_C/(dk))\) and union bounded over the \(dk\) coordinates,
the selected private bin intersects \(I_{jr}\) for every coordinate with
probability at least \(1-\alpha/8\), provided
\[
    m\ge
    \softOm{
        \frac{dk}{\epsilon_C}
    },
\]
with logarithmic dependence on \(dk,\alpha,\delta_C\) and the
\(K\log^a(mdkT/\alpha)\) factor hidden in \(\softOm{\cdot}\).

Let \(\bar g_{jr}\) be the left endpoint of the selected bin.  Since the
selected bin intersects \(I_{jr}\),
\[
    |\bar g_{jr}-(\mu_Q)_{jr}|
    \le
    L+\sqrt{\widehat\Lambda}
    \le
    C K\sqrt{\widehat\Lambda}
    \log^a\!\left(\frac{mdkT}{\alpha}\right).
\]
Thus, on the intersection of the tail and histogram events, for all \(i,j,r\),
\[
    |(Y_i)_{jr}-\bar g_{jr}|
    \le
    |(Y_i-\mu_Q)_{jr}|
    +
    |(\mu_Q)_{jr}-\bar g_{jr}|
    \le
    r
\]
after choosing \(C_R\) sufficiently large.  Hence no coordinate is clipped, so
\(\widetilde Y_i=Y_i\) for all \(i\).  Since \(Y_i=\Pi A_iQ\), we have
\(\Pi Y_i=Y_i\), and therefore
\[
    \Pi\left(\frac1m\sum_{i=1}^m\widetilde Y_i\right)
    =
    \frac1m\sum_{i=1}^mY_i .
\]
Summing the failure probabilities gives the stated probability after adjusting
constants.
\end{proof}

\begin{lemma}[Tangent perturbation bound]
\label{lem:tangent-perturbation-bound}
Fix an iteration and condition on the past and the randomness that
determines \(r_t\), so \(Q=Q_{t-1}\), \(\Pi=I-QQ^\top\), and
\(r_t\) are fixed.  Let \(m=|\mathcal B_t^{\rm mean}|\), and define
\[
    S_t
    :=
    \frac1m\sum_{i\in\mathcal B_t^{\rm mean}}
    \Pi(A_i-\Sigma)Q,
    \qquad
    E_t:=S_t+W_t,
\]
where \(W_t=\Pi Z_t, (Z_t)_{jr}\overset{\mathrm{i.i.d.}}{\sim}\mathcal
N(0,\sigma_t^2), \) with
\(
    \sigma_t
    =
    \frac{2r_t\sqrt{dk}}{m\epsilon_M}
    \sqrt{2\log(1.25/\delta_M)} .
\)
Let
\[
    Y_i:=\Pi(A_i-\Sigma)Q,
\]
and define
\[
    v_Q^2
    :=
    \max\left\{
        \left\|\mathbb E[Y_iY_i^\top]\right\|_2,
        \left\|\mathbb E[Y_i^\top Y_i]\right\|_2
    \right\}.
\]
Assume
\[
    \|Y_i\|_2\le M_\perp
    \qquad\text{a.s.}
\]
Then, on the event \( r_t
    \le
    C K g_Q\, \operatorname{polylog}(BdkT/\zeta), \)
with probability at least \(1-\xi\) over the mean batch and Gaussian noise,
we have,
    suppressing logarithmic factors,
\[
    \|E_t\|_F
    \le
    \softO{
        \sqrt{k}v_Qm^{-1/2}
        +
        \sqrt{k}M_\perp m^{-1}
        +
        \frac{K g_Qdk}{m\epsilon}
    }.
\]
\end{lemma}

\begin{proof}
    Conditionally on the past and on \(r_t\), the matrices
\(
    Y_i=\Pi(A_i-\Sigma)Q,~i\in\mathcal B_t^{\rm mean},
\) are independent and mean zero.  We first bound the stochastic term
\(S_t\). For a rectangular matrix \(Y\in\mathbb R^{d\times k}\),
define its self-adjoint dilation
\[
    \mathcal D(Y)
    :=
    \begin{pmatrix}
        0 & Y\\
        Y^\top & 0
    \end{pmatrix}
    \in\mathbb R^{(d+k)\times(d+k)}.
\]
Then \(  \|\mathcal D(Y_i)\|_2=\|Y_i\|_2\le M_\perp, \) and by the
definition of \(v_Q\),
\(
    \left\|
        \sum_{i\in\mathcal B_t^{\rm mean}}
        \mathbb E[\mathcal D(Y_i)^2]
    \right\|_2
    \le
    m v_Q^2 .
\)
Self-adjoint matrix Bernstein applied in dimension \(d+k\) gives, with
probability at least \(1-\xi/2\),
\[
    \left\|
        \frac1m\sum_{i\in\mathcal B_t^{\rm mean}}Y_i
    \right\|_2
    \le
    C\left(
        v_Q\sqrt{\frac{\log((d+k)/\xi)}{m}}
        +
        \frac{M_\perp\log((d+k)/\xi)}{m}
    \right).
\]

Since \(S_t\in\mathbb R^{d\times k}\), we have \( \|S_t\|_F\le
    \sqrt{k}\|S_t\|_2.  \) Therefore, with probability at least
    \(1-\xi/2\),
\[
    \|S_t\|_F
    \le
    C\sqrt{k}\left(
        v_Q\sqrt{\frac{\log((d+k)/\xi)}{m}}
        +
        \frac{M_\perp\log((d+k)/\xi)}{m}
    \right).
\]

It remains to bound the Gaussian term.  Let \(Q_\perp\) be an
orthonormal basis for \(\operatorname{range}(\Pi)\), so that
\(\Pi=Q_\perp Q_\perp^\top\).  Then \( W_t=\Pi
Z_t=Q_\perp(Q_\perp^\top Z_t),\) and \( \|W_t\|_F=\|Q_\perp^\top
Z_t\|_F.  \) By rotational invariance of the Gaussian distribution,
\(Q_\perp^\top Z_t\in\mathbb R^{(d-k)\times k}\) has i.i.d.
\(\mathcal N(0,\sigma_t^2)\) entries.  Thus
\(
    \frac{\|W_t\|_F^2}{\sigma_t^2}
    \sim
    \chi^2_{(d-k)k}.
\) A standard chi-square tail bound implies that, with probability at least
\(1-\xi/2\),
\[
    \|W_t\|_F
    \le
    C\sigma_t\sqrt{dk+\log(1/\xi)}.
\]
Substituting
\(
    \sigma_t
    =
    \frac{2r_t\sqrt{dk}}{m\epsilon_M}
    \sqrt{2\log(1.25/\delta_M)}
\) gives
\[
    \|W_t\|_F
    \le
    C\frac{r_t\sqrt{dk}}{m\epsilon_M}
    \sqrt{(dk+\log(1/\xi))\log(1/\delta_M)}.
\]

Combining the bounds for \(S_t\) and \(W_t\) by the triangle
inequality and using the event,
\[
    r_t
    \le
    C K g_Q\,
    \polylog{BdkT/\zeta},
\] completes the proof.
\end{proof}

\begin{lemma}[Local contraction of the noisy tangent step]
\label{lem:local-tangent-contraction}
There exist universal constants \(c_0,c,c',C>0\) such that the following holds.
Let \(P_\star=V_kV_k^\top\), let
\(\Delta=\lambda_k-\lambda_{k+1}>0\), and let
\(Q\in\bS_{d,k}\) satisfy
\[
    \|QQ^\top-P_\star\|_F\le c_0 .
\]
Let \(E\in\mathbb R^{d\times k}\) satisfy \(Q^\top E=0\).  If
 \(\eta\le c/(\lambda_1-\lambda_d)\). and
\[
    Q^+
    =
    \mathrm{QR}\!\left(
        Q+\eta(\Pi_Q\Sigma Q+E)
    \right),
    \qquad
    \Pi_Q=I-QQ^\top,
\]
then
\[
    \|Q^+(Q^+)^\top-P_\star\|_F
    \le
    (1-c'\eta\Delta)
    \|QQ^\top-P_\star\|_F
    +
    C\eta\|E\|_F .
\]
\end{lemma}

\begin{proof}
Let \(  G_Q:=\Pi_Q\Sigma Q.\) We first prove contraction for the
noiseless update
\[
    \bar Q=\mathrm{QR}(Q+\eta G_Q),
\]
and then compare \(\bar Q\) with the noisy update. Work in the
eigenbasis \([V_k,V_\perp]\) of \(\Sigma\), so that
\[
    \Sigma=\mathrm{diag}(\Lambda_1,\Lambda_2),
    \qquad
    \Lambda_1=\mathrm{diag}(\lambda_1,\dots,\lambda_k),
    \qquad
    \|\Lambda_2\|_2\le \lambda_{k+1}.
\]
Set
\[
    A:=V_k^\top Q,
    \qquad
    B:=V_\perp^\top Q,
    \qquad
    H:=Q^\top\Sigma Q,
    \qquad
    N:=\Lambda_2B-BH .
\]
Since \(Q^\top Q=I_k\), we have \(A^\top A+B^\top B=I_k\).  Moreover,
\[
    \|QQ^\top-P_\star\|_F=\sqrt{2}\|B\|_F,
\]

Let \(\bar Y:=Q+\eta G_Q .\) The bottom block of \(\bar Y\) is \(
    V_\perp^\top\bar Y=B+\eta N .  \) Also \(Q^\top G_Q=0\), and
    therefore
\[
    \bar Y^\top\bar Y
    =
    I_k+\eta^2G_Q^\top G_Q
    \succeq I_k .
\]
Thus QR normalization right-multiplies \(\bar Y\) by a matrix of operator norm
at most one, so
\[
    \|V_\perp^\top\bar Q\|_F^2
    \le
    \|B+\eta N\|_F^2  =
    \|B\|_F^2
    +
    2\eta\langle B,N\rangle
    +
    \eta^2\|N\|_F^2 .
\]

We now bound the two terms involving \(N\).  
\[
    -\langle B,N\rangle
    =
    \operatorname{tr}(B^\top B H)
    -
    \operatorname{tr}(B^\top\Lambda_2B).
\]
Equivalently, using \(A^\top A+B^\top B=I_k\) and
\(\lambda_i-\lambda_j\ge \Delta\) for \(i\le k<j\),
\[
    -\langle B,N\rangle
    =
    \sum_{i\le k<j}
    (\lambda_i-\lambda_j)
    |(AB^\top)_{ij}|^2  \ge
    \Delta\|AB^\top\|_F^2 .
\]
Finally,
\[
    \|AB^\top\|_F^2
    =
    \operatorname{tr}(B^\top B A^\top A)
    =
    \operatorname{tr}\!\left(B^\top B(I_k-B^\top B)\right)
    \ge
    (1-\|B\|_F^2)\|B\|_F^2 .
\]
Hence
\(
    -\langle B,N\rangle
    \ge
    \Delta(1-\|B\|_F^2)\|B\|_F^2 .
\) The same block calculation also gives, for \(c_0\) sufficiently small,
\[
    \|N\|_F^2
    \le
    C
    \sum_{i\le k<j}
    (\lambda_i-\lambda_j)^2
    |(AB^\top)_{ij}|^2
    \le
    C\lambda_1[-\langle B,N\rangle].
\]
Therefore, if \(\eta\le c/(\lambda_1-\lambda_d)\),
\[
\begin{aligned}
    \|V_\perp^\top\bar Q\|_F^2
    &\le
    \|B\|_F^2
    -
    2\eta[-\langle B,N\rangle]
    +
    C\eta^2\lambda_1[-\langle B,N\rangle]  \\
    &\le
    \|B\|_F^2
    -
    c\eta\Delta\|B\|_F^2  \\
    &\le
    (1-c\eta\Delta)\|B\|_F^2 ,
\end{aligned}
\]
where in the second line we used \(\|B\|_F\le c_0\) and chose \(c_0\) and
\(c\) small enough.  Taking square roots and adjusting constants,
\[
    \|V_\perp^\top\bar Q\|_F
    \le
    (1-c'\eta\Delta)\|B\|_F .
\]
Since projector Frobenius distance is \(\sqrt 2\|V_\perp^\top Q\|_F\),
\[
    \|\bar Q\bar Q^\top-P_\star\|_F
    \le
    (1-c'\eta\Delta)\|QQ^\top-P_\star\|_F .
\]

Now consider the noisy update
\[
    Y:=Q+\eta(G_Q+E).
\]
Since \(Q^\top G_Q=0\) and \(Q^\top E=0\),
\[
    Q^\top Y=Q^\top\bar Y=I_k .
\]
Thus both \(Y\) and \(\bar Y\) have smallest singular value at least one.  The
orthogonal projectors onto their column spaces therefore satisfy
\[
    \|P_Y-P_{\bar Y}\|_F
    \le
    C\|Y-\bar Y\|_F
    =
    C\eta\|E\|_F,
\]
where
\[
    P_Y=Q^+(Q^+)^\top,
    \qquad
    P_{\bar Y}=\bar Q\bar Q^\top .
\]
The triangle inequality gives
\[
\begin{aligned}
    \|Q^+(Q^+)^\top-P_\star\|_F
    &\le
    \|\bar Q\bar Q^\top-P_\star\|_F
    +
    \|P_Y-P_{\bar Y}\|_F  \\
    &\le
    (1-c'\eta\Delta)\|QQ^\top-P_\star\|_F
    +
    C\eta\|E\|_F .
\end{aligned}
\]
This proves the lemma.
\end{proof}

\begin{lemma}[Gaussian covariance]
\label{cor:gaussian-covariance}
Let \(A_i=x_ix_i^\top\) with \(x_i\sim\cN(0,\Sigma)\), and let
\[
    \operatorname{tr}_{>k}(\Sigma):=\sum_{j>k}\lambda_j.
\]
In a local basin around \(P_\star\), \Cref{ass:tangent-model} holds with
\[
    v_0\asymp \sqrt{\lambda_1(\operatorname{tr}_{>k}(\Sigma)+k\lambda_{k+1})},
    \qquad
    v_1\asymp \lambda_1\sqrt{k},
\]
and
\[
    g_0\asymp \sqrt{\lambda_1\lambda_{k+1}},
    \qquad
    g_1\asymp \lambda_1,
\]
up to logarithmic high-probability truncation factors in \(M_\perp\).  Consequently, under the absorption condition in \Cref{thm:tangent-general-utility},
\[
\norm{Q_TQ_T^\top-P_\star}_F
\le
\operatorname{Opt}_T
+
\softO{
    \frac{\sqrt{k\lambda_1(\operatorname{tr}_{>k}(\Sigma)+k\lambda_{k+1})}}{\Delta\sqrt n}
    +
    \frac{\sqrt{k}M_\perp}{\Delta n}
    +
    \frac{K\sqrt{\lambda_1\lambda_{k+1}}\,dk}{\Delta\epsilon n}
}.
\]
\end{lemma}

\begin{proof}
The Gaussian fourth-moment identity gives, for every fixed symmetric matrix \(B\),
\begin{equation}
\label{eq:gaussian-fourth}
    \bE\bs{(A-\Sigma)B(A-\Sigma)}
    =
    \tr(\Sigma B)\Sigma+\Sigma B\Sigma.
\end{equation}
Fix \(Q\in\bS_{d,k}\), set \(P=QQ^\top\), \(\Pi=I-P\), and define \(e_2=\norm{P-P_\star}_2\).  For a column \(q\) of \(Q\), \eqref{eq:gaussian-fourth} with \(B=qq^\top\) gives
\[
\Pi\bE\bs{(A-\Sigma)qq^\top(A-\Sigma)}\Pi
=
(q^\top\Sigma q)\Pi\Sigma\Pi
+
(\Pi\Sigma q)(\Pi\Sigma q)^\top.
\]
If \(z\in\operatorname{range}(\Pi)\) is unit norm, then \(Pz=0\), so \(P_\star z=(P_\star-P)z\) and \(\norm{P_\star z}\le e_2\).  Hence
\[
    z^\top\Sigma z
    \le
    \lambda_{k+1}+\lambda_1e_2^2,
    \qquad
    \norm{\Pi\Sigma\Pi}_2\le \lambda_{k+1}+\lambda_1e_2^2.
\]
Moreover,
\[
    \norm{\Pi\Sigma q}_2
    \le
    \norm{\Pi\Sigma(P-P_\star)}_2+
    \norm{\Pi P_\star\Sigma P_\star}_2
    \le
    2\lambda_1e_2.
\]
Since \(q^\top\Sigma q\le\lambda_1\),
\[
    g_Q^2
    \le
    C(\lambda_1\lambda_{k+1}+\lambda_1^2e_2^2),
\]
and therefore
\[
    g_Q\le C(\sqrt{\lambda_1\lambda_{k+1}}+\lambda_1e(Q)).
\]
This gives the stated choices of \(g_0,g_1\).

For \(v_Q\), write \(Y=\Pi(A-\Sigma)Q\).  Since \(P=QQ^\top\),
\[
    YY^\top=\Pi(A-\Sigma)P(A-\Sigma)\Pi.
\]
Applying \eqref{eq:gaussian-fourth} with \(B=P\),
\[
    \bE\bs{YY^\top}
    =
    \Pi\{\tr(\Sigma P)\Sigma+\Sigma P\Sigma\}\Pi.
\]
Using \(\tr(\Sigma P)\le k\lambda_1\), \(\norm{\Pi\Sigma\Pi}_2\le\lambda_{k+1}+\lambda_1e_2^2\), and \(\norm{\Pi\Sigma P}_2\le2\lambda_1e_2\), we obtain
\[
    \norm{\bE\bs{YY^\top}}_2
    \le
    Ck\lambda_1(\lambda_{k+1}+\lambda_1e_2^2).
\]
Similarly,
\[
    Y^\top Y=Q^\top(A-\Sigma)\Pi(A-\Sigma)Q.
\]
Using \eqref{eq:gaussian-fourth} with \(B=\Pi\),
\[
    \bE\bs{Y^\top Y}
    =
    Q^\top\{\tr(\Sigma\Pi)\Sigma+\Sigma\Pi\Sigma\}Q.
\]
Now
\[
    \tr(\Sigma\Pi)
    \le
    \operatorname{tr}_{>k}(\Sigma)+k\lambda_1e_2^2,
\]
and \(\norm{\Pi\Sigma Q}_2^2\le4\lambda_1^2e_2^2\).  Thus
\[
    \norm{\bE\bs{Y^\top Y}}_2
    \le
    C\left(\lambda_1\operatorname{tr}_{>k}(\Sigma)+k\lambda_1^2e_2^2\right).
\]
Combining the two variance bounds yields
\[
    v_Q^2
    \le
    C\left(\lambda_1[\operatorname{tr}_{>k}(\Sigma)+k\lambda_{k+1}]+k\lambda_1^2e_2^2\right),
\]
so
\[
    v_Q
    \le
    C\left(\sqrt{\lambda_1[\operatorname{tr}_{>k}(\Sigma)+k\lambda_{k+1}]}+\lambda_1\sqrt{k}\,e(Q)\right).
\]
This gives the stated \(v_0,v_1\).
\end{proof}

\begin{proof}[Proof of \Cref{cor:flat-tail}]
For flat-tail covariance, \(\operatorname{tr}_{>k}(\Sigma)=(d-k)\lambda_{k+1}\).
Hence \Cref{cor:gaussian-covariance} gives
\[
    v_0\asymp \sqrt{d\lambda_1\lambda_{k+1}},
    \qquad
    g_0\asymp \sqrt{\lambda_1\lambda_{k+1}} .
\]
Substituting these values into \Cref{thm:tangent-general-utility} yields
\[
\norm{Q_TQ_T^\top-P_\star}_F
\le
\operatorname{Opt}_T
+
\widetilde O\left(
\frac{\sqrt{\lambda_1\lambda_{k+1}}}{\Delta}
\left[
\sqrt{\frac{dk}{n}}
+
\frac{Kdk}{\varepsilon n}
\right]
+
\frac{\sqrt{k}M_\perp}{\Delta n}
\right).
\]
Since \(\Delta=\lambda_k-\lambda_{k+1}\), this gives the inequality.
\end{proof}

%% file: appendices/existing_lower_bounds.tex
\section{Results from Prior work}
\subsection{Existing Lower Bounds}\label{sec:lower-bounds}

\begin{theorem}[Lower bound, Gaussian distribution, Theorem 5.3 in \cite{liu2022dp}]\label{thm-gauss-lower-bound-k-1} Let $\mathcal{M}_{\epsilon}$ be a class of $(\epsilon, 0)$-DP estimators that map $n$ i.i.d. samples to an estimate $\hat{v} \in \mathbb{R}^d$. A set of Gaussian distributions with $(\lambda_1, \lambda_2)$ as the first and second eigenvalues of the covariance matrix is denoted by $\mathcal{P}_{(\lambda_1, \lambda_2)}$. There exists a universal constant $C >0$ such that 
\begin{align*}
    \inf_{\hat{v} \in \mathcal{M}_{\epsilon}} \sup_{P \in \mathcal{P}_{(\lambda_1, \lambda_2)}} \mathbb{E}_{S \sim P^n}[\sin(\hat{v}(S), v_1)] \geq C \min \left( \kappa \left(\sqrt{\frac{d}{n}} + \frac{d}{\epsilon n} \right) \sqrt{\frac{\lambda_2}{\lambda_1}}, 1 \right)
\end{align*}
where $v_1$ is the true top eigenvector of the expectation of the i.i.d. samples and $\kappa = \frac{\lambda_1}{\lambda_1 - \lambda_2}$.
\end{theorem}

\begin{theorem}[Theorem 4.2 in \citet{cai2024optimal}]\label{spiked-cov-lower-bound} Let the $d \times n$ data matrix $X$ have i.i.d. columns sampled from a distribution $P = \mathcal{N}(0, U \Lambda U^{\top} + \sigma^2 \mathbf{I}_d) \in \mathcal{P}(\lambda, \sigma^2)$, where $U \in \mathbb{R}^{d \times k}$ has orthonormal columns. Suppose $\delta \leq c_0' \exp \{2 \epsilon - c_0(\epsilon \sqrt{ndk} + dk) \}$ for some small constants $c_0, c'_0 > 0$. Then, there exists an absolute constant $c_1 > 0$ such that 
\begin{align*}
    \inf_{\tilde{U} \in \mathcal{U}_{\epsilon, \delta}} \sup_{P \in \mathcal{P}(\lambda, \sigma^2)} \frac{\mathbb{E} \| \tilde{U} \tilde{U}^{\top} - UU^{\top} \|_F}{\sqrt{k}} \geq c_1 \left( \left( \frac{\sigma \sqrt{\lambda + \sigma^2}}{\lambda}\right) \left( \sqrt{\frac{d}{n}} + \frac{d \sqrt{k}}{n \epsilon}\right) \bigwedge 1 \right)
\end{align*}
where the infimum is taken over all the possible $(\epsilon, \delta)$-DP algorithms, denoted by $\mathcal{U}_{\epsilon, \delta}$ and the expectation is taken with respect to both $\tilde{U}$ and $P$ and \[\mathcal{P}(\lambda, \sigma^2) := \{ \mathcal{N}(0, \Sigma): \Sigma = U \Lambda U^{\top} + \sigma^2 \mathbf{I}_d, U \in \mathbb{O}_{d,k} , \Lambda = \text{diag}(\lambda_1, \dots, \lambda_k), c_0 \lambda \leq \lambda_k \leq \lambda_1 \leq C_0 \lambda \}\]
\end{theorem}

\begin{corollary}[Corollary 3 in \citet{dungler2025iterative}]
\label{corollary-lower-bound-spiked-cov}
    Let the $d \times n$ data matrix $X$ have i.i.d. columns sampled from a distribution $P = \mathcal{N}(0, U \Lambda U^{\top} + \sigma^2 \mathbf{I}_d) \in \mathcal{P}(\lambda, \sigma^2)$ where $\mathcal{P}(\lambda, \sigma^2) = \{\mathcal{N}(0, \Sigma), \Sigma = U\Lambda U^{\top} + \sigma^2 \mathbf{I}_d, c \lambda \leq \lambda_k \leq \dots \leq \lambda_1 \leq C \lambda\}$. Suppose $\delta \leq c_0' \exp \{2 \epsilon - c_0(\epsilon \sqrt{ndk} + dk) \}$ for some small constants $c_0, c'_0 > 0$. Then, there exists an absolute constant $c_1 > 0$ such that 
    \begin{align*}
         \inf_{\tilde{U} \in \mathcal{U}_{\epsilon, \delta}} \sup_{P \in \mathcal{P}(\lambda, \sigma^2)} \mathbb{E}[\zeta]  \geq c_1 \left( \left(\frac{\sigma \sqrt{\lambda_1 + \sigma^2}}{\sum_{i=1}^k (\lambda_i + \sigma^2)} \right) \left( \sqrt{\frac{dk}{n}} + \frac{dk}{n \epsilon}\right) \bigwedge 1 \right).
    \end{align*}
\end{corollary}

\subsection{Existing Algorithms}

\begin{algorithm}[ht]\caption{Oja's Algorithm,~\citet{huang2021streaming}}\label{algo:ojas}
\textbf{Input:} symmetric matrices $S = \{A_1, \dots, A_{n} \}$ in $\mathbb{R}^{d \times d}$, $k \in [d]$ , learning rates $\{ \eta_t \}_{t=1}^{n}$
    \begin{algorithmic}[1]
    
        \State Choose $Q_0' \in \mathbb{R}^{d \times k}$ uniformly at random,~$Q_0 \gets \text{QR}[Q_0']$
       
        \For{$t = 1,2, \dots, T = n$}
            \State  $Q_t' \gets Q_{t-1} + \eta_t A_t Q_{t-1}$\label{step:ojas-update-step}
            \State $Q_t \gets \text{QR}[Q_t']$
        \EndFor
        \State \Return $Q_T$
    \end{algorithmic}
    
\end{algorithm}

\begin{theorem}[Main Theorem, ~\citet{huang2021streaming}]\label{thm:main-thm-streaming}
Given matrices $A_1, \dots, A_n \in \mathbb{R}^{d \times d}$ that are i.i.d., symmetric random matrices, satisfying $\mathbb{E}[A_i] = \Sigma$ for all $i$  with \( \rho_k = \lambda_k - \lambda_{k+1} > 0 \) the \(k\)th eigengap of \(\Sigma\) and the uniform bound
\[
    \sup_{i} \sup_{P \in \cS_{d,k}}\| P^\top (A_i - \Sigma) \|_F \leq M \text{ a.s.}
\] For every $\delta \in (0,1)$, define learning rates
\[
n_0 = \tilde{\Theta}\!\left(\frac{kM^2}{\delta^2 \rho_k^2}\right),
\qquad
\beta = \tilde{\Theta}\!\left(\frac{M^2}{\rho_k^2}\right),
\]
and
\[
\eta_t =
\begin{cases}
\tilde{\Theta}\!\left(\dfrac{1}{\rho_k n_0}\right), & t \le n_0, \\[1ex]
\Theta\!\left(\dfrac{1}{\rho_k(\beta+t-n_0)}\right), & t > n_0.
\end{cases}
\]

Let $V_k \in \mathbb R^{d\times k}$ be the orthogonal matrix whose columns are
the $k$ leading eigenvectors of $\Sigma$. Then for any $n > n_0$, the
output $Q_n$ of Oja’s algorithm satisfies
\[
  \big\|Q_nQ_n^\top-V_kV_k^\top\big\|_2
  \;\le\;C'\,\frac M{\rho_k}\sqrt{\frac{\log(Mk/(\rho_k\delta))}{n-n_0}}
\]
with probability at least $1-\delta$, where $C'$ is a universal positive
constant.
\end{theorem}

\begin{algorithm}[ht]
\caption{Private Top Eigenvalue Estimation, Algorithm 4 in~\citep{liu2022dp}}\label{algo:priv-top-eigenval-estimation-liu}
\textbf{Input:} $S = \{g_i\}_{i=1}^{B}$, $(\varepsilon,\delta)$-DP, failure probability $\zeta$
\begin{algorithmic}[1]
\State Let $\tilde{g}_i \gets g_{2i} - g_{2i-1}$ for $i \in 1, 2, \dots, \lfloor B/2 \rfloor$. Let $\tilde{S} = \{\tilde{g}_i\}_{i=1}^{\lfloor B/2 \rfloor}$
\State Partition $\tilde{S}$ into $k = C_1 \log(1/(\delta \zeta))/\varepsilon$ subsets and denote each dataset as $G_j \in \mathbb{R}^{d \times b}$, where each dataset is of size $b = \lfloor B/2k \rfloor$
\State Let $\lambda_1^{(j)}$ be the top eigenvalue of $(1/b)G_j G_j^\top$ for $\forall j \in [k]$
\State Partition $[0,\infty)$ into $\Omega \gets \{\dots, [2^{-2/4},2^{-1/4}), [2^{-1/4},1), [1,2^{1/4}), [2^{1/4},2^{2/4}), \dots\} \cup \{[0,0]\}$
\State Run $(\varepsilon,\delta)$-DP histogram learner of~\Cref{lem:stability-hist} on $\{\lambda_1^{(j)}\}_{j=1}^{k}$ over $\Omega$
\If{all the bins are empty}
    \State \Return $\perp$
\EndIf
\State Let $[l,r]$ be a non-empty bin that contains the maximum number of points in the DP histogram
\State \Return $\hat{\Lambda} = l$
\end{algorithmic}
\end{algorithm}

%% file: appendices/novel_ojas_bound_using_frobenius.tex
\section{A Frobenius-norm analysis of Oja's algorithm}\label{app:frob}

\Cref{thm:main-thm-streaming} is stated in the operator norm, while
\Cref{thm:utility-priv-ojas-batches,thm:utility-priv-ojas-adaptive,thm:tangent-general-utility}
and the lower bound of \Cref{spiked-cov-lower-bound} are stated in the Frobenius norm. Using \Cref{thm:main-thm-streaming} as a black box would cost a conversion factor: by
\Cref{lem:tansin},
\[
    \|Q_nQ_n^\top-V_kV_k^\top\|_F
    =\sqrt2\,\|\sin\Theta\|_F
    \le\sqrt{2k}\,\|Q_nQ_n^\top-V_kV_k^\top\|_2 ,
\]
with equality when the \(k\) principal angles coincide. The Frobenius
bound obtained this way is loose by a factor \(\sqrt k\) compared to
\Cref{spiked-cov-lower-bound}. This appendix shows that the factor is an artifact of the
conversion: under the same assumptions as \citet{huang2021streaming},
carrying both phases of their analysis out directly in the Frobenius norm gives the bound without
it. The conclusion is \Cref{thm:main}.

The obstruction is the norm in which the recursion runs. \citet{huang2021streaming} work in
$\|X\|_{p,p}=(\mathbb E\|X\|_{S_p}^p)^{1/p}$, where the single index $p$ is at once the Schatten
exponent inside the expectation and the moment exponent outside it. Exponential tails force
$p\asymp\log(1/\delta)$, and at that exponent $\|W\|_F\le k^{1/2-1/p}\|W\|_{S_p}$ loses
$\sqrt k$; the inequality is an identity at $p=2$, but $p$ cannot be small and large at once.
\Cref{def:Fp} pins the inner norm at Frobenius and lets the moment index run independently, which
is what makes a Frobenius conclusion with exponential tails available, and since $\|\cdot\|_F$ is
a Hilbert norm the martingale step costs only $\sqrt{p-1}$, with no dimensional factor. 
The analysis below follows Section 4 and Appendices A-D of \citet{huang2021streaming} closely.

Phase II requires nothing beyond \Cref{ass:C}. Phase I requires in addition that the updates be i.i.d., which the algorithms arrange by running their first $T_0$ steps non adaptively.~\Cref{app:G} lifts the finite support results below to that setting.

\subsection{Setting and notation}

\begin{assumption}[$(\Sigma,M)$-predictable Oja model]\label{ass:C}
Let $Z_0\in\mathbb R^{d\times k}$ be the initialisation matrix and let $(\mathcal F_t)_{t\ge0}$ be
a filtration with
\[
  \sigma(Z_0)\subseteq\mathcal F_0 .
\]
The matrices $C_1,\dots,C_T\in\mathbb R^{d\times d}$ are symmetric and satisfy:
\begin{enumerate}
\item[(C.1)] $C_t$ is $\mathcal F_t$-measurable and $\mathbb E[C_t\mid\mathcal F_{t-1}]=\Sigma$
 for every $t\le T$. The matrix $\Sigma\succeq0$ has eigenvalues $\lambda_1\ge\dots\ge
 \lambda_d\ge0$ with corresponding eigenvectors $v_1,\dots,v_d$, and eigengap
 $\rho_k=\lambda_k-\lambda_{k+1}>0$.
\item[(C.2)] The centred updates are uniformly bounded in the Ky Fan $2$-$k$ norm:
 $\|C_t-\Sigma\|_{(2,k)}\le M$ almost surely, for every $t\le T$.
\end{enumerate}
\end{assumption}

 The requirement $\sigma(Z_0)\subseteq\mathcal F_0$ is not an independence hypothesis
and must not be replaced by one: for the adaptive algorithms the updates depend on $Z_0$ from the
first step onwards, and (C.1) demands the conditional mean be $\Sigma$ \emph{given} the
initialisation. (C.2) bounds the total fluctuation energy seen through any $k$-dimensional subspace.

Throughout, $C_1,\dots,C_T$ satisfy Assumption \ref{ass:C}, and we write
$V:=[v_1,\dots,v_k]\in\mathbb R^{d\times k}$ and $U:=[v_{k+1},\dots,v_d]\in\mathbb R^{d\times(d-k)}$
for the leading and trailing eigenvector blocks of $\Sigma$, so that $[\,V\ U\,]$ is orthogonal
and $V^\top U=0$. 
Given the initialisation $Z_0$ of Assumption \ref{ass:C}, put $Q_0:=\mathrm{QR}(Z_0)$, and let
$\eta_t>0$ be step sizes. The Oja iterates are
\[
  Q_t:=\mathrm{QR}\big(Q_{t-1}+\eta_tC_tQ_{t-1}\big),\qquad
  Z_t:=(I+\eta_tC_t)\cdots(I+\eta_1C_1)\,Q_0 ,\qquad t\ge1 ,
\]
so that $Q_t$ is an orthonormal basis for the column span of $Z_t$, and whenever $V^\top Z_t$ is
invertible we set
\[
  W_t:=U^\top Z_t(V^\top Z_t)^{-1}=U^\top Q_t(V^\top Q_t)^{-1}.
\]

The analysis is carried out in $W_t$ rather than in the subspace error itself, because the two are
equivalent up to an absolute constant.

\begin{lemma}[Frobenius $\sin\Theta$ bound]\label{lem:tansin}
Let $Q\in\mathbb R^{d\times k}$ have orthonormal columns and let
$\theta_1,\dots,\theta_k\in[0,\pi/2]$ be the principal angles between $\mathrm{range}(Q)$ and
$\mathrm{range}(V)$, so that $\cos\theta_1,\dots,\cos\theta_k$ are the singular values of
$V^\top Q$, and write $\Theta:=\operatorname{diag}(\theta_1,\dots,\theta_k)$. Then
\[
  \big\|QQ^\top-VV^\top\big\|_F^2=2\|\sin\Theta\|_F^2 ,
  \qquad
  \big\|QQ^\top-VV^\top\big\|_2=\|\sin\Theta\|_2 .
\]
If moreover $V^\top Q$ is invertible and $W:=U^\top Q(V^\top Q)^{-1}$, then the singular values of
$W$ are $\tan\theta_1,\dots,\tan\theta_k$, and consequently
\[
  \big\|QQ^\top-VV^\top\big\|_F\le\sqrt2\,\|W\|_F .
\]
\end{lemma}

\begin{proof}
Both $QQ^\top$ and $VV^\top$ are rank-$k$ orthogonal projectors, so $(QQ^\top)^2=QQ^\top$,
$(VV^\top)^2=VV^\top$ and $\operatorname{tr}(QQ^\top)=\operatorname{tr}(VV^\top)=k$. Expanding,
\[
  \big\|QQ^\top-VV^\top\big\|_F^2=\operatorname{tr}\big((QQ^\top-VV^\top)^2\big)
  =2k-2\operatorname{tr}(QQ^\top VV^\top)=2k-2\|V^\top Q\|_F^2 ,
\]
since $\operatorname{tr}(QQ^\top VV^\top)=\operatorname{tr}(Q^\top VV^\top Q)=\|V^\top Q\|_F^2$.
As $\|V^\top Q\|_F^2=\sum_i\cos^2\theta_i$, the right-hand side is $2\sum_i(1-\cos^2\theta_i)
=2\|\sin\Theta\|_F^2$. For the second identity, the nonzero eigenvalues of $QQ^\top-VV^\top$ are
$\pm\sin\theta_1,\dots,\pm\sin\theta_k$, each appearing once, so the operator norm is
$\max_i\sin\theta_i=\|\sin\Theta\|_2$.

For the third claim put $Y_1:=V^\top Q$, $Y_2:=U^\top Q$. Since $[\,V\ U\,]$ is orthogonal,
$Y_1^\top Y_1+Y_2^\top Y_2=Q^\top Q=I_k$. Let $Y_1=ACB^\top$ be a singular value decomposition, so
$C=\cos\Theta$ after ordering; then $Y_2^\top Y_2=I_k-BC^2B^\top=B\sin^2\Theta\,B^\top$. As
$V^\top Q$ is invertible, every $\cos\theta_i>0$, so $C$ is invertible,
$W=Y_2Y_1^{-1}=Y_2BC^{-1}A^\top$ and
\[
  W^\top W=AC^{-1}B^\top Y_2^\top Y_2BC^{-1}A^\top=AC^{-1}\sin^2\Theta\,C^{-1}A^\top
   =A\tan^2\Theta\,A^\top .
\]
Hence the singular values of $W$ are the $\tan\theta_i$, and since $\sin\theta\le\tan\theta$ on
$[0,\pi/2)$, $\|W\|_F^2=\sum_i\tan^2\theta_i\ge\sum_i\sin^2\theta_i=\frac12
\|QQ^\top-VV^\top\|_F^2$.
\end{proof}

Applied at $Q=Q_t$, \Cref{lem:tansin} gives
$\|Q_tQ_t^\top-VV^\top\|_F\le\sqrt2\,\|W_t\|_F$ for every $t$ and for our analysis is therefore suffices to bound $\|W\|_F$. We write $\Lambda_1:=\operatorname{diag}(\lambda_1,\dots,\lambda_k)$ and $\Lambda_2:=
\operatorname{diag}(\lambda_{k+1},\dots,\lambda_d)$, so $\Sigma=V\Lambda_1V^\top+U\Lambda_2U^\top$,
and, for a threshold $\gamma\ge1$ to be fixed in each phase,
\[
  \varepsilon_t:=2\eta_tM(1+\gamma),\qquad
  \chi_t:=\frac{1+\eta_t\lambda_{k+1}}{1+\eta_t\lambda_k}\in(0,1] .
\]
When the step size is constant we drop the subscripts and write $\varepsilon,\chi,\eta$. Each
phase conditions on a sequence of decreasing \emph{good events} $G_0\supseteq G_1\supseteq\cdots$,
adapted to the filtration $(\mathcal F_t)$ of Assumption C, with indicators $\mathbf 1_t:=
\mathbf 1_{G_t}$; the two phases use different events, specified in Sections \ref{sec:phase1} and
\ref{sec:phase2}. We abbreviate the small-step conditions
\begin{align}
  \varepsilon_i&\le\tfrac12, \tag{S1}\label{S1}\\
  \eta_i\|\Sigma\|_2&\le\tfrac12, \tag{S2}\label{S2}\\
  e^{-\eta_i\rho_k/4}&\le\varepsilon_i/\varepsilon_{i-1}
   \quad\text{(vacuous at $i=1$ and for constant step size)}, \tag{S3}\label{S3}\\
  p\varepsilon_i^2&\le\eta_i\rho_k/400 . \tag{S4}\label{S4}
\end{align}
$C,C',C_0,C_1,\dots$ denote absolute positive constants. On the event where $V^\top Z_t$ fails to
be invertible, $W_t$ may be defined arbitrarily, as every quantity involving $W_t$ in the analysis is multiplied by $\mathbf 1_t$ or $\mathbf 1_{t-1}$, and by Lemma \ref{lem:welldef} $V^\top Z_t$ is invertible on theses events.

\subsection{Preliminaries}\label{sec:prelim}

\begin{definition}[Mixed norm]\label{def:Fp}
For a random matrix $X$ and $p\ge2$ set
$\displaystyle\|X\|_{F,p}:=\big(\mathbb E\|X\|_F^p\big)^{1/p}$.
\end{definition}

Both the Frobenius norm and the $p$-th power sit inside the expectation; $\|X\|_{F,p}$ is the
$L^p(\Omega)$ norm of the random variable $\|X\|_F$. In terms of singular values we have,
\[
  \|X\|_{F,p}^p=\mathbb E\Big(\sum_i\sigma_i(X)^2\Big)^{p/2},
\]
as opposed to
\[
\|X\|_{p,p}^p=\mathbb E\sum_i\sigma_i(X)^p,
\]
which is the norm used by~\citet{huang2021streaming}. In $\|\cdot\|_{p,p}$ the index $p$ serves both as the Schatten exponent inside and as the
moment exponent outside, while in $\|\cdot\|_{F,p}$ the inner norm is pinned at Frobenius and only
the outer moment index varies. The two agree at $p=2$. Decoupling the two indices is what allows a
Frobenius-norm conclusion (inner norm fixed) together with exponential tails (outer index large).

\begin{lemma}[Properties of $\|\cdot\|_{F,p}$]\label{lem:normprops}
Let $p\ge2$ and let $X,Y,Z$ be random matrices of the same size.
\begin{enumerate}
\item[(i)] $\|\cdot\|_{F,p}$ is a norm.
\item[(ii)] If $A,B$ are random matrices with $\|A\|_2\le a$ and $\|B\|_2\le b$ almost surely,
 then $\|AXB\|_{F,p}\le ab\,\|X\|_{F,p}$.
\item[(iii)] $\|X\mathbf 1_{G'}\|_{F,p}\le\|X\mathbf 1_G\|_{F,p}\le\|X\|_{F,p}$ whenever
 $G'\subseteq G$.
\item[(iv)] $\|X\|_{p,p}\le\|X\|_{F,p}$.
\item[(v)] $\mathbb P(\|X\|_F>s)\le s^{-p}\|X\|_{F,p}^p$ for every $s>0$.
\item[(vi)] \emph{(One-step smoothness.)} Let $\mathcal F$ be a $\sigma$-algebra, let $X$ be
 $\mathcal F$-measurable and let $\mathbb E[Y\mid\mathcal F]=0$. Then
 $\|X+Y\|_{F,p}^2\le\|X\|_{F,p}^2+2\cdot(p-1)\|Y\|_{F,p}^2$.
\item[(vii)] 
Under the hypotheses of (vi), for every $Z$ and every $\lambda > 0$:
\[
    \| X + Y + Z \|_{F,p}^2 \leq (1 + \lambda)(\|X \|_{F,p}^2 + 2(p-1)\|Y\|_{F,p}^2  + \lambda^{-1} \|Z\|_{F,p}^2)
\]
\item[(viii)] \emph{(Random class index.)} Let $\mathcal D$ be a finite set of deterministic
 matrices with $|\mathcal D|=m$, let $D$ be a random matrix taking values in $\mathcal D$, and let
 $Y$ be any random matrix of conforming size. Then
 $\|DY\|_{F,p}\le m^{1/p}\max_{D'\in\mathcal D}\|D'Y\|_{F,p}$; in particular
 $\|DY\|_{F,p}\le e\max_{D'\in\mathcal D}\|D'Y\|_{F,p}$ whenever $p\ge\log m$.
\end{enumerate}
\end{lemma}

\begin{proof}
(i) Note $\|X+Y\|_F\le\|X\|_F+\|Y\|_F$, so by monotonicity of $t\mapsto t^p$ and of the expectation, by Minkowski's inequality in $L^p(\Omega)$,
\[
  \|X+Y\|_{F,p}=\big(\mathbb E\|X+Y\|_F^p\big)^{1/p}
  \le\big(\mathbb E(\|X\|_F+\|Y\|_F)^p\big)^{1/p}
  \le\big(\mathbb E\|X\|_F^p\big)^{1/p}+\big(\mathbb E\|Y\|_F^p\big)^{1/p} .
\]
Homogeneity and definiteness are immediate.

(ii) holds point-wise from $\|AXB\|_F\le\|A\|_2\|X\|_F\|B\|_2$. \\ (iii) from $\|X\mathbf 1_G\|_F=\|X\|_F\mathbf 1_G$ and since $\mathbf{1}_G^p=\mathbf{1}_G$,
\[
\|X\mathbf{1}_G\|_{F,p}^p
= \mathbb{E}\big[\|X\|_F^p\,\mathbf{1}_G\big].
\]
This is monotone in the event: $G'\subseteq G$ implies
$\mathbf{1}_{G'}\leq\mathbf{1}_G$ pointwise, hence
\[
\mathbb{E}\big[\|X\|_F^p\,\mathbf{1}_{G'}\big]
\leq \mathbb{E}\big[\|X\|_F^p\,\mathbf{1}_G\big]
\leq \mathbb{E}\big[\|X\|_F^p\big].
\]
(iv) is $\|X\|_{S_p}\le\|X\|_{S_2}=\|X\|_F$ for $p\ge2$. \\(v) is Markov's
inequality applied to $\|X\|_F^p$.

(vi) follows from the $2$-smoothness of $L^p(\Omega;H)$ for a Hilbert space $H$; here
$H=\mathbb R^{(d-k)\times k}$ with the Frobenius inner product. Recall that a Banach space
$(\mathcal X,\|\cdot\|)$ is \emph{$(2,D)$-smooth} if
\[
  \|u+v\|^2+\|u-v\|^2\le2\|u\|^2+2D^2\|v\|^2
  \qquad\text{for all }u,v\in\mathcal X .
\]
A Hilbert space is $(2,1)$-smooth, with equality: this is the parallelogram law. By \citet[Proposition 2.1]{pinelis1994optimum}, $L^p$ is $(2,\sqrt{p-1})$-smooth for $p\ge2$. That proposition is stated for scalar-valued $L^p(T,\mathcal A,\nu)$, but its proof verifies the sufficient condition $(\|x\|^2)''(v,v)\le2D^2\|v\|^2$ by differentiating $\int_T|x|^p\mathrm d\nu$ twice and applying H\"older's inequality, and for $H$-valued integrands the same computation applies, Cauchy-Schwarz giving
$\frac{\mathrm d^2}{\mathrm ds^2}\|x+sv\|_H^p|_{s=0}\le p(p-1)\|x\|_H^{p-2}\|v\|_H^2$ where the
scalar case has equality. Hence $L^p(\Omega;H)$ is $(2,\sqrt{p-1})$-smooth, and taking $u=X$ and
$v=Y$,
\[
  \|X+Y\|_{F,p}^2\le2\|X\|_{F,p}^2+2(p-1)\|Y\|_{F,p}^2-\|X-Y\|_{F,p}^2 .
\]
Since $\mathbb E[Y\mid\mathcal F]=0$ and $X$ is $\mathcal F$-measurable, $-Y$ is again a
martingale increment and $\mathbb E[X-Y\mid\mathcal F]=X$, so conditional Jensen applied to the
convex map $u\mapsto\|u\|_F^p$ gives
\[
  \|X-Y\|_{F,p}^p=\mathbb E\|X-Y\|_F^p
  \ge\mathbb E\big\|\mathbb E[X-Y\mid\mathcal F]\big\|_F^p
  =\mathbb E\|X\|_F^p=\|X\|_{F,p}^p ,
\]
whence $\|X-Y\|_{F,p}\ge\|X\|_{F,p}$. Substituting gives (vi). At $p=2$ the mechanism is visible
directly:
\[
  \mathbb E\|X+Y\|_F^2=\mathbb E\|X\|_F^2+2\,\mathbb E\langle X,Y\rangle_F+\mathbb E\|Y\|_F^2,
  \qquad
  \mathbb E\langle X,Y\rangle_F=\mathbb E\big\langle X,\mathbb E[Y\mid\mathcal F]\big\rangle_F=0,
\]
the last step because $X$ is $\mathcal F$-measurable.

(vii): 
Using
\[ (a+b)^2 \leq (1+\lambda)\left(a^2+\lambda^{-1}b^2\right),
\]
we obtain
\begin{align*}
\norm{X+Y+Z}_{F,p}^{2}
&\leq
\left(\norm{X+Y}_{F,p}+\norm{Z}_{F,p}\right)^2\\
&\leq
(1+\lambda)
\left(
    \norm{X+Y}_{F,p}^{2}
    +\lambda^{-1}\norm{Z}_{F,p}^{2}
\right)\\
&\leq
(1+\lambda)
\left(
    \norm{X}_{F,p}^{2}
    +2(p-1)\norm{Y}_{F,p}^{2}
    +\lambda^{-1}\norm{Z}_{F,p}^{2}
\right),
\end{align*}
where the last inequality follows from (vi).

(viii): pointwise $\|DY\|_F^p\le\sum_{D'\in\mathcal D}\|D'Y\|_F^p$, since the left-hand side is
one of the summands and all are non-negative; taking expectations and the $p$-th root gives
$\|DY\|_{F,p}\le m^{1/p}\max_{D'}\|D'Y\|_{F,p}$, and $m^{1/p}=e^{(\log m)/p}\le e$ when
$p\ge\log m$. No measurability relation between $D$ and $Y$ is required. The factor $m^{1/p}$
cannot be removed: if $Y$ takes two values $y_1,y_2$ with probability $\frac12$ each and
$\mathcal D=\{D_1,D_2\}$ with $\|D_1y_1\|_F=\|D_2y_2\|_F=1$ and $\|D_1y_2\|_F=\|D_2y_1\|_F=0$,
then choosing $D=D_i$ on $\{Y=y_i\}$ gives $\|DY\|_{F,p}=1$ while
$\max_i\|D_iY\|_{F,p}=2^{-1/p}$.
\end{proof}

\subsection{The one-step decomposition}

\begin{lemma}[Frobenius bound on the $U$-block]\label{lem:Ublock}
Under Assumption C.2, almost surely $\|U^\top(C_t-\Sigma)V\|_F\le M$ and $\|C_t-\Sigma\|_2\le M$.
\end{lemma}

\begin{proof}
Since $U$ has orthonormal columns, $\|U^\top(C_t-\Sigma)V\|_F\le\|(C_t-\Sigma)V\|_F=
\|V^\top(C_t-\Sigma)\|_F$. As $V\in S_{d,k}$, the right-hand side is one of the quantities over
which the supremum defining $\|\cdot\|_{(2,k)}$ is taken, so it is at most
$\|C_t-\Sigma\|_{(2,k)}\le M$. The second claim is $\|X\|_2\le\|X\|_{(2,k)}$.
\end{proof}

\begin{lemma}[One-step decomposition, in factored form]\label{lem:onestep}
Fix $t\ge1$ and suppose $V^\top Z_{t-1}$ and $I-\Delta_t^2$ are invertible, where
\[
  H_t:=U^\top(I+\eta_t\Sigma)Z_{t-1}\big(V^\top(I+\eta_t\Sigma)Z_{t-1}\big)^{-1},\quad
  \begin{aligned}
   \Delta_t&:=\eta_tV^\top(C_t-\Sigma)Z_{t-1}\big(V^\top(I+\eta_t\Sigma)Z_{t-1}\big)^{-1},\\
   \widehat\Delta_t&:=\eta_tU^\top(C_t-\Sigma)Z_{t-1}\big(V^\top(I+\eta_t\Sigma)Z_{t-1}\big)^{-1}.
  \end{aligned}
\]
Then $V^\top Z_t=(I+\Delta_t)V^\top(I+\eta_t\Sigma)Z_{t-1}$,
\[
  W_t(I-\Delta_t^2)=H_t+J_{t,1}+J_{t,2},\qquad
  J_{t,1}:=\widehat\Delta_t-H_t\Delta_t,\quad J_{t,2}:=-\widehat\Delta_t\Delta_t ,
\]
and, writing $Z_{t-1}(V^\top Z_{t-1})^{-1}=V+UW_{t-1}$,
\[
  H_t=(I+\eta_t\Lambda_2)W_{t-1}(I+\eta_t\Lambda_1)^{-1},\quad
  \begin{aligned}
   \Delta_t&=\eta_tV^\top(C_t-\Sigma)(V+UW_{t-1})(I+\eta_t\Lambda_1)^{-1},\\
   \widehat\Delta_t&=\eta_tU^\top(C_t-\Sigma)(V+UW_{t-1})(I+\eta_t\Lambda_1)^{-1}.
  \end{aligned}
\]
\end{lemma}

\begin{proof}
The identity $W_t(I-\Delta_t^2)=H_t+J_{t,1}+J_{t,2}$ is Lemma 2.7 of
\citet{huang2021streaming}, proved in their Appendix G.1, where $J_{t,1}=\widehat\Delta_t
-H_t\Delta_t$ and $J_{t,2}=-\widehat\Delta_t\Delta_t$ are defined as above; their statement
records only that these two matrices have norms $O(\eta_t)$ and $O(\eta_t^2)$. The argument is
short, so we give it in full. The closed forms for $H_t$, $\Delta_t$ and $\widehat\Delta_t$ are
not stated there.

$V^\top Z_t=V^\top(I+\eta_tC_t)Z_{t-1}=V^\top(I+\eta_t\Sigma)Z_{t-1}+\eta_tV^\top(C_t-\Sigma)
Z_{t-1}=(I+\Delta_t)V^\top(I+\eta_t\Sigma)Z_{t-1}$, by the definition of $\Delta_t$. Hence
$(V^\top Z_t)^{-1}=(V^\top(I+\eta_t\Sigma)Z_{t-1})^{-1}(I+\Delta_t)^{-1}$, and since
$U^\top Z_t=U^\top(I+\eta_t\Sigma)Z_{t-1}+\eta_tU^\top(C_t-\Sigma)Z_{t-1}$,
\[
  W_t(I+\Delta_t)=U^\top Z_t\big(V^\top(I+\eta_t\Sigma)Z_{t-1}\big)^{-1}=H_t+\widehat\Delta_t .
\]
Multiplying on the right by $I-\Delta_t$ gives the decomposition.

For the closed forms, $U^\top(I+\eta_t\Sigma)=(I+\eta_t\Lambda_2)U^\top$ and
$V^\top(I+\eta_t\Sigma)=(I+\eta_t\Lambda_1)V^\top$, so $V^\top(I+\eta_t\Sigma)Z_{t-1}=
(I+\eta_t\Lambda_1)V^\top Z_{t-1}$; multiplying each of $U^\top(I+\eta_t\Sigma)Z_{t-1}$,
$\eta_tV^\top(C_t-\Sigma)Z_{t-1}$ and $\eta_tU^\top(C_t-\Sigma)Z_{t-1}$ on the right by
$(V^\top Z_{t-1})^{-1}(I+\eta_t\Lambda_1)^{-1}$ and using
$Z_{t-1}(V^\top Z_{t-1})^{-1}=V+UW_{t-1}$ gives the three expressions.
\end{proof}

\begin{lemma}[Contraction]\label{lem:contr}
Whenever $V^\top Z_{t-1}$ is invertible, $\|H_t\mathbf 1_{t-1}\|_F\le\chi_t
\|W_{t-1}\mathbf 1_{t-1}\|_F$ pointwise, and hence $\|H_t\mathbf 1_{t-1}\|_{F,p}\le\chi_t
\|W_{t-1}\mathbf 1_{t-1}\|_{F,p}$.
\end{lemma}

\begin{proof}
By \Cref{lem:onestep}, $H_t=(I+\eta_t\Lambda_2)W_{t-1}(I+\eta_t\Lambda_1)^{-1}$, so
$\|AXB\|_F\le\|A\|_2\|X\|_F\|B\|_2$ gives, pointwise,
\[
  \|H_t\mathbf 1_{t-1}\|_F\le\|I+\eta_t\Lambda_2\|_2\,
   \|(I+\eta_t\Lambda_1)^{-1}\|_2\,\|W_{t-1}\mathbf 1_{t-1}\|_F .
\]
Both $\Lambda_1$ and $\Lambda_2$ are diagonal with non-negative entries, since $\Sigma\succeq0$ by
Assumption C.1, so $\|I+\eta_t\Lambda_2\|_2=1+\eta_t\lambda_{k+1}$ and
$\|(I+\eta_t\Lambda_1)^{-1}\|_2=(1+\eta_t\lambda_k)^{-1}$, the largest entry of $\Lambda_2$ being
$\lambda_{k+1}$ and the smallest of $\Lambda_1$ being $\lambda_k$. Their product is $\chi_t$, and
taking $L^p$ norms gives the second claim.
\end{proof}

\begin{lemma}[Step bounds]\label{lem:steps}
Under the hypotheses of \Cref{lem:welldef}, put
$\mathcal E_t:=1+2\|W_{t-1}\mathbf 1_{t-1}\|_{F,p}$. Then
\begin{align}
  \|\widehat\Delta_t\mathbf 1_{t-1}\|_{F,p}&\le\eta_tM\big(1
   +\|W_{t-1}\mathbf 1_{t-1}\|_{F,p}\big),\label{eq:FDhat}\\
  \|J_{t,1}\mathbf 1_{t-1}\|_{F,p}&\le\eta_tM+\varepsilon_t
   \|W_{t-1}\mathbf 1_{t-1}\|_{F,p}\;\le\;\varepsilon_t\mathcal E_t,\label{eq:FJ1}\\
  \|J_{t,2}\mathbf 1_{t-1}\|_{F,p}&\le\varepsilon_t^2\mathcal E_t .\label{eq:FJ2}
\end{align}
\end{lemma}

\begin{proof}
All three bounds rest on one substitution. By \Cref{lem:welldef}, $V^\top Z_{t-1}$ is invertible on
$G_{t-1}$; since $UU^\top+VV^\top=I$ and $U^\top Z_{t-1}(V^\top Z_{t-1})^{-1}=W_{t-1}$,
\[
  Z_{t-1}=\big(VV^\top+UU^\top\big)Z_{t-1}=(V+UW_{t-1})\,V^\top Z_{t-1} ,
\]
while $V^\top(I+\eta_t\Sigma)U=0$ and $V^\top(I+\eta_t\Sigma)V=I+\eta_t\Lambda_1$ give
$V^\top(I+\eta_t\Sigma)Z_{t-1}=(I+\eta_t\Lambda_1)V^\top Z_{t-1}$. The right factor common to
$\Delta_t$ and $\widehat\Delta_t$ therefore collapses:
\[
  Z_{t-1}\big(V^\top(I+\eta_t\Sigma)Z_{t-1}\big)^{-1}
  =(V+UW_{t-1})\,V^\top Z_{t-1}\,(V^\top Z_{t-1})^{-1}(I+\eta_t\Lambda_1)^{-1}
  =(V+UW_{t-1})(I+\eta_t\Lambda_1)^{-1},
\]
which is the closed form of \Cref{lem:onestep}. Since $\Lambda_1\succeq0$ by Assumption C.1,
$\|(I+\eta_t\Lambda_1)^{-1}\|_2\le1$.

\eqref{eq:FDhat}: splitting $V+UW_{t-1}$ and discarding the right factor gives, pointwise,
\begin{align*}
  \|\widehat\Delta_t\mathbf 1_{t-1}\|_F
  &\le\eta_t\big\|U^\top(C_t-\Sigma)V\big\|_F
   +\eta_t\big\|U^\top(C_t-\Sigma)U\big\|_2\,\big\|W_{t-1}\mathbf 1_{t-1}\big\|_F\\
  &\le\eta_tM\big(1+\|W_{t-1}\mathbf 1_{t-1}\|_F\big),
\end{align*}
both terms by \Cref{lem:Ublock}. Taking $L^p$ norms gives \eqref{eq:FDhat}; in particular
$\|U^\top(C_t-\Sigma)V\|_{F,p}\le M$, since the bound holds almost surely.
\emph{This is the only step at which the factor $k^{1/p}$ of Huang et al.\ arises}: their
Proposition A.1 bounds the first term by $k^{1/p}\|C_t-\Sigma\|_2$ rather than by
\Cref{lem:Ublock}. That is a sharpening of Assumption C.2 and not a consequence of the choice of
norm; no product class stands in front of the block here, which is why the point arises at all.

\eqref{eq:FJ1}: since $J_{t,1}=\widehat\Delta_t-H_t\Delta_t$, pointwise
\begin{align*}
  \|J_{t,1}\mathbf 1_{t-1}\|_F
  &\le\|\widehat\Delta_t\mathbf 1_{t-1}\|_F
   +\|H_t\mathbf 1_{t-1}\|_F\,\|\Delta_t\mathbf 1_{t-1}\|_2\\
  &\le\eta_tM\big(1+\|W_{t-1}\mathbf 1_{t-1}\|_F\big)
   +\tfrac12\varepsilon_t\|W_{t-1}\mathbf 1_{t-1}\|_F ,
\end{align*}
by \Cref{lem:contr} with $\chi_t\le1$ and \eqref{eq:FD}. Taking $L^p$ norms and using
$\eta_tM\le\frac12\varepsilon_t$,
\[
  \|J_{t,1}\mathbf 1_{t-1}\|_{F,p}
  \le\eta_tM+\varepsilon_t\|W_{t-1}\mathbf 1_{t-1}\|_{F,p}
  \le\varepsilon_t\mathcal E_t .
\]

\eqref{eq:FJ2}: since $J_{t,2}=-\widehat\Delta_t\Delta_t$, pointwise
\[
  \|J_{t,2}\mathbf 1_{t-1}\|_F
  \le\|\widehat\Delta_t\mathbf 1_{t-1}\|_F\,\|\Delta_t\mathbf 1_{t-1}\|_2
  \le\tfrac12\varepsilon_t\big(1+\|W_{t-1}\mathbf 1_{t-1}\|_F\big)\cdot\tfrac12\varepsilon_t ,
\]
again by \eqref{eq:FD}, and taking $L^p$ norms gives
$\|J_{t,2}\mathbf 1_{t-1}\|_{F,p}\le\frac14\varepsilon_t^2\mathcal E_t\le\varepsilon_t^2
\mathcal E_t$.

No upper bound on $\varepsilon_t$ is used in \eqref{eq:FDhat}--\eqref{eq:FJ2}: they hold at every
step size. The condition \eqref{S1} enters only through \Cref{lem:welldef}, which supplies the
invertibility and \eqref{eq:FD}.
\end{proof}

\begin{lemma}[Well-definedness and the bound on $\Delta_t$]\label{lem:welldef}
Assume \eqref{S1} and $\|V^\top(C_t-\Sigma)UW_{t-1}\mathbf 1_{t-1}\|_2\le M\gamma$ almost surely,
for every $t\le T$. Then for every $t\le T$, on $G_{t-1}$ the matrix $V^\top Z_{t-1}$ is
invertible, so that $H_t$, $\Delta_t$ and $\widehat\Delta_t$ are defined and have the closed forms
of \Cref{lem:onestep};
\begin{equation}\label{eq:FD}
  \|\Delta_t\mathbf 1_{t-1}\|_2\le\tfrac12\varepsilon_t\le\varepsilon_t\le\tfrac12
  \qquad\text{almost surely};
\end{equation}
and $I-\Delta_t^2$ and $V^\top Z_t$ are invertible on $G_{t-1}$, with
$\|(I-\Delta_t^2)^{-1}\mathbf 1_{t-1}\|_2\le(1-\varepsilon_t^2)^{-1}$.
\end{lemma}

\begin{proof}
By induction on $t$. At $t=0$ the matrix $V^\top Z_0=V^\top Q_0$ is invertible almost surely.

Suppose $V^\top Z_{t-1}$ is invertible on $G_{t-1}$. Since $\Lambda_1\succeq0$ by Assumption C.1,
$V^\top(I+\eta_t\Sigma)Z_{t-1}=(I+\eta_t\Lambda_1)V^\top Z_{t-1}$ is invertible there as well, so
$H_t$, $\Delta_t$ and $\widehat\Delta_t$ are defined on $G_{t-1}$ and \Cref{lem:onestep} gives
$\Delta_t=\eta_tV^\top(C_t-\Sigma)(V+UW_{t-1})(I+\eta_t\Lambda_1)^{-1}$. Discarding the right
factor, whose operator norm is at most $1$, and splitting $V+UW_{t-1}$,
\[
  \|\Delta_t\mathbf 1_{t-1}\|_2
  \le\eta_t\big\|V^\top(C_t-\Sigma)V\big\|_2
   +\eta_t\big\|V^\top(C_t-\Sigma)UW_{t-1}\mathbf 1_{t-1}\big\|_2
  \le\eta_tM+\eta_tM\gamma=\tfrac12\varepsilon_t ,
\]
the first term by \Cref{lem:Ublock} and the second by hypothesis; with \eqref{S1} this is
\eqref{eq:FD}. In particular $\|\Delta_t\mathbf 1_{t-1}\|_2\le\frac12<1$, so \Cref{lem:neumann}
with $A=\Delta_t$, $G=G_{t-1}$ and $\varepsilon=\varepsilon_t$ shows that $I+\Delta_t$ and
$I-\Delta_t^2$ are invertible on $G_{t-1}$ and, by \eqref{eq:neu1}, that
$\|(I-\Delta_t^2)^{-1}\mathbf 1_{t-1}\|_2\le(1-\varepsilon_t^2)^{-1}$. Finally
$V^\top Z_t=(I+\Delta_t)V^\top(I+\eta_t\Sigma)Z_{t-1}$ by \Cref{lem:onestep} is a product of three
matrices each invertible on $G_{t-1}$, and $G_t\subseteq G_{t-1}$, so the induction continues.
\end{proof}

\begin{remark}
The base case is supplied by the initialisation: $Q_0=\mathrm{QR}(Q_0')$ with $Q_0'$ having
i.i.d.\ $\mathcal N(0,1)$ entries, so $V^\top Q_0=V^\top Q_0'R^{-1}$ with $V^\top Q_0'$ a
$k\times k$ standard Gaussian matrix, which is invertible almost surely.
\end{remark}

\subsection{The main recursive bound}\label{sec:recursion}

The two results below bound $\|W_t\mathbf 1_t\|_{F,p}$ itself and they are stated for an arbitrary decreasing sequence of good events $(G_t)$ and an
arbitrary threshold $\gamma$; each phase supplies its own events and its own $\gamma$, and with
them the hypothesis $\|V^\top(C_t-\Sigma)UW_{t-1}\mathbf 1_{t-1}\|_2\le M\gamma$. Phase I applies
\Cref{thm:F10} at a constant step size in \Cref{prop:F3}, Phase II at a decaying one in
\Cref{prop:phase2}.

\begin{theorem}[Unrolled recursion]\label{thm:F10}
Let $t\ge1$, set $s_t:=\sum_{i\le t}\eta_i$, and assume \eqref{S1}--\eqref{S3} and
$\|V^\top(C_i-\Sigma)UW_{i-1}\mathbf 1_{i-1}\|_2\le M\gamma$ almost surely, for every $i\le t$.
Then for every $p\ge2$,
\begin{equation}\label{eq:F5}
  \|W_t\mathbf 1_t\|_{F,p}^2\le\|W_t\mathbf 1_{t-1}\|_{F,p}^2
  \le e^{-s_t\rho_k}\|W_0\mathbf 1_0\|_{F,p}^2
   +C_1p\varepsilon_t^2\sum_{i=0}^{t-1}\|W_i\mathbf 1_i\|_{F,p}^2+C_2p\varepsilon_t^2t ,
\end{equation}
with $C_1=39$ and $C_2=10$; and if \eqref{S4} also holds for every $i\le t$, then
\begin{equation}\label{eq:F7}
  \|W_t\mathbf 1_t\|_{F,p}^2\le\|W_t\mathbf 1_{t-1}\|_{F,p}^2
  \le e^{-s_t\rho_k/2}\|W_0\mathbf 1_0\|_{F,p}^2+C_2p\varepsilon_t^2t .
\end{equation}
These are the two conclusions of Theorem 3.1 of Huang et al.\ with $k^{2/p}$ replaced by $1$.
\end{theorem}

\begin{proof}
\emph{Bounding $K_{1,i}$.} Expanding \Cref{prop:F10step},
\[
  K_{1,i}=\chi_i^2+16p\varepsilon_i^2+5\varepsilon_i^2\chi_i^2+80p\varepsilon_i^4 ,
\]
and $\chi_i\le1$ together with $\varepsilon_i^2\le\frac14$ from \eqref{S1} give
$5\varepsilon_i^2\chi_i^2\le5\varepsilon_i^2$ and $80p\varepsilon_i^4\le20p\varepsilon_i^2$, so
that
\[
  K_{1,i}\le\chi_i^2+(5+36p)\varepsilon_i^2 .
\]
For $\chi_i$, write $\lambda_{k+1}=\lambda_k-\rho_k$ and use $1+\eta_i\lambda_k\le2$ from
\eqref{S2}:
\[
  \chi_i=\frac{1+\eta_i\lambda_{k+1}}{1+\eta_i\lambda_k}
  =\frac{1+\eta_i\lambda_k-\eta_i\rho_k}{1+\eta_i\lambda_k}
  =1-\frac{\eta_i\rho_k}{1+\eta_i\lambda_k}
  \le1-\tfrac12\eta_i\rho_k ,
\]
whence $\chi_i^2\le(1-\frac12\eta_i\rho_k)^2\le e^{-\eta_i\rho_k}$. Finally $5+36p\le39p$ for
$p\ge2$, so $K_{1,i}\le e^{-\eta_i\rho_k}+C_1p\varepsilon_i^2$.
Condition \eqref{S2} in fact gives $\eta_i\lambda_k\le\frac12$, hence $1+\eta_i\lambda_k\le
\frac32$ and the sharper $\chi_i\le1-\frac23\eta_i\rho_k$, which is what the proof of Theorem
\ref{thm:F11} uses; the weaker form is kept here because it delivers the conclusion in exactly the
form of Theorem 3.1 of Huang et al.

\emph{Proof of \eqref{eq:F5}, by induction on $t$.} At $t=1$, \Cref{prop:F10step} gives
$\|W_1\mathbf 1_1\|_{F,p}^2\le K_{1,1}\|W_0\mathbf 1_0\|_{F,p}^2+K_{2,1}\le e^{-\eta_1\rho_k}
\|W_0\mathbf 1_0\|_{F,p}^2+C_1p\varepsilon_1^2\|W_0\mathbf 1_0\|_{F,p}^2+C_2p\varepsilon_1^2$.
Assume \eqref{eq:F5} at $t-1$. Then
\begin{align*}
  \|W_t\mathbf 1_t\|_{F,p}^2
  &\le e^{-\eta_t\rho_k}\|W_{t-1}\mathbf 1_{t-1}\|_{F,p}^2
    +C_1p\varepsilon_t^2\|W_{t-1}\mathbf 1_{t-1}\|_{F,p}^2+C_2p\varepsilon_t^2\\
  &\le e^{-\eta_t\rho_k}\Big(e^{-s_{t-1}\rho_k}\|W_0\mathbf 1_0\|_{F,p}^2
    +C_1p\varepsilon_{t-1}^2\!\!\sum_{i=0}^{t-2}\!\|W_i\mathbf 1_i\|_{F,p}^2
    +C_2p\varepsilon_{t-1}^2(t-1)\Big)\\
  &\qquad+C_1p\varepsilon_t^2\|W_{t-1}\mathbf 1_{t-1}\|_{F,p}^2+C_2p\varepsilon_t^2\\
  &\le e^{-s_t\rho_k}\|W_0\mathbf 1_0\|_{F,p}^2
    +C_1p\varepsilon_t^2\sum_{i=0}^{t-1}\|W_i\mathbf 1_i\|_{F,p}^2+C_2p\varepsilon_t^2t ,
\end{align*}
the last step using $e^{-\eta_t\rho_k}\varepsilon_{t-1}^2\le\varepsilon_t^2$, which is \eqref{S3}
squared together with $e^{-\eta_t\rho_k}\le e^{-\eta_t\rho_k/2}$.

\emph{Proof of \eqref{eq:F7}.} Here $\eta_i\rho_k\le\eta_i\lambda_1\le\frac12\le1$ by \eqref{S2},
so $\chi_i^2\le1-\eta_i\rho_k+\frac14(\eta_i\rho_k)^2\le1-\frac34\eta_i\rho_k$, and using
$1+x\le e^x$ twice,
\[
  K_{1,i}\le(1+5\varepsilon_i^2)\big(1-\tfrac34\eta_i\rho_k+16p\varepsilon_i^2\big)
  \le\exp\big(-\tfrac34\eta_i\rho_k+(5+16p)\varepsilon_i^2\big)\le e^{-\eta_i\rho_k/2},
\]
since $5+16p\le\frac{37}2p$ for $p\ge2$ and \eqref{S4} give $(5+16p)\varepsilon_i^2\le\frac14
\eta_i\rho_k$. The induction is then immediate, again by $e^{-\eta_t\rho_k/2}\varepsilon_{t-1}^2
\le\varepsilon_t^2$.
\end{proof}

\begin{proposition}[One-step recursion without product classes]\label{prop:F10step}
Assume \eqref{S1}, \eqref{S2} and $\|V^\top(C_t-\Sigma)UW_{t-1}\mathbf 1_{t-1}\|_2\le M\gamma$
almost surely. Then for $p\ge2$,
\[
  \|W_t\mathbf 1_t\|_{F,p}^2\le\|W_t\mathbf 1_{t-1}\|_{F,p}^2
  \le K_{1,t}\|W_{t-1}\mathbf 1_{t-1}\|_{F,p}^2+K_{2,t},
\]
\[
  K_{1,t}=(1+5\varepsilon_t^2)\big(\chi_t^2+16p\varepsilon_t^2\big),
  \qquad K_{2,t}=10p\varepsilon_t^2 .
\]
This is Proposition A.2 of Huang et al.\ with $K_{2,t}=5pk^{2/p}\varepsilon_t^2$ replaced by
$10p\varepsilon_t^2$.
\end{proposition}

\begin{proof}
The first inequality is \Cref{lem:normprops}(iii), since $G_t\subseteq G_{t-1}$.

By \Cref{lem:onestep},
\[
  W_t\mathbf 1_{t-1}(I-\Delta_t^2)
  =H_t\mathbf 1_{t-1}+J_{t,1}\mathbf 1_{t-1}+J_{t,2}\mathbf 1_{t-1},
  \qquad J_{t,1}=\widehat\Delta_t-H_t\Delta_t .
\]
Assumption C.1 gives $\mathbb E[C_t-\Sigma\mid\mathcal F_{t-1}]=0$, and both $\Delta_t$ and
$\widehat\Delta_t$ are linear in $C_t-\Sigma$ with $\mathcal F_{t-1}$-measurable coefficients, so
\[
  \mathbb E\big[\widehat\Delta_t\mid\mathcal F_{t-1}\big]
  =\eta_tU^\top\,\mathbb E\big[C_t-\Sigma\mid\mathcal F_{t-1}\big]\,
   Z_{t-1}\big(V^\top(I+\eta_t\Sigma)Z_{t-1}\big)^{-1}=0 ,
\]
and likewise $\mathbb E[\Delta_t\mid\mathcal F_{t-1}]=0$. The matrix $H_t$ is
$\mathcal F_{t-1}$-measurable, being a function of $Z_{t-1}$, as is $\mathbf 1_{t-1}$; hence
\[
  \mathbb E\big[J_{t,1}\mathbf 1_{t-1}\mid\mathcal F_{t-1}\big]
  =\Big(\mathbb E\big[\widehat\Delta_t\mid\mathcal F_{t-1}\big]
   -H_t\,\mathbb E\big[\Delta_t\mid\mathcal F_{t-1}\big]\Big)\mathbf 1_{t-1}=0 ,
\]
which is the hypothesis of \Cref{lem:normprops}(vii) with $\mathcal F=\mathcal F_{t-1}$ and
$X=H_t\mathbf 1_{t-1}$.

\Cref{lem:normprops}(vii) with $X=H_t\mathbf 1_{t-1}$,
$Y=J_{t,1}\mathbf 1_{t-1}$ and $Z=J_{t,2}\mathbf 1_{t-1}$ gives, for every $\lambda>0$,
\[
  \|W_t\mathbf 1_{t-1}(I-\Delta_t^2)\|_{F,p}^2
  \le(1+\lambda)\Big(\|H_t\mathbf 1_{t-1}\|_{F,p}^2
   +2(p-1)\|J_{t,1}\mathbf 1_{t-1}\|_{F,p}^2
   +\lambda^{-1}\|J_{t,2}\mathbf 1_{t-1}\|_{F,p}^2\Big) ,
\]
and inserting $\|J_{t,1}\mathbf 1_{t-1}\|_{F,p}\le\varepsilon_t\mathcal E_t$ and
$\|J_{t,2}\mathbf 1_{t-1}\|_{F,p}\le\varepsilon_t^2\mathcal E_t$ from \Cref{lem:steps},
\[
  \|W_t\mathbf 1_{t-1}(I-\Delta_t^2)\|_{F,p}^2
  \le(1+\lambda)\Big(\|H_t\mathbf 1_{t-1}\|_{F,p}^2
   +2(p-1)\varepsilon_t^2\mathcal E_t^2+\lambda^{-1}\varepsilon_t^4\mathcal E_t^2\Big) .
\]
Choosing $\lambda=\varepsilon_t^2$ collapses the last two terms, since $2(p-1)+1=2p-1\le2p$:
\[
  \|W_t\mathbf 1_{t-1}(I-\Delta_t^2)\|_{F,p}^2
  \le(1+\varepsilon_t^2)\Big(\|H_t\mathbf 1_{t-1}\|_{F,p}^2
   +2p\,\mathcal E_t^2\varepsilon_t^2\Big).
\]

By \eqref{eq:FD} and \eqref{S1},
$\|\Delta_t\mathbf 1_{t-1}\|_2\le\varepsilon_t\le\frac12$, and
$W_t\mathbf 1_{t-1}=(W_t\mathbf 1_{t-1})\mathbf 1_{t-1}$, so \eqref{eq:neu3} of
\Cref{lem:neumann} with $A=\Delta_t$ and $G=G_{t-1}$ gives, pointwise and hence in
$\|\cdot\|_{F,p}$,
\[
  \|W_t\mathbf 1_{t-1}\|_{F,p}\le(1-\varepsilon_t^2)^{-1}
   \|W_t\mathbf 1_{t-1}(I-\Delta_t^2)\|_{F,p} .
\]
Squaring and combining,
\[
  \|W_t\mathbf 1_{t-1}\|_{F,p}^2
  \le\frac{1+\varepsilon_t^2}{(1-\varepsilon_t^2)^2}
   \Big(\|H_t\mathbf 1_{t-1}\|_{F,p}^2+2p\,\mathcal E_t^2\varepsilon_t^2\Big)
  \le(1+5\varepsilon_t^2)
   \Big(\|H_t\mathbf 1_{t-1}\|_{F,p}^2+2p\,\mathcal E_t^2\varepsilon_t^2\Big),
\]
the last step because $(1+\varepsilon^2)(1-\varepsilon^2)^{-2}\le1+5\varepsilon^2$ for
$\varepsilon\le\frac12$.

\Cref{lem:contr} gives $\|H_t\mathbf 1_{t-1}\|_{F,p}^2\le\chi_t^2
\|W_{t-1}\mathbf 1_{t-1}\|_{F,p}^2$, and $(a+b)^2\le2a^2+2b^2$ applied to
$\mathcal E_t=1+2\|W_{t-1}\mathbf 1_{t-1}\|_{F,p}$ gives
$\mathcal E_t^2\le2+8\|W_{t-1}\mathbf 1_{t-1}\|_{F,p}^2$. Hence
\[
  \|W_t\mathbf 1_{t-1}\|_{F,p}^2
  \le(1+5\varepsilon_t^2)\big(\chi_t^2+16p\varepsilon_t^2\big)
   \|W_{t-1}\mathbf 1_{t-1}\|_{F,p}^2+4(1+5\varepsilon_t^2)p\varepsilon_t^2 ,
\]
and $4(1+5\varepsilon_t^2)\le10$ for $\varepsilon_t\le\frac12$, which is $K_{2,t}$.
\end{proof}

\subsection{Phase I}\label{sec:phase1}

Phase I uses a constant step size $\eta$, a threshold $\gamma\ge2$, and the product-class
construction of Section 5 of \citet{huang2021streaming}, which we recall. Suppose there is a
deterministic finite set $\mathcal A_\star\subseteq\mathbb R^{d\times d}$ with $C_t\in
\mathcal A_\star$ almost surely for every $t\le T_0$, and put $m:=|\mathcal A_\star|$. Set
\[
  \mathcal E:=\{M^{-1}(A-\Sigma)UU^\top:A\in\mathcal A_\star\},
\]
\[
  \mathcal E_{r,\ell}:=\big\{V^\top F_1\cdots F_rU:\ F_i\in\mathcal E\ \text{for at most $\ell$
   distinct $i\in[r]$, and}\ F_i=(1+\eta\lambda_{k+1})^{-1}(I+\eta\Sigma)UU^\top
   \ \text{otherwise}\big\},
\]

\begin{lemma}[Properties of the product classes]\label{lem:classprops}
Let $r,\ell\ge1$ and let $E=V^\top F_1\cdots F_rU\in\mathcal E_{r,\ell}$. Then
\[
  \|E\|_F\le1,
\]
\begin{equation}\label{eq:absorb}
  EU^\top=V^\top F_1\cdots F_rUU^\top=V^\top F_1\cdots F_r ,
\end{equation}
and $\mathcal E_{r,\ell}\subseteq\mathcal E_{r,\ell+1}$.
\end{lemma}

\begin{proof}
\emph{Every factor is a contraction in operator norm.} For a noise factor,
$\|M^{-1}(A-\Sigma)UU^\top\|_2\le M^{-1}\|A-\Sigma\|_2\le1$ by Assumption C.2; for the drift
factor, $(1+\eta\lambda_{k+1})^{-1}\|(I+\eta\Sigma)UU^\top\|_2=1$, since $\Sigma\succeq0$ and its
largest eigenvalue on $\mathrm{range}(U)$ is $\lambda_{k+1}$.

\emph{The leading factor is a contraction in Frobenius norm.} If $F_1$ is a noise factor then
\[
  \|V^\top F_1\|_F=M^{-1}\|V^\top(A-\Sigma)UU^\top\|_F\le M^{-1}\|V^\top(A-\Sigma)\|_F\le1
\]
by Assumption C.2; and if $F_1$ is the drift factor then $V^\top F_1=0$, since
$V^\top\Sigma U=\Lambda_1V^\top U=0$ and $V^\top U=0$.

\emph{The bound.} Splitting $E=\big(V^\top F_1\big)\big(F_2\cdots F_rU\big)$ and applying
\Cref{lem:normprops}(ii) to the second group,
\[
  \|E\|_F\le\|V^\top F_1\|_F\,\prod_{i\ge2}\|F_i\|_2\,\|U\|_2\le1 .
\]
One factor has to be bounded in Frobenius norm: operator-norm bounds alone would give only
$\|E\|_F\le\sqrt k\,\|E\|_2$.

Every factor ends in $UU^\top$, and $U^\top U=I_{d-k}$ gives $F_rUU^\top=F_r$,
which is \eqref{eq:absorb}. And finally an element with at most $\ell$ noise factors has at most $\ell+1$.
\end{proof}

Define the Phase I good events
\[
  G_t:=G_{t-1}\cap\Big\{\max_{E\in\mathcal E}\|V^\top EUW_t\|_2\le\gamma\Big\},\qquad t\ge1 ,
\]
with $G_0$ the initialisation event
\[
  G_0:=\bigcap_{r,\ell=1}^{T_0+1}\Big\{\max_{E\in\mathcal E_{r,\ell}}\|EW_0\|_F\le
   \sqrt\ell\,\gamma_{\rm init}\Big\}\cap\big\{\|W_0\|_F\le\sqrt d\,\gamma\big\},
  \qquad \gamma_{\rm init}:=\frac\gamma{e\sqrt2} .
\]

These events satisfy $\|V^\top(C_t-\Sigma)UW_{t-1}\mathbf 1_{t-1}\|_2\le M\gamma$ almost surely,
which is needed for the main recursive bound~(\Cref{thm:F10}): for
$F_t:=M^{-1}(C_t-\Sigma)UU^\top\in\mathcal E$ one has $\|V^\top F_tUW_{t-1}\|_2\le\gamma$ on
$G_{t-1}$.

\begin{proposition}[One-step product estimate]\label{prop:F1}
Assume \eqref{S1}, \eqref{S2}, let $E\in\mathcal E_{r,\ell}$ and let $p\ge\max\{2,\log m\}$. Put
\[
  \overline E_t:=1+2e\max_{E''\in\mathcal E_{r+1,\ell+1}}
   \|E''W_{t-1}\mathbf 1_{t-1}\|_{F,p} .
\]
Then $EW_t(I-\Delta_t^2)=EH_t+EJ_{t,1}+EJ_{t,2}$, and
\begin{align}
  \|EJ_{t,1}\mathbf 1_{t-1}\|_{F,p}&\le\overline E_t\varepsilon , \label{eq:F1a}\\
  \|EJ_{t,2}\mathbf 1_{t-1}\|_{F,p}&\le\overline E_t\varepsilon^2 , \label{eq:F1b}\\
  \|EH_t\mathbf 1_{t-1}\|_{F,p}&\le\chi\,\|E'W_{t-1}\mathbf 1_{t-1}\|_{F,p}
   \quad\text{for }E':=(1+\eta\lambda_{k+1})^{-1}EU^\top(I+\eta\Sigma)U
   \in\mathcal E_{r+1,\ell} , \label{eq:F1c}\\
  \mathbb E\big[EJ_{t,1}\mathbf 1_{t-1}\mid\mathcal F_{t-1}\big]&=0 . \label{eq:F1d}
\end{align}
\end{proposition}

\begin{proof}
The decomposition is the identity of \Cref{lem:onestep} multiplied on the left by $E$.

\eqref{eq:F1c}: by \Cref{lem:onestep}, $H_t=(I+\eta\Lambda_2)W_{t-1}(I+\eta\Lambda_1)^{-1}$. The
columns of $U$ are the eigenvectors of $\Sigma$ with eigenvalues $\lambda_{k+1},\dots,\lambda_d$,
so $U^\top\Sigma U=\Lambda_2$ and $U^\top U=I_{d-k}$, giving $I+\eta\Lambda_2=U^\top(I+\eta\Sigma)U$
and
\[
  EH_t=\underbrace{EU^\top(I+\eta\Sigma)U}_{=(1+\eta\lambda_{k+1})E'}\,
   W_{t-1}(I+\eta\Lambda_1)^{-1} .
\]
Since $\Lambda_1\succeq0$ with smallest entry $\lambda_k$, $\|(I+\eta\Lambda_1)^{-1}\|_2
=(1+\eta\lambda_k)^{-1}$, so \Cref{lem:normprops}(ii) yields
\[
  \|EH_t\mathbf 1_{t-1}\|_{F,p}\le\frac{1+\eta\lambda_{k+1}}{1+\eta\lambda_k}
   \|E'W_{t-1}\mathbf 1_{t-1}\|_{F,p}=\chi\,\|E'W_{t-1}\mathbf 1_{t-1}\|_{F,p} .
\]
That $E'$ lies in $\mathcal E_{r+1,\ell}$ follows from \eqref{eq:absorb}: with
$F_{r+1}:=(1+\eta\lambda_{k+1})^{-1}(I+\eta\Sigma)UU^\top$, a drift factor of the class,
$V^\top F_1\cdots F_rF_{r+1}U=(1+\eta\lambda_{k+1})^{-1}EU^\top(I+\eta\Sigma)U=E'$, a product of
$r+1$ factors whose new factor is not a noise factor, so the count is still at most $\ell$.

\eqref{eq:F1a}: by \Cref{lem:onestep}, $J_{t,1}=\widehat\Delta_t-H_t\Delta_t$, so
\[
  \|EJ_{t,1}\mathbf 1_{t-1}\|_{F,p}
  \le\underbrace{\|E\widehat\Delta_t\mathbf 1_{t-1}\|_{F,p}}_{(\ast)}
   +\underbrace{\|EH_t\mathbf 1_{t-1}\|_{F,p}}_{(\ast\ast)}\,
    \|\Delta_t\mathbf 1_{t-1}\|_2 .
\]
For $(\ast)$, the closed form $\widehat\Delta_t=\eta U^\top(C_t-\Sigma)(V+UW_{t-1})
(I+\eta\Lambda_1)^{-1}$ of \Cref{lem:onestep} gives
\[
  (\ast)\le\eta\Big(\|EU^\top(C_t-\Sigma)V\|_{F,p}
   +\|EU^\top(C_t-\Sigma)UW_{t-1}\mathbf 1_{t-1}\|_{F,p}\Big),
\]
the right factor having operator norm at most $1$ and being discarded by
\Cref{lem:normprops}(ii). The first summand is at most $M$, since $\|E\|_2\le\|E\|_F\le1$ by
\Cref{lem:classprops} and $\|U^\top(C_t-\Sigma)V\|_F\le M$ almost surely by \Cref{lem:Ublock}.
For the second, put $F_{r+1}:=M^{-1}(C_t-\Sigma)UU^\top$, which lies in $\mathcal E$ because
$C_t\in\mathcal A_\star$ almost surely, and $E'':=V^\top F_1\cdots F_rF_{r+1}U$; then
\eqref{eq:absorb} and $U^\top U=I_{d-k}$ give
\[
  E''=M^{-1}V^\top F_1\cdots F_r(C_t-\Sigma)UU^\top U=M^{-1}EU^\top(C_t-\Sigma)U
  \ \in\ \mathcal E_{r+1,\ell+1},
\]
the new factor being a noise factor, so the count rises from $\ell$ to $\ell+1$, and hence
$EU^\top(C_t-\Sigma)UW_{t-1}=M\,E''W_{t-1}$. Here $E''$ is \emph{random}: as $C_t$ ranges over
$\mathcal A_\star$ it takes $m$ values, all of them in the deterministic class
$\mathcal E_{r+1,\ell+1}$. \Cref{lem:normprops}(viii) with
$\mathcal D=\{M^{-1}EU^\top(A-\Sigma)U:A\in\mathcal A_\star\}$ and
$Y=W_{t-1}\mathbf 1_{t-1}$ therefore gives
\begin{equation}\label{eq:randidx}
  \|E''W_{t-1}\mathbf 1_{t-1}\|_{F,p}\le m^{1/p}\max_{E'''\in\mathcal E_{r+1,\ell+1}}
   \|E'''W_{t-1}\mathbf 1_{t-1}\|_{F,p}\le e\max_{E'''\in\mathcal E_{r+1,\ell+1}}
   \|E'''W_{t-1}\mathbf 1_{t-1}\|_{F,p},
\end{equation}
using $p\ge\log m$; the maximum is over a finite set of \emph{deterministic} matrices, so
$\overline E_t$ is deterministic. Combining, and using $\eta M\le\frac12\varepsilon$ from
$\varepsilon=2\eta M(1+\gamma)$ with $\gamma\ge1$,
\[
  (\ast)\le\eta M\Big(1+e\max_{E'''}\|E'''W_{t-1}\mathbf 1_{t-1}\|_{F,p}\Big)
  \le\tfrac12\varepsilon\,\overline E_t .
\]
For $(\ast\ast)$, since $\chi\le1$ by \Cref{lem:contr} and
$\mathcal E_{r+1,\ell}\subseteq\mathcal E_{r+1,\ell+1}$ by \Cref{lem:classprops},
\[
  (\ast\ast)\le\chi\max_{E'''\in\mathcal E_{r+1,\ell+1}}\|E'''W_{t-1}\mathbf 1_{t-1}\|_{F,p}
  \le e\max_{E'''\in\mathcal E_{r+1,\ell+1}}\|E'''W_{t-1}\mathbf 1_{t-1}\|_{F,p}
  =\frac{\overline E_t-1}2\le\tfrac12\overline E_t .
\]
With $\|\Delta_t\mathbf 1_{t-1}\|_2\le\varepsilon$ from \eqref{eq:FD} and
\Cref{lem:normprops}(ii),
\[
  \|EJ_{t,1}\mathbf 1_{t-1}\|_{F,p}
  \le\tfrac12\varepsilon\overline E_t+\tfrac12\overline E_t\varepsilon=\overline E_t\varepsilon .
\]

\eqref{eq:F1b}: by \Cref{lem:onestep}, $J_{t,2}=-\widehat\Delta_t\Delta_t$, so
\[
  \|EJ_{t,2}\mathbf 1_{t-1}\|_{F,p}
  \le\underbrace{\|E\widehat\Delta_t\mathbf 1_{t-1}\|_{F,p}}_{\le\frac12\varepsilon\overline E_t}
   \underbrace{\|\Delta_t\mathbf 1_{t-1}\|_2}_{\le\varepsilon}
  \le\overline E_t\varepsilon^2 .
\]

\eqref{eq:F1d}: $E$ is deterministic, and $H_t$, $W_{t-1}$ and $\mathbf 1_{t-1}$ are
$\mathcal F_{t-1}$-measurable, so
\[
  \mathbb E\big[EJ_{t,1}\mathbf 1_{t-1}\mid\mathcal F_{t-1}\big]
  =E\Big(\mathbb E\big[\widehat\Delta_t\mid\mathcal F_{t-1}\big]
   -H_t\,\mathbb E\big[\Delta_t\mid\mathcal F_{t-1}\big]\Big)\mathbf 1_{t-1} .
\]
Both $\widehat\Delta_t$ and $\Delta_t$ are linear in $C_t-\Sigma$ with
$\mathcal F_{t-1}$-measurable coefficients, so by Assumption C.1,
\[
  \mathbb E\big[\widehat\Delta_t\mid\mathcal F_{t-1}\big]
  =\eta\,U^\top\underbrace{\mathbb E\big[C_t-\Sigma\mid\mathcal F_{t-1}\big]}_{=0}
   Z_{t-1}\big(V^\top(I+\eta\Sigma)Z_{t-1}\big)^{-1}=0 ,
\]
and likewise $\mathbb E[\Delta_t\mid\mathcal F_{t-1}]=0$.
\end{proof}

\begin{proposition}[One-step recursion with product classes]\label{prop:F2}
Under the hypotheses of \Cref{prop:F1}, for every $t\le T_0$,
\[
  \max_{E\in\mathcal E_{r,\ell}}\|EW_t\mathbf 1_{t-1}\|_{F,p}^2
  \le\overline K_1\!\!\max_{E'\in\mathcal E_{r+1,\ell}}\!\!\|E'W_{t-1}\mathbf 1_{t-1}\|_{F,p}^2
   +\overline K_2\!\!\max_{E''\in\mathcal E_{r+1,\ell+1}}\!\!\|E''W_{t-1}\mathbf 1_{t-1}\|_{F,p}^2
   +\overline K_2 ,
\]
with $\overline K_1=(1+5\varepsilon^2)\chi^2$ and $\overline K_2=(1+5\varepsilon^2)16e^2p
\varepsilon^2$. These are the constants of Huang et al.\ with $\overline K_2$ multiplied by two
independent factors: $e^2$, supplied by \eqref{eq:randidx}, and $2$, from the proof of
\Cref{lem:normprops}(vi).
\end{proposition}

\begin{proof}
The argument is that of \Cref{prop:F10step}, with \Cref{prop:F1} in place of \Cref{lem:steps}.
Fix $E\in\mathcal E_{r,\ell}$. By \Cref{prop:F1},
\[
  \|EW_t\mathbf 1_{t-1}(I-\Delta_t^2)\|_{F,p}^2
  =\|EH_t\mathbf 1_{t-1}+EJ_{t,1}\mathbf 1_{t-1}+EJ_{t,2}\mathbf 1_{t-1}\|_{F,p}^2 ,
\]
where $EH_t\mathbf 1_{t-1}$ is $\mathcal F_{t-1}$-measurable and
$\mathbb E[EJ_{t,1}\mathbf 1_{t-1}\mid\mathcal F_{t-1}]=0$ by \eqref{eq:F1d}, so
\Cref{lem:normprops}(vii) applies with $\lambda=\varepsilon^2$, and \eqref{eq:F1a} and
\eqref{eq:F1b} bound the two fluctuation terms:
\begin{align*}
  \|EW_t\mathbf 1_{t-1}(I-\Delta_t^2)\|_{F,p}^2
  &\le(1+\varepsilon^2)\Big(\|EH_t\mathbf 1_{t-1}\|_{F,p}^2
   +2(p-1)\underbrace{\|EJ_{t,1}\mathbf 1_{t-1}\|_{F,p}^2}_{\le\overline E_t^{\,2}\varepsilon^2}
   +\frac1{\varepsilon^2}
    \underbrace{\|EJ_{t,2}\mathbf 1_{t-1}\|_{F,p}^2}_{\le\overline E_t^{\,2}\varepsilon^4}\Big)\\
  &\le(1+\varepsilon^2)\Big(\|EH_t\mathbf 1_{t-1}\|_{F,p}^2
   +2p\,\overline E_t^{\,2}\varepsilon^2\Big),
\end{align*}
since $2(p-1)+1=2p-1\le2p$.

Note $\|\Delta_t\mathbf 1_{t-1}\|_2\le\varepsilon\le\frac12$ by \eqref{eq:FD} and \eqref{S1}, and
$EW_t\mathbf 1_{t-1}=(EW_t\mathbf 1_{t-1})\mathbf 1_{t-1}$, so \eqref{eq:neu3} of
\Cref{lem:neumann} with $A=\Delta_t$ and $G=G_{t-1}$ gives
$\|EW_t\mathbf 1_{t-1}\|_{F,p}\le(1-\varepsilon^2)^{-1}
\|EW_t\mathbf 1_{t-1}(I-\Delta_t^2)\|_{F,p}$, whence
\[
  \|EW_t\mathbf 1_{t-1}\|_{F,p}^2
  \le\underbrace{\frac{1+\varepsilon^2}{(1-\varepsilon^2)^2}}_{\le\,1+5\varepsilon^2}
   \Big(\|EH_t\mathbf 1_{t-1}\|_{F,p}^2+2p\,\overline E_t^{\,2}\varepsilon^2\Big) .
\]

Now $\overline E_t=1+2e\max_{E''\in\mathcal E_{r+1,\ell+1}}\|E''W_{t-1}\mathbf 1_{t-1}\|_{F,p}$,
so $(a+b)^2\le2a^2+2b^2$ gives
\[
  \overline E_t^{\,2}\le2+8e^2\!\!\max_{E''\in\mathcal E_{r+1,\ell+1}}\!\!
   \|E''W_{t-1}\mathbf 1_{t-1}\|_{F,p}^2 ,
\]
and therefore
\[
  (1+5\varepsilon^2)2p\,\overline E_t^{\,2}\varepsilon^2
  \le(1+5\varepsilon^2)16e^2p\varepsilon^2\!\!\max_{E''\in\mathcal E_{r+1,\ell+1}}\!\!
   \|E''W_{t-1}\mathbf 1_{t-1}\|_{F,p}^2+4(1+5\varepsilon^2)p\varepsilon^2
  \le\overline K_2\!\!\max_{E''\in\mathcal E_{r+1,\ell+1}}\!\!
   \|E''W_{t-1}\mathbf 1_{t-1}\|_{F,p}^2+\overline K_2 ,
\]
the last step because $4\le16e^2$.

Finally $\|EH_t\mathbf 1_{t-1}\|_{F,p}^2\le\chi^2\max_{E'\in\mathcal E_{r+1,\ell}}
\|E'W_{t-1}\mathbf 1_{t-1}\|_{F,p}^2$ by \eqref{eq:F1c}. The right-hand side does not depend on
$E\in\mathcal E_{r,\ell}$, so the maximum over $E$ may be taken on the left.
\end{proof}

\begin{theorem}[Iteration of the product-class recursion]\label{thm:F11}
Assume \eqref{S1}, \eqref{S2}, \eqref{S4} with constant step size, $p\ge\max\{2,\log m\}$,
$\gamma\ge2$, and that on $G_0$ one has $\max_{1\le r,\ell\le T_0+1}\max_{E\in\mathcal E_{r,\ell}}
\|EW_0\|_F\le\sqrt\ell\,\gamma_{\rm init}$. Then for every $1\le t\le T_0$ and every
$1\le r,\ell\le T_0-t+1$,
\[
  \max_{E\in\mathcal E_{r,\ell}}\|EW_t\mathbf 1_t\|_{F,p}^2
  \le\max_{E\in\mathcal E_{r,\ell}}\|EW_t\mathbf 1_{t-1}\|_{F,p}^2
  \le\ell\gamma_{\rm init}^2e^{-t\eta\rho_k/2}+12e^2p\gamma^2\varepsilon^2t .
\]
\end{theorem}

\begin{proof}
The first inequality is \Cref{lem:normprops}(iii). Writing
$B_{t,r,\ell}:=\max_{E\in\mathcal E_{r,\ell}}\|EW_t\mathbf 1_{t-1}\|_{F,p}^2$ and
$A_{t,r,\ell}:=\max_{E\in\mathcal E_{r,\ell}}\|EW_t\mathbf 1_t\|_{F,p}^2\le B_{t,r,\ell}$,
\Cref{prop:F2} reads
\[
  B_{t,r,\ell}\le\overline K_1A_{t-1,r+1,\ell}+\overline K_2A_{t-1,r+1,\ell+1}+\overline K_2 ,
  \qquad
  \overline K_1=(1+5\varepsilon^2)\chi^2,\quad
  \overline K_2=(1+5\varepsilon^2)16e^2p\varepsilon^2 .
\]

Put $x:=\eta\rho_k$, so $x\le\frac12$ by \eqref{S2}. As in the proof of \Cref{thm:F10},
$\chi=1-\frac{\eta\rho_k}{1+\eta\lambda_k}$, and \eqref{S2} in the form $\eta\lambda_k\le\frac12$
gives $1+\eta\lambda_k\le\frac32$, so $\chi\le1-\frac23x$. Further, \eqref{S4} with $p\ge2$ gives
$\varepsilon^2\le x/800$, hence
\[
  16e^2p\varepsilon^2\le\frac{16e^2}{400}\,x\le\tfrac3{10}x ,
  \qquad
  1+5\varepsilon^2\le e^{5\varepsilon^2}\le e^{x/160} ,
\]
and for $x\le\frac12$, $(1-\frac23x)^2\le e^{-4x/3}\le1-\frac43x+\frac89x^2\le1-\frac89x$. Hence
\[
  \overline K_1+\overline K_2=(1+5\varepsilon^2)\big(\chi^2+16e^2p\varepsilon^2\big)
   \le e^{x/160}\big(1-\tfrac89x+\tfrac3{10}x\big)
   \le e^{x/160}e^{-(8/9-3/10)x}\le e^{-x/2}=e^{-\eta\rho_k/2}\le1 ,
\]
using $1-u\le e^{-u}$ and $\frac89-\frac3{10}-\frac1{160}\ge\frac12$; and
$\overline K_2\le\frac94\cdot16e^2p\varepsilon^2=36e^2p\varepsilon^2$ from $\varepsilon\le\frac12$.

We show by induction that $B_{t,r,\ell}\le\ell\gamma_{\rm init}^2e^{-t\eta\rho_k/2}
+\frac{\gamma^2}3\overline K_2t$ for all $1\le r,\ell\le T_0-t+1$.

At $t=1$, the hypothesis on $G_0$ gives
\[
  B_{1,r,\ell}
  \le\overline K_1\underbrace{\max_{E'\in\mathcal E_{r+1,\ell}}
    \|E'W_0\mathbf 1_0\|_{F,p}^2}_{\le\,\ell\gamma_{\rm init}^2}
   +\overline K_2\underbrace{\max_{E''\in\mathcal E_{r+1,\ell+1}}
    \|E''W_0\mathbf 1_0\|_{F,p}^2}_{\le\,(\ell+1)\gamma_{\rm init}^2}+\overline K_2
  \le\ell\gamma_{\rm init}^2\big(\overline K_1+\overline K_2\big)
   +\big(1+\gamma_{\rm init}^2\big)\overline K_2 ,
\]
and $\gamma\ge2$ gives $1+\gamma_{\rm init}^2=1+\gamma^2/2e^2\le\gamma^2/3$, which is the claim.

For $t>1$, note $r+1,\ell+1\le T_0-(t-1)+1$, so the induction hypothesis applies to both terms on
the right, and
\begin{align*}
  B_{t,r,\ell}&\le\overline K_1\Big(\ell\gamma_{\rm init}^2e^{-(t-1)\eta\rho_k/2}
   +\tfrac{\gamma^2}3\overline K_2(t-1)\Big)
   +\overline K_2\Big((\ell+1)\gamma_{\rm init}^2e^{-(t-1)\eta\rho_k/2}
   +\tfrac{\gamma^2}3\overline K_2(t-1)\Big)+\overline K_2\\
  &\le\ell\gamma_{\rm init}^2\big(\overline K_1+\overline K_2\big)e^{-(t-1)\eta\rho_k/2}
   +\big(1+\gamma_{\rm init}^2\big)\overline K_2
   +\tfrac{\gamma^2}3\overline K_2(t-1)\big(\overline K_1+\overline K_2\big)\\
  &\le\ell\gamma_{\rm init}^2e^{-t\eta\rho_k/2}+\tfrac{\gamma^2}3\overline K_2t ,
\end{align*}
using $\overline K_1+\overline K_2\le e^{-\eta\rho_k/2}\le1$ and $1+\gamma_{\rm init}^2\le
\gamma^2/3$ again. Finally $\frac{\gamma^2}3\overline K_2t\le12e^2p\gamma^2\varepsilon^2t$.
\end{proof}

\begin{proposition}[Phase I moment estimates]\label{prop:F3}
Let the updates $C_1,\dots,C_{T_0}$ satisfy Assumption C and lie almost surely in a deterministic
finite set $\mathcal A_\star$ with $m:=|\mathcal A_\star|$, let the step size be constant equal to
$\eta$, let $\gamma\ge2$, let $\mathcal E$ and $\mathcal E_{r,\ell}$ be the classes built from
$\mathcal A_\star$ at the start of this section, and let $(G_t)_{t\ge0}$ be the good events defined
there with $\gamma_{\rm init}=\gamma/(e\sqrt2)$. Let $\delta\in(0,1)$, put
\[
  p_0:=\max\Big\{2,\ \log\frac6\delta\Big\},\qquad
  p_1:=\max\Big\{2,\ \log\frac{12T_0(m+1)}\delta\Big\},\qquad
  p_\star:=\max\{p_0,p_1\},
\]
and assume \eqref{S1}, \eqref{S2}, \eqref{S4} with $p=p_\star$, together with
\[
  12e^2p_\star\varepsilon^2T_0\le\frac1{2e^2},
  \qquad
  d\gamma^2e^{-T_0\eta\rho_k/2}\le\frac1{2e^2} .
\]
Then
\[
  \|W_{T_0}\mathbf 1_{T_0}\|_{F,p_0}\le e^{-1}
  \qquad\text{and}\qquad
  \max_{1\le j\le T_0}\max_{F\in\mathcal E}\|V^\top FUW_j\mathbf 1_{j-1}\|_{F,p_1}
   \le\frac\gamma e .
\]
\end{proposition}

\begin{proof}
\emph{Admissible exponents.} Definition \ref{def:Fp} defines $\|\cdot\|_{F,p}$ only for
$p\ge2$, and Lemma \ref{lem:normprops} and Theorems \ref{thm:F10}--\ref{thm:F11} are stated under
that restriction, so both exponents are truncated at $2$; this matches the convention
$p_\delta=\max\{2,\log(2/\delta)\}$ used in Phase II. The truncation is active only for $p_0$, and
only when $\delta>6e^{-2}$: since $12T_0(m+1)\ge24$ we have $p_1=\log\frac{12T_0(m+1)}\delta>\log
24>2$ always, whence also $p_\star=p_1$. The truncation costs nothing downstream, because the only
property of $p_0$ used outside this proof is $e^{-p_0}\le\delta/6$ (Theorem \ref{thm:F12}), which
holds with equality when $\log\frac6\delta\ge2$ and, when $\log\frac6\delta<2$, holds because
$\delta>6e^{-2}$ gives $e^{-p_0}=e^{-2}<\delta/6$.

\emph{Terminal bound.} The Phase I good events supply the hypothesis
$\|V^\top(C_i-\Sigma)UW_{i-1}\mathbf 1_{i-1}\|_2\le M\gamma$ of Theorem \ref{thm:F10}, and
\eqref{S3} is vacuous for a constant step size, so \eqref{eq:F7} at $p=p_0$ gives
\[
  \|W_{T_0}\mathbf 1_{T_0}\|_{F,p_0}^2\le e^{-T_0\eta\rho_k/2}\|W_0\mathbf 1_0\|_{F,p_0}^2
   +10p_0\varepsilon^2T_0 .
\]
By the definition of $G_0$, $\|W_0\mathbf 1_0\|_F^2\le d\gamma^2$ pointwise, hence
$\|W_0\mathbf 1_0\|_{F,p_0}^2\le d\gamma^2$. The second displayed condition bounds the first term
by $\frac1{2e^2}$, and $p_0\le p_\star$ with the first condition bounds the second by
$10p_0\varepsilon^2T_0\le12e^2p_\star\varepsilon^2T_0\le\frac1{2e^2}$. Hence
$\|W_{T_0}\mathbf 1_{T_0}\|_{F,p_0}^2\le e^{-2}$.

\emph{Good-event moment bound.} For $F\in\mathcal E$ one has $V^\top FU\in\mathcal E_{1,1}$, and
the definition of $G_0$ gives $\max_{E\in\mathcal E_{1,1}}\|EW_0\mathbf 1_0\|_{F,p_1}^2\le
\gamma_{\rm init}^2=\gamma^2/2e^2$. Theorem \ref{thm:F11} with $r=\ell=1$ and $p=p_1$ therefore
gives $\|V^\top FUW_j\mathbf 1_{j-1}\|_{F,p_1}^2\le\frac{\gamma^2}{2e^2}e^{-j\eta\rho_k/2}
+12e^2p_1\gamma^2\varepsilon^2j\le\frac{\gamma^2}{2e^2}+\frac{\gamma^2}{2e^2}$, using
$e^{-j\eta\rho_k/2}\le1$, $p_1\le p_\star$, $j\le T_0$ and the first condition. Theorem
\ref{thm:F11} is applicable at $p=p_1$ because $p_1\ge\log\frac{12T_0(m+1)}\delta\ge\log m$, which
is the hypothesis $p\ge\log m$ of Propositions \ref{prop:F1}--\ref{prop:F2}; this is the only
exponent at which the product-class recursion is used.
\end{proof}

\begin{theorem}[Finite-support Phase I]\label{thm:F12}
Let the updates satisfy the hypotheses of Proposition \ref{prop:F3}, with deterministic finite
support of cardinality $m$ and constant step size $\eta$, and let
$Q_0=\mathrm{QR}(Z_0)$ for $Z_0$ with
i.i.d.\ $\mathcal N(0,1)$ entries and $\sigma(Z_0)\subseteq\mathcal F_0$. Let $\delta\in(0,1)$,
$C_\gamma$ be a sufficiently large absolute constant, and set
\[
  \gamma:=\max\Big\{e,\ C_\gamma\min\Big\{\frac{\sqrt{k\log(emT_0/\delta)}}\delta,\
   \frac d{\delta^2}\Big\}\Big\},\quad
  L_\gamma:=\max\{1,\log(2e^2d\gamma^2)\},\quad
  \eta:=\frac{C_\eta L_\gamma}{\rho_kT_0},
\]
with $C_\eta\ge2$. If
\[
  T_0\ge C\max\Big\{\frac{L_\gamma M(1+\gamma)}{\rho_k},\ \frac{L_\gamma\|\Sigma\|_2}{\rho_k},\
   \frac{p_\star L_\gamma M^2(1+\gamma)^2}{\rho_k^2},\
   \frac{p_\star L_\gamma^2M^2(1+\gamma)^2}{\rho_k^2}\Big\},
\]
then $\mathbb P(\|W_{T_0}\|_F>1)\le\delta/3$.
\end{theorem}

\begin{proof}
The first three lower bounds on $T_0$ give \eqref{S1}, \eqref{S2} and \eqref{S4} with
$p=p_\star$, and the fourth gives $12e^2p_\star\varepsilon^2T_0\le\frac1{2e^2}$. The computation of
Theorem D.5 of Huang et al.\ gives the same bound without the factor $2e^2$; the two factors are
independent, $e^2$ coming from the random class index of \eqref{eq:randidx} and $2$ from the
constant in Lemma \ref{lem:normprops}(vi), and both are absorbed by the free constant in the lower
bound on $T_0$. For the contraction condition,
$T_0\eta\rho_k=C_\eta L_\gamma$ and $C_\eta\ge2$ give $e^{-T_0\eta\rho_k/2}\le e^{-L_\gamma}$, and
the definition of $L_\gamma$ then gives $d\gamma^2e^{-T_0\eta\rho_k/2}\le\frac1{2e^2}$, the second
condition of Proposition \ref{prop:F3}. The corresponding definition in Huang et al.\ is
$L_\gamma=\max\{1,\log(2e^2d\gamma^2/k^{2/p_0})\}$, and the fourth lower bound on $T_0$ carries a
denominator $k^{2/p_0}$; since $k^{2/p_0}\in[1,e^2]$ whenever $p_0\ge\log k$, the two differ by at
most $2$ additively and by a factor at most $e^2$ respectively, so $T_0$ is unchanged in order.

By Lemma \ref{lem:welldef}, on $G_{T_0}$ the matrix $V^\top Z_{T_0}$ is invertible and
$W_{T_0}=U^\top Q_{T_0}(V^\top Q_{T_0})^{-1}$, so
\[
  \mathbb P\big(\|W_{T_0}\|_F>1\big)\le\mathbb P\big(\|W_{T_0}\mathbf 1_{T_0}\|_F>1\big)
   +\mathbb P(G_{T_0}^c),
\]
and by Proposition \ref{prop:F3} with Lemma \ref{lem:normprops}(v) at $s=1$,
\[
  \mathbb P\big(\|W_{T_0}\mathbf 1_{T_0}\|_F>1\big)\le
   \|W_{T_0}\mathbf 1_{T_0}\|_{F,p_0}^{p_0}\le e^{-p_0}\le\frac\delta6 .
\]
The corresponding step of Huang et al.\ produces $k\,e^{-p_0}$ and so requires
$p_0=\log(6k/\delta)$; the factor $k$ there is the $k^{1/p_0}$ of their terminal bound, which
Proposition \ref{prop:F3} does not carry. For $\mathbb P(G_{T_0}^c)$, the event $G_0$ concerns $W_0$ alone, so a union bound over the finitely many elements of the classes
$\mathcal E_{r,\ell}$, combined with Gaussian bounds on $\|U^\top Z_0\|_2$ and
$\|(V^\top Z_0)^{-1}\|_F$, gives
\begin{equation}\label{eq:G0init}
  \mathbb P(G_0^c)\le\frac\delta{12}
\end{equation}
whenever $\gamma\ge C_\gamma\min\{\sqrt{k\log(emT_0/\delta)}/\delta,\ d/\delta^2\}$, which is the
choice made in the statement; this is Lemma H.4 of \citet{huang2021streaming}.
and for $j\ge1$, Markov's inequality in $\|\cdot\|_2\le\|\cdot\|_F$ together with the second bound
of Proposition \ref{prop:F3} gives $\mathbb P(G_j^c\cap G_{j-1})\le\sum_{F\in\mathcal E}
\gamma^{-p_1}\|V^\top FUW_j\mathbf 1_{j-1}\|_{F,p_1}^{p_1}\le m\,e^{-p_1}$, so
$\sum_{j\le T_0}\mathbb P(G_j^c\cap G_{j-1})\le mT_0e^{-p_1}\le\delta/12$ by the choice of $p_1$.
Hence $\mathbb P(G_{T_0}^c)\le\delta/6$ and the two terms sum to $\delta/3$.
\end{proof}

\subsection{Phase II}\label{sec:phase2}

For Phase II the good events are taken in the Frobenius norm. Fix a starting time $\tau$ and
relabel it as the index $0$, so that $W_0=W_\tau$; fix a threshold $\gamma\ge1$ and set
\[
  G_0:=\{\|W_0\|_F\le1\},\qquad
  G_m:=G_{m-1}\cap\{\|W_m\|_F\le\gamma\},\qquad
  \mathbf 1_m:=\mathbf 1_{G_m}.
\]
These events are decreasing, and since $\|W_s\mathbf 1_s\|_2\le\|W_s\mathbf 1_s\|_F\le\gamma$ they
imply the hypothesis $\|V^\top(C_m-\Sigma)UW_{m-1}\mathbf 1_{m-1}\|_2\le M\gamma$ of Lemmas
\ref{lem:steps} and \ref{lem:welldef}. No product class and no finite-support hypothesis is used
in this subsection.

Phase II is Theorem \ref{thm:F10} with a decaying step size, as in Section 4 of Huang et al. The
one point to notice is that the hypotheses \eqref{S1}--\eqref{S3} of its first conclusion
\eqref{eq:F5} do not involve $p$, while the conclusion holds for every $p\ge2$. The good-event
check at step $j$ may therefore use its own exponent $p_j$. Choosing $p_j$ to grow linearly in
$j$, as large as the fluctuation term permits, makes the per-step failure probabilities decay
geometrically, so the union bound over the run costs $O(1)$ rather than a factor $T-\tau$. This is
the device of Theorem 4.2 of Huang et al.; in the Frobenius norm it is better still, because their
Markov step pays $k\,e^{-p}$ where ours pays $e^{-p}$.

\begin{proposition}[Phase II with decaying step size]\label{prop:phase2}
Let $\|W_0\|_F\le1$. Fix $c_\eta\ge8$, put $\alpha:=\tfrac12c_\eta$ and $\gamma:=2e$, and for
$m\ge1$ set
\[
  \eta_m=\frac{c_\eta}{\rho_k(\beta+m)},\qquad
  \beta\ge\max\Big\{1,\ \frac{4c_\eta\lambda_1}{\rho_k},\ \frac{27c_\eta M}{\rho_k},\
    \frac{C_3c_\eta^2M^2}{\rho_k^2}\Lambda_\delta\Big\},
\]
\[
  \Lambda_\delta:=\log\Big(e+\frac{c_\eta^2M^2}{\rho_k^2\delta}\Big)+\log\frac2\delta,
  \qquad p_\delta:=\max\{2,\log(2/\delta)\}.
\]
Then with probability at least $1-\delta$,
\[
  \|W_T\|_F\le e\Big(\frac{\beta+1}{\beta+T}\Big)^{\alpha}
   +C_4\,\frac{c_\eta M}{\rho_k}\sqrt{\frac{p_\delta}{\beta+T}} .
\]
\end{proposition}

\begin{remark}
    Neither $\beta$ nor the bound depends on $T$, $d$ or $k$. The constant $c_\eta$ is free
subject to $c_\eta\ge8$, and may be chosen as a function of the sample size, as
\Cref{thm:utility-priv-ojas-batches,thm:utility-priv-ojas-adaptive} do; $\beta$ then depends on it
through the displayed branches, but still not on $T$. Every constant in the proof is independent
of $c_\eta$: in particular $C_5=2(1+\gamma)\sqrt{C_1\gamma^2+C_2}$ with $\gamma=2e$, and
$C_4=eC_5$.
\end{remark}

\begin{proof}
\emph{Hypotheses of Theorem \ref{thm:F10}.} The good events are decreasing and supply
$\|V^\top(C_m-\Sigma)UW_{m-1}\mathbf 1_{m-1}\|_2\le M\gamma$. Since $\eta_m$ decreases it suffices
to check \eqref{S1}--\eqref{S2} at $m=1$. For \eqref{S2}, $\eta_1\lambda_1=c_\eta\lambda_1/
(\rho_k(\beta+1))\le\tfrac14$ by the second bound on $\beta$. For \eqref{S1}, $\varepsilon_1=
2c_\eta M(1+2e)/(\rho_k(\beta+1))\le\tfrac12$ by the third, since $4(1+2e)<27$. For \eqref{S3},
write $x:=1/(\beta+m)\le\tfrac12$; then $\varepsilon_m/\varepsilon_{m-1}=\eta_m/\eta_{m-1}=1-x$,
while $e^{-\eta_m\rho_k/4}=e^{-c_\eta x/4}\le e^{-2x}\le1-x$, the middle inequality because
$c_\eta\ge8$ and the last because $x\le\tfrac12$. Condition \eqref{S4} is never used.

\emph{Moment bound.} Fix $p\ge2$ and write $S_m:=\beta+m$. We have $\|W_0\mathbf 1_0\|_{F,p}\le1$
and $\|W_i\mathbf 1_i\|_{F,p}\le\gamma$ pointwise, so, with $\gamma\ge1$, the sum in \eqref{eq:F5}
is at most $m\gamma^2$ and
\[
  \|W_m\mathbf 1_\ast\|_{F,p}^2\le e^{-s_m\rho_k}+(C_1\gamma^2+C_2)\,p\,\varepsilon_m^2m,
  \qquad\mathbf 1_\ast\in\{\mathbf 1_m,\mathbf 1_{m-1}\}.
\]
For the first term, $\rho_ks_m=c_\eta\sum_{i=1}^m(\beta+i)^{-1}\ge c_\eta\log\frac{\beta+m}
{\beta+1}$, so $e^{-s_m\rho_k}\le((\beta+1)/S_m)^{c_\eta}$. For the second, $\varepsilon_m^2m=
4c_\eta^2M^2(1+\gamma)^2m/(\rho_k^2S_m^2)\le4c_\eta^2M^2(1+\gamma)^2/(\rho_k^2S_m)$ because
$m\le S_m$. Hence, by $\sqrt{u+v}\le\sqrt u+\sqrt v$,
\begin{equation}\label{eq:theta}
  \|W_m\mathbf 1_\ast\|_{F,p}\le\Theta_m(p):=\Big(\frac{\beta+1}{S_m}\Big)^{\alpha}
   +C_5\,\frac{c_\eta M}{\rho_k}\sqrt{\frac p{S_m}},\qquad
  C_5:=2(1+\gamma)\sqrt{C_1\gamma^2+C_2},
\end{equation}
with $\alpha=\tfrac12c_\eta$ as stated.

\emph{Good event.} Put $\nu:=\rho_k^2/(C_5^2c_\eta^2M^2)$ and $p_j:=\nu S_j$. The fourth bound on
$\beta$ gives $\nu\beta\ge2$, as $\Lambda_\delta\ge1$ and $C_3\ge2C_5^2$, so $p_j\ge2$ and
\eqref{eq:theta} applies at $p=p_j$. By construction the second term of $\Theta_j(p_j)$ equals
$1$, and the first is at most $1$; hence $\Theta_j(p_j)\le2=\gamma/e$, and Markov's inequality
gives
\[
  \mathbb P\big(\|W_j\|_F>\gamma,\ G_{j-1}\big)\le\gamma^{-p_j}
   \|W_j\mathbf 1_{j-1}\|_{F,p_j}^{p_j}\le e^{-p_j}=e^{-\nu(\beta+j)} .
\]
Summing and using $e^\nu-1\ge\nu$,
\[
  \mathbb P(G_T^c)\le\sum_{j\ge1}e^{-\nu(\beta+j)}=\frac{e^{-\nu\beta}}{e^\nu-1}
   \le\frac{e^{-\nu\beta}}\nu\le\frac\delta2 ,
\]
the last step because $\nu\beta\ge\log\frac2{\nu\delta}$, which is the fourth bound on $\beta$
once $1/\nu=C_5^2c_\eta^2M^2/\rho_k^2$ is substituted and $C_3$ is taken large enough in terms of
$C_5$. The number of steps never enters.

\emph{Conclusion.} Apply \eqref{eq:theta} once more at $p=p_\delta$: Markov's inequality gives
$\mathbb P\big(\|W_T\|_F>e\Theta_T(p_\delta),\,G_T\big)\le e^{-p_\delta}\le\delta/2$, and
$e\Theta_T(p_\delta)$ is the asserted bound with $C_4:=eC_5$.
\end{proof}

No constant in this subsection depends on $k$, $d$ or $k^{2/p}$.

\subsection{From the error matrix to the subspace error}

\begin{theorem}[Phase II in Frobenius norm]\label{thm:phase2}
Let $T_0<T$ and suppose Phase I delivers $\|W_{T_0}\|_F\le1$ with probability at least
$1-\delta_I$. Run Proposition \ref{prop:phase2} on $(T_0,T]$ with $\tau=T_0$, failure probability
$\delta_{II}$ and the $\beta$ defined there, and set $S:=\beta+T-T_0$ and
$\alpha:=\tfrac12c_\eta\ge4$. Then for every $T>T_0$, with probability at least
$1-\delta_I-\delta_{II}$,
\[
  \big\|Q_TQ_T^\top-VV^\top\big\|_F
  \;\le\;\sqrt2\,e\Big(\frac{\beta+1}S\Big)^{\alpha}
   \;+\;C_8\,\frac M{\rho_k}\sqrt{\frac{\Lambda_{\delta_{II}}}S} .
\]
No burn-in stage is required between the two phases, and no lower bound on $T-T_0$ is needed.
\end{theorem}

\begin{proof}
On the Phase I event $\|W_{T_0}\|_F\le1$, the entry hypothesis of Proposition \ref{prop:phase2}
holds. By that proposition, with probability $1-\delta_{II}$,
$\|W_T\|_F\le e(\frac{\beta+1}S)^{\alpha}+C_4\frac{c_\eta M}{\rho_k}\sqrt{p_{\delta_{II}}/S}$.
For the second term, $\Lambda_\delta\ge1$ and $\Lambda_\delta\ge\log\frac2\delta$ give
$p_\delta=\max\{2,\log\frac2\delta\}\le2\Lambda_\delta$. Lemma \ref{lem:tansin} converts,
multiplying both terms by $\sqrt2$, and the two failure probabilities add; take
$C_8:=2C_4c_\eta$.
\end{proof}

\subsection{The main theorem}

\begin{theorem}[Main theorem, Frobenius form; replaces \Cref{thm:main-thm-streaming}]\label{thm:main}
Let $C_1,\dots,C_n$ satisfy Assumption C with boundedness constant $M$ and $\rho_k>0$, and let
$C_1,\dots,C_{n_0}$ be in addition i.i.d. and independent of $Z_0$, as in~\Cref{app:G}. Let $Z_0$
have i.i.d.\ $\mathcal N(0,1)$ entries with $\sigma(Z_0)\subseteq\mathcal F_0$, and let $Z_0$ be
independent of the auxiliary array of \Cref{prop:G4}; set
$Q_0:=\mathrm{QR}(Z_0)$. Let $V_k$ contain the $k$ leading eigenvectors of $\Sigma$. Fix
$\delta\in(0,1)$ and put $\delta_I:=\delta/2$, $\xi:=\delta/6$, $\delta_{II}:=2\delta/3$. Let
$n_0$ be the Phase I length, let $L$ and $m_\star=n_0(L+1)+1$ be as in \Cref{prop:G4} at parameters $(\delta_I,\xi)$, and let $\gamma,L_\gamma,\eta$ and the lower bound
on $n_0$ be as in Theorem \ref{thm:F12} at failure probability $\delta_I$ and support size
$m_\star$, so that
$n_0=\widetilde\Theta\big(kM^2/\delta^2\rho_k^2\big)$; and let $\beta,\Lambda:=
\Lambda_{\delta_{II}}$ and $c_\eta$ be as in Proposition \ref{prop:phase2} at failure probability
$\delta_{II}$. Take
\[
  \eta_t=\begin{cases}
    \dfrac{C_\eta L_\gamma}{\rho_k n_0}, & t\le n_0,\\[2.2ex]
    \dfrac{c_\eta}{\rho_k(\beta+t-n_0)}, & t>n_0 .
  \end{cases}
\]
Then for every $n>n_0$, the output $Q_n$ satisfies, with probability at least $1-\delta$,
\begin{equation}\label{eq:maintwoterm}
  \boxed{\;\big\|Q_nQ_n^\top-V_kV_k^\top\big\|_F\;\le\;
   \sqrt2\,e\Big(\frac{\beta+1}{\,\beta+n-n_0\,}\Big)^{\alpha}
   \;+\;C_8\,\frac M{\rho_k}\sqrt{\frac{\Lambda}{\,n-n_0\,}}\;}
\end{equation}
where $\alpha=\tfrac12c_\eta\ge4$ and
$\Lambda\asymp\log\frac1\delta+\log\big(e+\frac{M^2}{\rho_k^2\delta}\big)$.
\end{theorem}

\begin{remark}[on the two terms]\label{rem:twoterms}
The second term is the rate; the first is the cost of forgetting the Phase I output and decays
like $(n-n_0)^{-\alpha}$ with $\alpha=c_\eta/2\ge4$, against $(n-n_0)^{-1/2}$ for the second. It
falls below the second once
\[
  n-n_0\;\gtrsim\;\beta\,\Gamma^{1/(2\alpha-1)},
  \qquad
  \Gamma:=\frac{\beta\rho_k^2}{C_3c_\eta^2M^2\Lambda}\;\ge\;1,
\]
where $\Gamma$ measures how far $\beta$ exceeds its last branch; $\Gamma=O(1)$ exactly when that
branch attains the maximum defining $\beta$, and $2\alpha-1\ge7$, so the extra factor is at worst
$\Gamma^{1/7}$. Both displays treat $c_\eta$ as an absolute constant. \Cref{prop:phase2} requires only
$c_\eta\ge8$, so $c_\eta$ may instead be chosen as a function of the horizon, and this is what
\Cref{thm:utility-priv-ojas-batches,thm:utility-priv-ojas-adaptive} do. The first term of
\eqref{eq:maintwoterm} is at most $\epsilon_\star$ as soon as
$\frac{\beta+n-n_0}{\beta+1}\ge(\sqrt2e/\epsilon_\star)^{1/\alpha}$. Take
$\epsilon_\star=n^{-1/2}$ and $c_\eta\ge\max\{8,\log n\}$: the right-hand side is
$(\sqrt2e)^{2/c_\eta}n^{1/c_\eta}\le(\sqrt2e)^{1/4}\,e<3.81$, since $n^{1/c_\eta}\le n^{1/\log n}
=e$, while $n-n_0\ge7\beta$ together with $\beta\ge1$ gives
$\frac{\beta+n-n_0}{\beta+1}\ge\frac{8\beta}{2\beta}=4$. Hence
\[
  c_\eta\ge\max\{8,\log n\}\ \text{ and }\ n-n_0\ge7\beta
  \quad\Longrightarrow\quad
  \sqrt2\,e\Big(\frac{\beta+1}{\,\beta+n-n_0\,}\Big)^{\alpha}\;\le\;\frac1{\sqrt n},
\]
so the crossover is a fixed multiple of $\beta$ and no $\Gamma$ factor survives. This is what lets
the batch count of \Cref{thm:utility-priv-ojas-batches,thm:utility-priv-ojas-adaptive} be
$\Theta(\varkappa\log n)$ with $\varkappa:=\lambda_1/\rho_k$, since the branch
$4c_\eta\lambda_1/\rho_k$ of $\beta$ then reads $4\varkappa\lceil\log n\rceil$. At constant
$c_\eta$ the first display would instead force $n-n_0\gtrsim\beta\,n^{1/(2\alpha)}$, polynomial in
$n$.
\end{remark}

\begin{remark}[on the hypotheses and on $\widetilde\Theta$]\label{rem:mainhyp}
Phase II uses nothing beyond Assumption C, and its only distributional input is the conditional
mean of C.1, through \Cref{thm:F10}; the almost-sure bound of C.2 enters only pointwise, through \Cref{lem:Ublock}. Independence of the updates is required by Phase I alone, and only on the first $n_0$ updates,
where it is what makes the atom array of \Cref{prop:G4} drawable in advance. It is the hypothesis
of Huang et al., and on the batches where it is needed it holds because
\Cref{algo:priv-ojas-batches,algo:priv-ojas-adaptive} run their first $n_0$ batches with a noise
multiplier fixed before the run.

The notation $\widetilde\Theta$ suppresses factors polylogarithmic in $d$, $k$, $n_0$, $1/\delta$
and $1/\xi$. The support size is $m_\star=n_0(L+1)+1$, and since
$L=\widetilde O(kn_0^2/\xi^2)$ this gives $\log m_\star=O\big(\log\frac{kdn_0}\xi\big)$. That $m_\star$ depends on $n_0$ while the lower
bound on $n_0$ depends on $m_\star$, through $p_\star$ and $\gamma$, is a fixed point rather than a
circularity: the dependence is logarithmic, so the lower bound is of the form
$n_0\ge A\log^b(Cn_0)$ with $A,b,C$ free of $n_0$, which is solved by any
$n_0\ge C'A\log^b(CA)$ and leaves the order $\widetilde\Theta\big(kM^2/(\delta^2\rho_k^2)\big)$
unchanged.
\end{remark}

\begin{proof}
\Cref{prop:G4} gives $\|W_{n_0}\|_F\le1$ with probability at least $1-\delta_I/3-\xi
=1-\delta/3$. On that event Theorem \ref{thm:phase2} applies with $T_0=n_0$, Phase I failure
probability $\delta/3$ and Phase II failure probability $\delta_{II}=2\delta/3$, and its
conclusion is \eqref{eq:maintwoterm}, using $\beta+n-n_0\ge n-n_0$ in the second term. The two
failure probabilities sum to $\delta$.
\end{proof}

\begin{corollary}[Operator-norm form]\label{cor:op}
Under the hypotheses of Theorem \ref{thm:main}, with probability at least $1-\delta$,
\[
  \big\|Q_nQ_n^\top-V_kV_k^\top\big\|_2\;\le\;
   \sqrt2\,e\Big(\frac{\beta+1}{\,\beta+n-n_0\,}\Big)^{\alpha}
   \;+\;C_8\,\frac M{\rho_k}\sqrt{\frac{\Lambda}{\,n-n_0\,}} .
\]
\end{corollary}

\begin{proof}
$\|X\|_2\le\|X\|_F$ for every matrix $X$.
\end{proof}

\begin{remark}
\Cref{cor:op} shows that \Cref{thm:main} implies \Cref{thm:main-thm-streaming} at no cost, since
$\|\cdot\|_2\le\|\cdot\|_F$ pointwise, and with $\log(Mk/\rho_k\delta)$ improved to $\Lambda$. The
converse implication is the expensive one, and is what this appendix exists to avoid.
\end{remark}

\section{Lifting from finite support to i.i.d.\ bounded updates}\label{app:G}

Phase I of~\Cref{algo:priv-ojas-batches} and of the warm-up branch of
\Cref{algo:priv-ojas-adaptive} is run at a fixed clipping radius and a fixed noise multiplier.
On the no-clipping event the update is therefore
\[
  C_t=\frac1B\sum_{i=B(t-1)+1}^{tB}A_i+G_t,
  \qquad G_t\sim\mathrm{GOE}_d(\sigma^2),
\]
which does not depend on $Q_{t-1}$: the Phase I updates $C_1,\dots,C_{T_0}$ are i.i.d., independent
of the initialisation $Z_0$, have mean $\Sigma$, and satisfy $\|C_t-\Sigma\|_{(2,k)}\le M$ almost
surely. Write $K$ for their common law. They are not finitely supported, so
\Cref{thm:F12} does not apply to them directly; this section supplies the replacement of $K$ by a
finitely supported law with the same mean and the same $(2,k)$-bound. Phase II needs nothing from this section, since it uses only \Cref{ass:C}.

\subsection{The adaptive updates and the pre-update filtration}\label{sec:Gverify}

The warm-up updates above are i.i.d., and nothing more is needed from them. The adaptive updates of
\Cref{algo:priv-ojas-adaptive,alg:tangent-adaptive} are not, and they enter Phase II, which uses
only~\Cref{ass:C}. The next lemma records that they satisfy it, with respect to a filtration that
places the step-$t$ multiplier randomness in the conditioning.

\begin{lemma}[The private updates satisfy~\Cref{ass:C}]\label{lem:Gverify}
The blocks $I_t$ are disjoint and deterministic with $|I_t|=B/2$,
the matrices $A_1,\dots,A_n$ are independent and symmetric with $\mathbb E[A_i]=\Sigma$ and
$\|A_i-\Sigma\|_{(2,k)}\le M_A$ almost surely,
\[
  C_t=\frac2B\sum_{i\in I_t}A_i+G_t,
  \qquad G_t\mid\mathcal H_t\sim\mathrm{GOE}_d(\sigma_t^2)\ \text{conditioned on }
  \|G_t\|_{(2,k)}\le\mathsf N_B ,
\]
with $\sigma_t$ measurable with respect to $\mathcal H_t=\sigma(\mathcal F_{t-1},\mathcal R_t)$ and
the block $I_t$ independent of $\mathcal H_t$, where $\mathcal R_t$ is the randomness the algorithm
draws at step $t$ to form $\sigma_t$ and $\mathcal F_t:=\sigma(\mathcal F_{t-1},\mathcal R_t,C_t)$.
Define the \emph{pre-update filtration}
\begin{equation}\label{eq:preupdate}
  \widetilde{\mathcal F}_{t-1}:=\mathcal H_t=\sigma(\mathcal F_{t-1},\mathcal R_t),\qquad t\ge1,
  \qquad \widetilde{\mathcal F}_t=\sigma(\mathcal F_t,\mathcal R_{t+1}) .
\end{equation}
Then~\Cref{ass:C} holds with respect to $(\widetilde{\mathcal F}_t)_{t\ge0}$, with
$M=M_A+\mathsf N_B$.
\end{lemma}

\begin{proof}
Write $\mu_t$ for the law of $\frac2B\sum_{i\in I_t}A_i$, $P_\sigma$ for the law of
$\mathrm{GOE}_d(\sigma^2)$ conditioned on $\{\|G\|_{(2,k)}\le\mathsf N_B\}$, and set
$K_t^{(\sigma)}:=\mu_t*P_\sigma$.

\emph{$(\widetilde{\mathcal F}_t)$ is a filtration, and the process is adapted to it.} Since
$\mathcal F_t=\sigma(\mathcal F_{t-1},\mathcal R_t,C_t)$ we have $\mathcal F_{t-1}\subseteq
\mathcal F_t$ and $\mathcal R_t\subseteq\mathcal F_t$, hence
$\widetilde{\mathcal F}_{t-1}=\sigma(\mathcal F_{t-1},\mathcal R_t)\subseteq\mathcal F_t\subseteq
\sigma(\mathcal F_t,\mathcal R_{t+1})=\widetilde{\mathcal F}_t$, so
$(\widetilde{\mathcal F}_t)_{t\ge0}$ is increasing.
Moreover $C_t$ is $\mathcal F_t$-measurable and hence $\widetilde{\mathcal F}_t$-measurable, and $Q_t$, being a
function of $Q_0$ and $C_1,\dots,C_t$, is $\widetilde{\mathcal F}_t$-measurable as well. Finally
$Z_0\in\mathcal F_0\subseteq\widetilde{\mathcal F}_0$, so the requirement $\sigma(Z_0)\subseteq\mathcal F_0$ of Assumption \ref{ass:C} is met
for $(\widetilde{\mathcal F}_t)$.

\emph{The conditional law.} The blocks are disjoint and the $A_i$ independent, so $\frac2B\sum_{i\in I_t}A_i$ is
independent of $\mathcal H_t=\widetilde{\mathcal F}_{t-1}$; and $G_t$ is conditionally $P_{\sigma_t}$ given
$\widetilde{\mathcal F}_{t-1}$ with $\sigma_t$ being $\widetilde{\mathcal F}_{t-1}$-measurable. Hence
\[
  \mathcal L\big(C_t\mid\widetilde{\mathcal F}_{t-1}\big)=\mu_t*P_{\sigma_t}=K_t^{(\sigma_t)},
  \qquad \sigma_t\in\widetilde{\mathcal F}_{t-1},
\]
so $(\sigma_t)$ is predictable for $(\widetilde{\mathcal F}_t)$. The indexing matters: $\sigma_t$ is computed from
$\mathcal R_t$ and so is \emph{not} $\mathcal F_{t-1}$-measurable, and conditioning on
$\mathcal F_{t-1}$ alone would return the mixture $\mathbb E[K_t^{(\sigma_t)}\mid
\mathcal F_{t-1}]$; shifting
to the pre-update filtration \eqref{eq:preupdate} places the fresh randomness of step $t$ in the
conditioning and recovers the single kernel $K_t^{(\sigma_t)}$. The filtration $(\widetilde{\mathcal F}_t)$ is strictly
larger than the one generated by $C_1,\dots,C_t$; Assumption C and the whole of Sections
\ref{sec:phase1}--\ref{sec:phase2} are to be read with respect to it. This is legitimate because
the computation below gives $\mathbb E[C_t\mid\widetilde{\mathcal F}_{t-1}]=\Sigma$ directly, which is stronger than
Assumption C.1 for the smaller filtration, while every quantity the recursions require to be
predictable --- $W_{t-1}$, $\mathbf 1_{t-1}$ and the good events $G_{t-1}$, all functions of $Z_0$
and $C_1,\dots,C_{t-1}$ --- is $\widetilde{\mathcal F}_{t-1}$-measurable.

\emph{Mean and bound.} Since $|I_t|=B/2$ and $\mathbb E[A_i]=\Sigma$ for every $i$,
$\int C\,\mu_t(dC)=\frac2B\cdot\frac B2\,\Sigma=\Sigma$, for every block. The truncated Gaussian is
centred: $\mathrm{GOE}_d(\sigma^2)$ is symmetric under $G\mapsto-G$ and the truncation event is
invariant under that map, so $P_\sigma$ has mean $0$. Hence $\int C\,K_t^{(\sigma)}(dC)=\Sigma$ for
every $t$ and every $\sigma$. For the bound, almost surely
\[
  \Big\|\frac2B\sum_{i\in I_t}A_i-\Sigma\Big\|_{(2,k)}
  =\Big\|\frac2B\sum_{i\in I_t}(A_i-\Sigma)\Big\|_{(2,k)}
  \le\frac2B\sum_{i\in I_t}\|A_i-\Sigma\|_{(2,k)}\le\frac2B\cdot\frac B2\,M_A=M_A ,
\]
and $\|G\|_{(2,k)}\le\mathsf N_B$ holds $P_\sigma$-almost surely by construction, so
$\|C-\Sigma\|_{(2,k)}\le M_A+\mathsf N_B=M$ holds $K_t^{(\sigma)}$-almost surely, again for every
$t$ and every $\sigma$. Neither computation sees $\sigma$ or $t$.
\end{proof}

\subsection{The reduction}

The reduction uses four standard ingredients, which we record first: a matrix Bernstein
inequality, its consequence for the empirical mean of a $(2,k)$-bounded law, the mean-preserving
replacement that turns such a law into a finitely supported one, and a sequential coupling bound.

\begin{lemma}[Matrix Bernstein input]\label{lem:bern}
Let $X_1,\dots,X_L\in\mathbb R^{d\times d}$ be independent, mean-zero, self-adjoint matrices with
$\|X_\ell\|_2\le R$ almost surely and $\big\|\sum_{\ell=1}^L\mathbb E[X_\ell^2]\big\|_2\le
\sigma^2$. Then for every $s>0$,
\[
  \mathbb P\Big(\Big\|\sum_{\ell=1}^LX_\ell\Big\|_2\ge s\Big)
   \le 2d\exp\Big(-\frac{s^2/2}{\sigma^2+Rs/3}\Big) .
\]
\end{lemma}

\begin{proof}
Theorem 6.1.1 of \citet{tropp2015introduction} gives
$\mathbb P\big(\lambda_{\max}(\sum_\ell X_\ell)\ge s\big)\le d\exp\big(-\frac{s^2/2}
{\sigma^2+Rs/3}\big)$; applying it to $\sum_\ell X_\ell$ and to $-\sum_\ell X_\ell$, whose
summands satisfy the same hypotheses, and taking a union bound gives the factor $2d$.
\end{proof}

\begin{lemma}[Conditional empirical-mean bound]\label{lem:empmean}
Let $K$ be a probability measure on symmetric $d\times d$ matrices with $\int C\,K(dC)=\Sigma$ and
$\|C-\Sigma\|_{(2,k)}\le M$ for $K$-almost every $C$. Let $\widehat C_1,\dots,\widehat C_L$ be
i.i.d.\ $K$ and put $\widehat\Sigma_L:=\frac1L\sum_{\ell\le L}\widehat C_\ell$. Then for every
$u>0$,
\[
  \mathbb P\big(\|\widehat\Sigma_L-\Sigma\|_2\ge u\big)
   \le 2d\exp\Big(-\frac{Lu^2/2}{M^2+Mu/3}\Big) .
\]
Consequently, if $0<\alpha\le\frac12$, $0<\beta<1$ and
\[
  L\ \ge\ C_0\,\frac k{\alpha^2}\,\log\frac{2d}\beta
\]
for a sufficiently large universal constant $C_0$, then
\[
  \mathbb P\Big(\|\widehat\Sigma_L-\Sigma\|_2>\frac{\alpha M}{(1-\alpha)\sqrt k}\Big)\le\beta .
\]
\end{lemma}

\begin{proof}
Set $X_\ell:=\widehat C_\ell-\Sigma$; these are independent, mean-zero and self-adjoint, and
$\|X_\ell\|_2\le\|X_\ell\|_{(2,k)}\le M$. Hence $X_\ell^2\preceq M^2I_d$ and
$\big\|\sum_{\ell\le L}\mathbb E[X_\ell^2]\big\|_2\le LM^2$. Lemma \ref{lem:bern} applied with
$R=M$, $\sigma^2=LM^2$ and $s=Lu$ gives the first claim. Substituting
$u=\alpha M/((1-\alpha)\sqrt k)$ and using $\alpha\le\frac12$ gives the second, after increasing
$C_0$.
\end{proof}

\begin{lemma}[Mean-preserving finite-support replacement]\label{lem:meanpres}
Let $\widehat C_1,\dots,\widehat C_L$ be symmetric matrices with
$\|\widehat C_\ell-\Sigma\|_{(2,k)}\le M$ for every $\ell\le L$, and put
$\widehat\Sigma_L:=\frac1L\sum_{\ell\le L}\widehat C_\ell$. Fix $0<\alpha\le\frac12$ and assume
$\|\widehat\Sigma_L-\Sigma\|_2\le\alpha M/((1-\alpha)\sqrt k)$. Define
\[
  \widehat C_0:=\Sigma-\frac{1-\alpha}\alpha\big(\widehat\Sigma_L-\Sigma\big),
  \qquad
  \widehat K^\sharp:=\alpha\,\delta_{\widehat C_0}
   +(1-\alpha)\frac1L\sum_{\ell=1}^L\delta_{\widehat C_\ell} .
\]
Then $\int C\,\widehat K^\sharp(dC)=\Sigma$, $\|C-\Sigma\|_{(2,k)}\le M$ for
$\widehat K^\sharp$-almost every $C$, and $\widehat K^\sharp$ has support of cardinality at most
$L+1$.
\end{lemma}

\begin{proof}
The mean identity is immediate:
\[
  \int C\,\widehat K^\sharp(dC)=\alpha\widehat C_0+(1-\alpha)\widehat\Sigma_L
  =\alpha\Sigma-(1-\alpha)\big(\widehat\Sigma_L-\Sigma\big)+(1-\alpha)\widehat\Sigma_L=\Sigma .
\]
The original atoms obey the $(2,k)$-bound by hypothesis, and for the correction atom
\[
  \|\widehat C_0-\Sigma\|_{(2,k)}=\frac{1-\alpha}\alpha\|\widehat\Sigma_L-\Sigma\|_{(2,k)}
  \le\frac{1-\alpha}\alpha\sqrt k\,\|\widehat\Sigma_L-\Sigma\|_2\le M
\]
by the assumed bound. The support bound is clear from the definition.
\end{proof}

\begin{lemma}[Sequential coupling bound]\label{lem:seqcoup}
Let $\mathsf P$ and $\mathsf P^\sharp$ be the path laws of two predictable processes on $T_0$
steps. Suppose that at every time $t$, whenever the two histories agree up to time $t-1$, the two
one-step transition laws admit a coupling under which the next updates disagree with probability
at most $\alpha$. Then $d_{\rm TV}(\mathsf P,\mathsf P^\sharp)\le T_0\alpha$, and consequently
$\mathsf P(A)\le\mathsf P^\sharp(A)+T_0\alpha$ for every measurable event $A$ depending on the
update path.
\end{lemma}

\begin{proof}
Build the two processes recursively, using the assumed one-step coupling while the histories
agree and continuing arbitrarily once they differ. By a union bound the probability that the
paths ever disagree is at most $T_0\alpha$, and total variation distance is bounded by the
disagreement probability under any coupling.
\end{proof}

\begin{proposition}[Finite-support reduction for i.i.d.\ updates]\label{prop:G4}
Let $C_1,\dots,C_{T_0}$ be i.i.d.\ symmetric matrices with common law $K$, mean $\Sigma$ and
$\|C_t-\Sigma\|_{(2,k)}\le M$ almost surely, independent of $Z_0$, and let $Q_t$ be the Oja
iterates with constant Phase I step size $\eta_t\equiv\eta$. Let $\delta,\xi\in(0,1)$ and set
\[
  \alpha:=\frac\xi{4T_0},\qquad \beta:=\frac\xi{4T_0},
\]
and let $L$ be an integer with
\[
  L\ \ge\ C_0\,\frac{kT_0^2}{\xi^2}\,\log\frac{8dT_0}\xi .
\]
Assume that the hypotheses of \Cref{thm:F12} hold uniformly for every deterministic finite support
$\mathcal A_\star\subseteq\mathbb R^{d\times d}$ with
\[
  |\mathcal A_\star|\ \le\ m_\star:=T_0(L+1)+1 ,
\]
with the same constants $M,\eta,\gamma,\delta$. Then
\[
  \mathbb P\big(\|W_{T_0}\|_F>1\big)\le\frac\delta3+\xi .
\]
\end{proposition}

\begin{proof}
\emph{The atom array, drawn in advance.} On an auxiliary probability space, independent of
$C_1,\dots,C_{T_0}$ and of $Q_0$, draw
\[
  \widehat C_{t,1},\dots,\widehat C_{t,L}\ \overset{\rm i.i.d.}\sim\ K,
  \qquad t\le T_0,
\]
all independent, and write $\mathcal A$ for the array and $\{\widehat C_{t,\ell}:\ell\le L\}$ for
the \emph{time-$t$ row}. The rows are independent of one another; in particular the time-$t$ row is
independent of all rows at times $s<t$.

Put $\widehat\Sigma_t:=\frac1L\sum_{\ell\le L}\widehat C_{t,\ell}$ and
\[
  \mathcal G_t:=\Big\{\big\|\widehat\Sigma_t-\Sigma\big\|_2\le
   \frac{\alpha M}{(1-\alpha)\sqrt k}\Big\} .
\]
Since $K$ has mean $\Sigma$ and is $M$-bounded in $\|\cdot\|_{(2,k)}$,~\Cref{lem:empmean} applies:
the lower bound on $L$ is $L\ge C_0\alpha^{-2}k\log(2d/\beta)$ after absorbing the factor $16$ into
$C_0$, and therefore
\begin{equation}\label{eq:Gbeta}
  \mathbb P(\mathcal G_t^c)\le\beta\qquad\text{for every }t\le T_0 .
\end{equation}
On $\mathcal G_t$ define the correction atom and the replacement law of~\Cref{lem:meanpres},
\[
  \widehat C_{t,0}:=\Sigma-\frac{1-\alpha}\alpha\big(\widehat\Sigma_t-\Sigma\big),
  \qquad
  \widehat K^\sharp_t:=\alpha\,\delta_{\widehat C_{t,0}}
   +(1-\alpha)\frac1L\sum_{\ell=1}^L\delta_{\widehat C_{t,\ell}} ,
\]
and on $\mathcal G_t^c$ set $\widehat K^\sharp_t:=\delta_\Sigma$. By~\Cref{lem:meanpres} in the
first case and trivially in the second, \emph{in both cases}
\begin{equation}\label{eq:Gsharp}
  \int C\,\widehat K^\sharp_t(dC)=\Sigma,\qquad
  \widehat K^\sharp_t\big(\{\|C-\Sigma\|_{(2,k)}\le M\}\big)=1,\qquad
  \big|\mathrm{supp}(\widehat K^\sharp_t)\big|\le L+1 .
\end{equation}

\emph{The auxiliary process.} Define $C^\sharp_t\sim\widehat K^\sharp_t$ independently across $t$,
using randomisation independent of $\mathcal A$, and set
$Q^\sharp_t:=\mathrm{QR}(Q^\sharp_{t-1}+\eta C^\sharp_tQ^\sharp_{t-1})$. Let
$\mathcal F^\sharp_t:=\sigma(\mathcal A,C^\sharp_1,\dots,C^\sharp_t)$. By \eqref{eq:Gsharp},
\emph{with no conditioning on a good event},
\[
  \mathbb E\big[C^\sharp_t\mid\mathcal F^\sharp_{t-1}\big]=\Sigma,
  \qquad \big\|C^\sharp_t-\Sigma\big\|_{(2,k)}\le M\ \text{ a.s.},
\]
and $C^\sharp_t$ takes values in
\[
  \mathcal A^\sharp_\star:=\{\Sigma\}\cup\bigcup_{t\le T_0}
   \mathrm{supp}\big(\widehat K^\sharp_t\big),
  \qquad |\mathcal A^\sharp_\star|\le T_0(L+1)+1=m_\star ,
\]
which is $\mathcal A$-measurable.

\emph{Applying \Cref{thm:F12}.} Condition on $\mathcal A$. The auxiliary process is driven by
randomisation independent of $\mathcal A$, so its conditional law given $\mathcal A$ is again that
of a process with conditional mean $\Sigma$ and $\|C^\sharp_t-\Sigma\|_{(2,k)}\le M$, now with the
\emph{deterministic} finite support $\mathcal A^\sharp_\star$ of cardinality at most $m_\star$. The
initialisation $Z_0$ is independent of $\mathcal A$, so its law is unchanged by the conditioning.
\Cref{thm:F12} therefore applies and gives
\[
  \mathbb P\Big(\big\|U^\top Q^\sharp_{T_0}(V^\top Q^\sharp_{T_0})^{-1}\big\|_F>1
   \ \Big|\ \mathcal A\Big)\le\frac\delta3 ,
\]
and averaging over $\mathcal A$,
\begin{equation}\label{eq:Gaux}
  \mathbb P\Big(\big\|U^\top Q^\sharp_{T_0}(V^\top Q^\sharp_{T_0})^{-1}\big\|_F>1\Big)
   \le\frac\delta3 .
\end{equation}

\emph{Coupling.} We realise the two processes on a common space and verify the hypothesis
of~\Cref{lem:seqcoup} with per-step disagreement probability $\alpha+\beta=\xi/(2T_0)$. Draw $J_t$
uniformly from $\{1,\dots,L\}$ and $B_t\sim\mathrm{Bernoulli}(\alpha)$, independently of
$\mathcal A$ and of everything else, and realise \emph{unconditionally}
\[
  C_t:=\widehat C_{t,J_t},
  \qquad
  C^\sharp_t:=\begin{cases}
   \widehat C_{t,0},&\text{on }\mathcal G_t\text{ with }B_t=1,\\
   \widehat C_{t,J_t},&\text{on }\mathcal G_t\text{ with }B_t=0,\\
   \Sigma,&\text{on }\mathcal G^c_t .\end{cases}
\]
That $C_t$ is realised from the array on $\mathcal G^c_t$ as well as on $\mathcal G_t$ is what
keeps its law exact: conditioning the realisation on $\mathcal G_t$, an event about the time-$t$
row, would bias the row.

\emph{The two marginals.} The common past $\mathcal F_{t-1}$ is a function of $Q_0$, of
$(J_s,B_s)_{s<t}$ and of the rows at times $s<t$, so the time-$t$ row is independent of
$\mathcal F_{t-1}$. Two conditionings must be kept apart. Given $\mathcal A$ as well as the past,
$C_t$ is a uniform draw from the $L$ realised atoms of row $t$, so its law is the \emph{empirical}
measure of that row,
\[
  \mathcal L\big(C_t\mid\mathcal F_{t-1},\mathcal A\big)
   =\frac1L\sum_{\ell\le L}\delta_{\widehat C_{t,\ell}} ,
\]
which is random and is not $K$. Averaging over the time-$t$ row recovers the law: since $J_t$ is
uniform and independent and $\widehat C_{t,1},\dots,\widehat C_{t,L}$ are i.i.d.\ $K$ and
independent of $\mathcal F_{t-1}$, for every bounded measurable $f$
\[
  \mathbb E\big[f(C_t)\mid\mathcal F_{t-1}\big]
  =\frac1L\sum_{\ell\le L}\mathbb E\big[f(\widehat C_{t,\ell})\big]
  =\int f\,\mathrm dK ,
  \qquad\text{that is}\qquad
  \mathcal L\big(C_t\mid\mathcal F_{t-1}\big)=K ,
\]
which is the law of the original update; and
$\mathcal L(C^\sharp_t\mid\mathcal A,\mathcal F_{t-1})=\widehat K^\sharp_t$ by \eqref{eq:Gsharp},
in both of its cases. So the coupling has the two intended marginals, and in particular its
$\sharp$-marginal is the process to which \eqref{eq:Gaux} applies.

\emph{Disagreement.} The two updates differ only on $\{B_t=1\}\cup\mathcal G^c_t$. The first has
probability $\alpha$ by construction; the second has probability at most $\beta$ by
\eqref{eq:Gbeta}, and $\mathcal G_t$ is an event about the time-$t$ row, hence independent of
$\mathcal F_{t-1}$. Note that it is \emph{not} true that $\mathcal A$ is independent of
$\mathcal F_{t-1}$ --- the past is built from the earlier rows --- and only the independence of the
time-$t$ row from the past is used.

By~\Cref{lem:seqcoup}, $d_{\rm TV}(\mathsf P,\mathsf P^\sharp)\le T_0\cdot\frac{\xi}{2T_0}
=\frac\xi2$, and with \eqref{eq:Gaux},
\[
  \mathbb P\big(\|U^\top Q_{T_0}(V^\top Q_{T_0})^{-1}\|_F>1\big)
  \le\frac\delta3+\frac\xi2\le\frac\delta3+\xi . \qedhere
\]
\end{proof}

\begin{remark}[what $m_\star$ costs, and the condition on $L$]\label{rem:Gmstar}
Only $\log m_\star$ enters~\Cref{thm:F12}, through $\gamma$ and through
$p_1=\log\frac{12T_0(m_\star+1)}\delta$. With $L=\widetilde O(kT_0^2/\xi^2)$,
\[
  \log m_\star=O\Big(\log\frac{kdT_0}{\xi}\Big) ,
\]
so the order of $T_0$ is unchanged.
\end{remark}

\begin{remark}
The array must be drawn in advance, and for an adaptive update it cannot instead be drawn along
the path: the product class $\mathcal E_{r,\ell}$ of Theorem \ref{thm:F12} must be deterministic,
because the centring step of \Cref{prop:F1} needs $E$ to be
$\mathcal F_{t-1}$-measurable while the class elements are built from noise factors at times after
$t$. Conditioning on a path-dependent array would make the class deterministic but would destroy
the conditional mean, since the time-$s$ atoms ($s>t$) are drawn from $K_s(\omega_{s-1},\cdot)$
and $\omega_{s-1}$ contains $C_t$. Appendix E of Huang et al.\ draws its atoms from a single fixed
law, so the obstruction does not arise there and their construction does not transfer verbatim.
\end{remark}

%% file: appendices/experiments.tex
\section{Experiments}\label{app:experiments}

\begin{figure}[t] 
  \centering
  \begin{subfigure}[b]{\textwidth}\centering
     \includegraphics[width=0.9\textwidth]{figs/legend_horizontal_all_algos.png}
  \end{subfigure}
  \vspace{1em}  %
  \begin{subfigure}[b]{\textwidth}
    \centering
    \includegraphics[width=\linewidth]{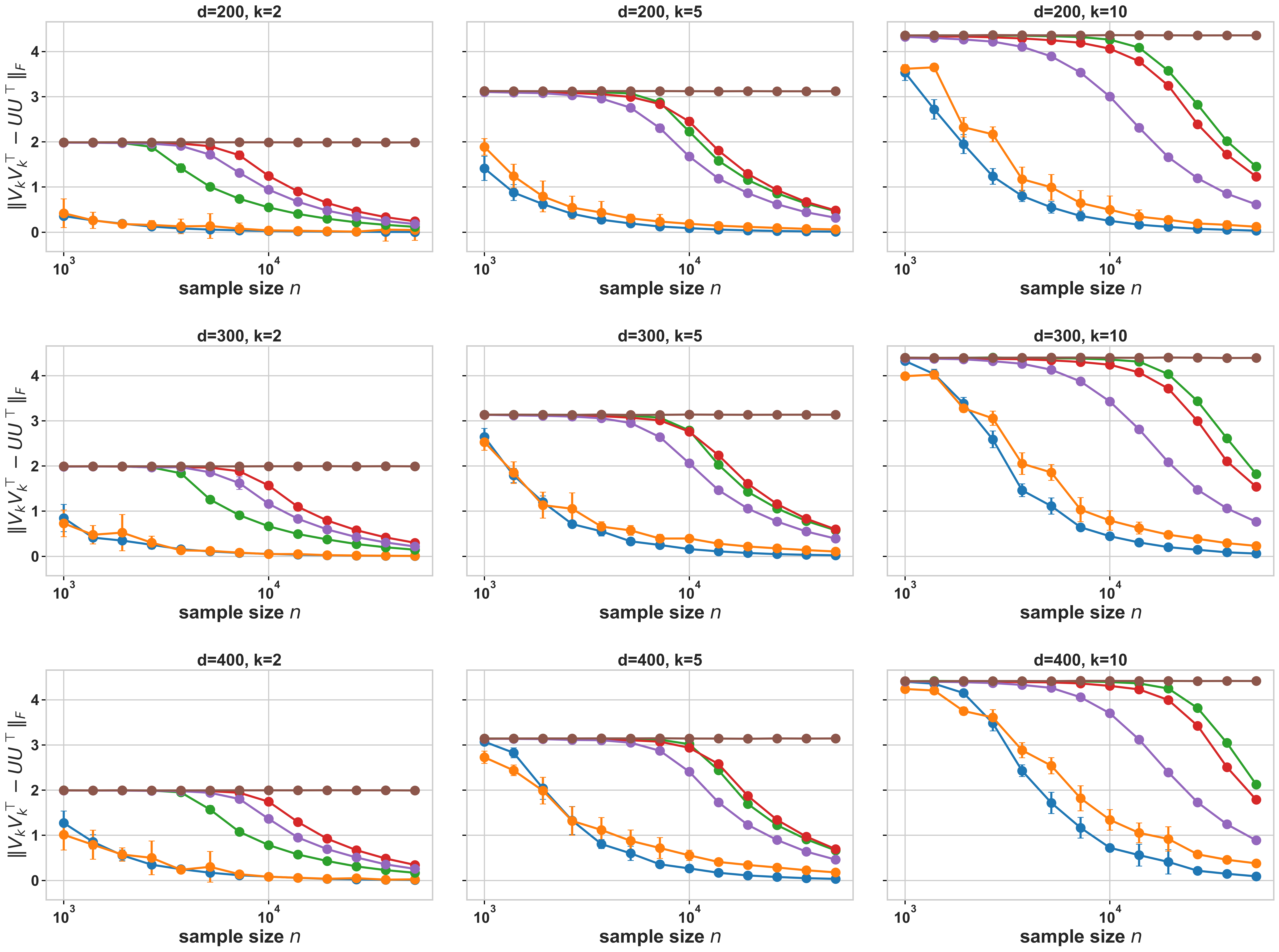}
  \end{subfigure}
  \vspace{-5 mm}
  \label{fig:ojas-vs-kdppca}
  \caption{\small  Comparison of \AdaptPrivOja with baselines for varying $k$ and $d$ on the spiked covariance model. We plot the mean over 10 trials, with the bars representing the standard deviation.}\label{fig:adadpojas-sweep-extended} 
\end{figure}

This appendix documents the implementation choices and algorithmic details omitted in the main paper. All experiments
were run under the replace-model \((\varepsilon,\delta)\)-DP setting with
\(\varepsilon=1\) and \(\delta=0.01\), unless explicitly stated otherwise. In~\Cref{fig:adadpojas-sweep} we give additional experiment results that had to be omitted from the main paper due to space constraints.

In~\Cref{subsec:experiments} we compare the performance of \AdaptPrivOja to \kdppca and \kdpojas~\citep{dungler2025iterative}, two adapted versions of the DP‐Gauss algorithms of \citet{dwork2014analyze}, we refer to as \texttt{DP-Gauss-1} and \texttt{DP-Gauss-2} respectively, and an adapted version of the noisy power method~\citep{hardt2014noisy}.

Given a stream of matrices \(\{A_i\}\) and a clipping threshold \(\beta\), chose $n$ according to the input distribution, \texttt{DP-Gauss-1} first rescales each matrix so that its trace is at most \(\beta^2\):\[    \tilde A_i    =    A_i \cdot \min\{1,\beta^2/\operatorname{Tr}(A_i)\}.\]It then forms the clipped sum \(X=\sum_i \tilde A_i\) and applies the Gaussian mechanism,\[    X' = X + E,\]where \(E\) is symmetric and its upper-triangular entries, including the diagonal, are sampled independently from \(\mathcal N(0,\Delta_1^2\mathbf I_d)\), with\[    \Delta_1    =    \frac{\beta^2\sqrt{2\log(1.25/\delta)}}{\epsilon}.\]The algorithm finally computes an eigendecomposition of \(X'\) and releases its top \(k\) eigenvectors. The second variant, \texttt{DP-Gauss-2}, uses the same clipping and summation steps to obtain \(X\). It then computes the top \(k\) eigenvectors \(V_k\) of \(X\), and privatizes the eigengap by setting\[    g_k = \lambda_k-\lambda_{k+1}+z,    \qquad    z\sim \operatorname{Lap}(2/\varepsilon).\]Next, it applies the Gaussian mechanism to \(V_k\):\[    W = V_kV_k^\top + E,\]where \(E\) is symmetric with independent upper-triangular entries, including the diagonal, sampled from \(\mathcal N(0,\Delta_2^2\mathbf I_d)\), and\[    \Delta_2    =    \frac{\beta^2\left(1+\sqrt{2\log(1/\delta)}/\epsilon\right)}    {\left|g_k-2\left(1+\log(1/\delta)/\epsilon\right)\right|}.\]Since the added noise may destroy orthogonality, we perform a final eigendecomposition of \(W\) and release the resulting top \(k\) eigenvectors. If \(g_k\leq 0\), the procedure is not differentially private, even though it follows Algorithm~2 of~\citet{dwork2014analyze}. A fully compliant version, also described in that work, would use the PTR mechanism, but this incurs additional privacy loss. For simplicity and to make the implementation more flexible, we instead resample the noise whenever \(g_k\leq 0\).Finally, \texttt{DP-Power-Method} clips the matrices using both the square root of the trace and the trace of the square root of the diagonal. In the rank-one case \(A=aa^\top\), with \(a\in\mathbb R^d\), these correspond respectively to clipping \(\|a\|_2\leq \beta\) and \(\|a\|_1\leq \alpha\), where \(\beta\) is chosen as in the DP-Gauss baselines. The algorithm then sums the clipped matrices and runs the noisy power method, using the version described in~\citet{nicolas2024differentially}. The Gaussian noise added at each power-iteration is scaled by an additional factor of \(\beta\alpha\).

\texttt{k-DP-PCA} and \texttt{k-DP-Ojas} are developed for stochastic data, and therefore need no adaptation. However, they both require a learning rate schedule and \kdppca requires a batch size. We discuss the hyperparameter tuning for this together with the hyperparameter tuning done for \AdaptPrivOja.

\subsection{Data Generation}

We generate synthetic data from a spiked covariance model. Each matrix
\(A_i\in\mathbb{R}^{d\times d}\) contains a deterministic rank-\(k\)
signal component together with a random noise component that makes the
sample full rank. For \(k=1\), we draw
\[
    x_i=s_i+n_i,
\]
where
\[
    s_i\sim \operatorname{Unif}\{\lambda_1 v,-\lambda_1 v\},
\]
\(v\in\mathbb{R}^d\) is a unit vector, \(\lambda_1\in\mathbb{R}\) is a
signal-strength parameter, and
\[
    n_i\sim \mathcal N(0,\sigma^2 I_d).
\]
We then set \(A_i=x_ix_i^\top\). The quantities \(\lambda_1\) and
\(\sigma\) are supplied as inputs to the sampler, while \(v\) is obtained by
sampling a standard Gaussian vector in \(\mathbb{R}^d\) and normalizing it.

For \(k>1\), we use a different construction. We first sample a matrix
\(V\in\mathbb{R}^{d\times k}\) with i.i.d. standard Gaussian entries and
orthonormalize its columns via Gram--Schmidt, obtaining
\(V_k\in\mathbb{R}^{d\times k}\). We then define
\[
    A_i = V_k \Lambda V_k^\top + z_i z_i^\top,
\]
where
\[
    z_i\sim \mathcal N(0,\sigma^2 I_d),
\]
and \(\Lambda\in\mathbb{R}^{k\times k}\) is diagonal with user-specified
eigenvalues. This \(k>1\) construction is not a direct generalization of the
\(k=1\) sampling scheme. Indeed, independently drawing \(k\) vectors as in the
rank-one case and summing their outer products would produce a mixture of
Gaussians rather than a single spiked covariance model. We therefore fix the
signal subspace and impose the rank-\(k\) structure deterministically through
\(V_k\Lambda V_k^\top\).

For \texttt{DP-Gauss-1} and \texttt{DP-Gauss-2}, we set
\[
    \beta = C\sqrt{\lambda_1}+\sigma\sqrt{d\log(n/\zeta)},
\]
where \(n\) is the sample size and \(1-\zeta\) is the target probability that
no clipping occurs. Throughout all experiments, we use \(\zeta=0.01\) for every
method. \(\AdaptPrivOja\), \(\kdppca\) and \(\kdpojas\) require the parameters
\(K\) and \(a\) from~\Cref{assumption-1}. For the synthetic model above, one has
\(a=1\) and \(K=O(1)\), so we take \(a=1\) and \(K=1\) in all experiments.

\subsection{Hyperparameter Tuning}

\paragraph{General protocol.}
Hyperparameters are tuned by running each candidate configuration across
the full sweep grid, averaging the target subspace error over all settings and
trials, and selecting the configuration with minimum mean error. Exact ties are
broken uniformly at random.

\paragraph{Learning-rate schedules.}
Let \(T\) denote the number of update steps, let \(n\) denote the total sample
size, let \(\tilde \lambda_1\) denote the leading population eigenvalue used by
the implementation, and let
\[
\Delta_1 := \tilde \lambda_1 - \tilde \lambda_2
\]
denote the leading eigengap. The learning-rate candidates used in the
experiments are as follows.

\begin{enumerate}
    \item
    For an offset \(c>0\), a cutoff fraction \(\rho \in (0,1]\), and an extra
    decay power \(p \ge 0\), define
    \[
    \tau := \left\lceil \rho T \right\rceil .
    \]
    The schedule is
    \[
    \eta_t
    =
    \frac{1}{t+c}
    \qquad \text{for } 1 \le t \le \tau,
    \]
    and
    \[
    \eta_t
    =
    \frac{1}{t+c}
    \left(\frac{\tau+c}{t+c}\right)^p
    \qquad \text{for } t > \tau.
    \]
    The two step-decay candidates used in the experiments are (\(\rho=\tfrac12\), \(p=\tfrac12\)) and (\(\rho=\tfrac13\), \(p=2\)).

    \item \textbf{\texttt{lambda1\_spiked}.}
    \[
    \eta_t
    =
    \frac{1}{
        20\,\sigma\,\tilde \lambda_1
        +
        \Delta_1 \, t / \log n
    },
    \qquad t=1,\dots,T.
    \]
    \item 
    \[
    \eta_t
    =
    \frac{1}{
        20\,\sigma\,\tilde \lambda_1\, s_d
        +
        \Delta_1 \, t / \log n
    },
    \qquad t=1,\dots,T,
    \]
    where \(s_d\) is a dimension factor. In the spiked-data implementation, \(s_d=d\) is replaced by the ambient eigenvalue-array
    length; in the restrained-Gaussian implementation, \(s_d=d\) exactly.
\end{enumerate}

\paragraph{Component-wise \(k\)-DP-PCA schedules.}
For \(k\)-DP-PCA, the implementation maintains a separate schedule for each
deflation step \(j=1,\dots,k\). Let
\[
\Delta_j := \tilde\lambda_j - \tilde\lambda_{j+1},
\]
 the component-wise learning rate is
\[
\eta_t^{(j)}
=
\frac{1}{
    b_j + \Delta_j \, t / \log n
},
\]
where
\[
b_j \in \{ 20\,\sigma\,\tilde\lambda_j, 20\,\sigma\,\tilde\lambda_j\, d\}.
\]

\paragraph{Adaptive Private Oja tuning.}
\AdaptPrivOja is tuned over:
\begin{itemize}
    \item batch rules
    \( B \in \{
    \sqrt{n}, n/\log n \}
    \)
    projected to an even batch size and then increased if necessary to satisfy the
    range-estimation feasibility condition;
    \item learning-rate offset
    \(
    10.0;
    \)
    \item learning-rate families as described above
\end{itemize}

\paragraph{\(k\)-DP-PCA tuning.}
\kdppca is tuned independently over:
\begin{itemize}
    \item batch rules
    \( B \in \{
    \sqrt{n}/k, n/(\log n \cdot k) \}
    \)
    together with an internal feasibility correction enforcing the truncation/range
    requirements of the implementation;
    \item the same learning-rate family set as Adaptive Private Oja.
\end{itemize}
The final sweep plots use the best Adaptive Private Oja configuration and best
\(k\)-DP-PCA configuration selected independently in this way.

\paragraph{\(k\)-DP-Ojas tuning.}

For \(\kdpojas\), we found empirically that a simple decreasing schedule,
independent of the eigenvalues, works well, and therefore set
\[
    \eta_j=\frac{1}{1+j},
    \qquad j\in[n],
\]
for all \(k\) iterations of \(\kdpojas\).

\paragraph{Finetuning in Tangent Space}
The hybrid method uses \AdaptPrivOja for the warm start and then runs \TangAlg on the remaining matrices. The two methods are both finetuned with the options described above, and allowed to have different batch sizes and learning-rates

\subsection{Error Metric}

All reported curves use the projection Frobenius error
\[
\|UU^\top - V_kV_k^\top\|_F,
\]
where \(U\) is the estimated \(k\)-dimensional subspace and \(V_k\) is the true
population top-\(k\) eigenspace.